\documentclass[preprint,1p,authoryear]{elsarticle}

\usepackage{graphicx}
\usepackage{multirow}
\usepackage{multicol}
\usepackage{amsmath,amssymb,amsfonts}
\usepackage{amsthm}
\usepackage{mathrsfs}
\usepackage[title]{appendix}
\usepackage{xcolor}
\usepackage{textcomp}
\usepackage{manyfoot}
\usepackage{booktabs}
\usepackage{algorithm}
\usepackage{algorithmicx}
\usepackage{algpseudocode}
\usepackage{listings}
\usepackage{ifpdf}
\usepackage{array}
\usepackage{url}
\usepackage{subcaption}
\usepackage{caption}
\usepackage{tabularray}
\usepackage{tikz}
\usepackage{float}
\usepackage{hyperref}
\usepackage{colortbl}
\usepackage{lscape}
\usepackage{rotating}
\usepackage{soul}
\usepackage{natbib}
\usepackage{todonotes}
\usepackage{etoolbox}
\usepackage[utf8]{inputenc}
\usepackage{enumitem}
\usepackage{pifont}
\usepackage[dvipsnames]{xcolor}
\usepackage{ragged2e}

\newcommand{\cmark}{\ding{51}}
\newcommand{\xmark}{\ding{55}}
\newcommand{\cback}[2]{{\setlength{\fboxsep}{0pt}\colorbox{#1}{#2}}}

\setcitestyle{numbers}

\begin{document}

\begin{frontmatter}

\title{Visual Hand Gesture Recognition with Deep Learning: A Comprehensive Review of Methods, Datasets, Challenges and Future Research Directions}

\author[1]{Konstantinos Foteinos}\ead{kfoteinos@hua.gr}
\author[1]{Manousos Linardakis}\ead{manouslinard@gmail.com}
\author[2,3]{Panagiotis Radoglou-Grammatikis}\ead{pradoglou@k3y.bg}
\author[4]{Vasileios Argyriou}\ead{vasileios.argyriou@kingston.ac.uk}
\author[2]{Panagiotis Sarigiannidis}\ead{psarigiannidis@uowm.gr}
\author[1]{Iraklis Varlamis}\ead{varlamis@hua.gr}
\author[1]{Georgios Th. Papadopoulos}\ead{G.Th.Papadopoulos@hua.gr}

\affiliation[1]{organization={Department of Informatics and Telematics, Harokopio University of Athens},
          addressline={Thiseos 70}, 
          city={Athens},
          postcode={GR 17676}, 
          state={Attiki},
          country={Greece}}
\affiliation[2]{organization={Department of Electrical and Computer Engineering, University of Western Macedonia},
          addressline={Active Urban Planning Zone}, 
          city={Kozani},
          postcode={GR 50150}, 
          state={Kozani},
          country={Greece}}
\affiliation[3]{organization={K3Y},
          addressline={Studentski district, Vitosha quarter, bl. 9}, 
          city={Sofia},
          postcode={BG 1700}, 
          state={Sofia City Province},
          country={Bulgaria}}
\affiliation[4]{organization={Department of Networks and Digital Media, Kingston University},
          addressline={53-57 High Street}, 
          city={Kingston upon Thames},
          postcode={KT1 1LQ}, 
          state={Surrey},
          country={United Kingdom}}

\begin{abstract}
The rapid evolution of deep learning (DL) models and the ever-increasing size of available datasets have raised the interest of the research community in the always-important field of \textbf{\textit{visual hand gesture recognition}} (VHGR), and delivered a wide range of applications, such as sign language understanding and human-computer interaction using cameras. Despite the large volume of research works in the field, a structured and complete survey on VHGR is still missing, leaving researchers to navigate through hundreds of papers in order to find the current state-of-the-art (SOTA). The current survey aims to fill this gap by presenting a comprehensive overview of this computer vision field. With a systematic research methodology that identifies the SOTA works and a structured presentation of the various methods, datasets, and evaluation metrics, this review aims to constitute a useful guideline for researchers, helping them to propose improvements. Specifically, this survey focuses on four fundamental questions: what are the main VHGR aspects, what are the current SOTA methods, what comparative insights can be drawn across methods and tasks, and which challenges shape future research. Starting with the methodology used to locate the related literature, the survey identifies and organizes the key VHGR approaches in a taxonomy-based format, and presents the various dimensions that affect the final method choice, such as input modality, task type, and application domain. The SOTA techniques are grouped across three primary VHGR tasks: static, isolated dynamic and continuous gesture recognition. For each task, the architectural trends and learning strategies are listed. To support the experimental evaluation of future methods in the field, the study reviews commonly used datasets and presents the standard performance metrics. Our survey concludes by identifying the major challenges in VHGR, including both general computer vision issues and domain-specific obstacles, and outlines promising directions for future research.
\end{abstract}


\end{frontmatter}

\section{Introduction}
\label{sec:intro}

    The recognition and interpretation of hand gestures \citep{ADADOGLOU} play a central role in advancing Human-Computer Interaction (HCI) and assistive communication systems. In recent years, deep learning-based vision-based hand gesture recognition (VHGR) has gained significant momentum as an active subfield of computer vision, building upon progress in related areas such as action~\citep{survey-HAR-introduction-shafizadegan2024multimodal} and facial expression recognition~\citep{survey-FER-JESWA-2024}. This has led to increased focus on Hand Gesture Recognition (HGR) across various applications. For example, HGR is widely used in Human-Robot Interaction (HRI) in industrial settings (e.g., robotic arm control \citep{Papadopoulos2021Access,ROBOTIC_ARM}), in healthcare (e.g., robot-assisted surgery \citep{TELEOPERATED_SURGICAL_ROBOT}), and in the autonomous navigation of mobile robots such as Unmanned Aerial \citep{MD-UHGRD_SA-YOLO} and Ground \citep{ULTRA-RANGE-UGV-Bamani,Linardakis2024Access} Vehicles (UAVs and UGVs). It also plays a key role in HCI, especially in VR/AR environments \citep{OO-dMVMT,EgoGesture_DATASET,BOAT-MI_VR}. In addition, the large population of deaf and hard-of-hearing individuals has driven research into automatic Sign Language Recognition (SLR) systems, aiming to improve communication and inclusion for these communities \citep{survey-deaf-mute-AIR-2024,NEURAL-SIGN-LANGUAGE-TRANSLATION-CVPR-CAMGOZ}.
    
    Given the importance of the applications mentioned above, HGR has grown into a broad research field, generally divided into two main approaches: vision-based (VHGR) and sensor-based. VHGR relies on visual data such as RGB and depth images, as well as features derived from them, such as pose heatmaps or optical flow. Sensor-based HGR, on the other hand, uses data from devices such as wearable gloves or surface electromyography sensors, and even non-wearable sources like acoustic signals. Researchers have also explored multi-modal methods that combine inputs from different sources. In this study, we focus on VHGR, as sensor-based methods often involve high cost and complex setup, making them less practical for real-world use. Vision-based approaches better align with the goal of HGR: to enable more natural, unobtrusive, and accessible gesture-based interaction.
    
    \begin{sidewaystable}[]
        \caption{Comparison of this survey with selected others. ML refers to traditional machine learning/computer vision techniques. The current survey mainly examines publications from the last four years, focusing on recent advancements regarding architectures, learning strategies, and related challenges, although older datasets that are relevant to this day have also been included.}
        \centering
        \tiny
        \setlength{\tabcolsep}{2pt}
        \renewcommand{\arraystretch}{3}
        \begin{tabular}{llllllll}
            \toprule
            \textbf{Publication} & \textbf{Range} & \textbf{Papers} & \textbf{Input data} & \textbf{Models} & \textbf{Scope} & \textbf{Comments} & \parbox{6cm}{\textbf{Aspects that this survey aims to complement}} \\
            \midrule
            \citep{ADADOGLOU} & 2016-2020 & - & Vision-based & DL & CSLR & \parbox{6cm}{Provides an in-depth comparison of existing approaches, loss functions and pretraining schemes for CSLR.} & \parbox{6cm}{This survey expands the scope including non-CSLR methods and examines more recent methods.} \\
            \midrule
            \citep{survey-5-9622242} & 2014-2020 & 98 & Vision-based & DL \& ML & General & \parbox{6cm}{Investigates the progress on hand gesture representation and Data Acquisition (DAQ) techniques as well as their performance and future research directions.} & \parbox{6cm}{It examines relatively older methods. The current work aims to emphasize recent DL approaches and related challenges and future perspectives.} \\
            \midrule
            \citep{survey-10-al2021deep}& 2014-2021 & 138 & \parbox{1.5cm}{\raggedright Sensor- \& Vision-based} & DL & SLR & \parbox{6cm}{Categorizes Continuous and Isolated SLR methods based on their architectures and discusses pose estimation for ISLR in detail.}  & \parbox{6cm}{This survey covers more topics including HCI \& HRI, and highlights task-specific architectural patterns and challenges for CSLR.}\\
            \midrule
            \citep{survey-11-adeyanju2021machine} & 2001-2021 & 649 & Vision-based & DL \& ML & SLR & \parbox{6cm}{Discusses relationships between research teams and DAQ techniques.}  & \parbox{6cm}{It includes relatively old methods. This survey emphasizes recently proposed ones in order to facilitate the evolution of the field.} \\
            \midrule
            \citep{survey-6-ojeda2022survey} & 2000-2020 & 571 & \parbox{1.5cm}{\raggedright Sensor- \& Vision-based} & DL \& ML & General & \parbox{6cm}{Provides many statistical results and bibliometric charts.}  & \parbox{6cm}{It does not identify contemporary challenges or future research directions; this work seeks to bridge this gap.} \\
            \midrule
            \citep{survey-3-el2022comprehensive} & 1998-2020 & - & \parbox{1.5cm}{\raggedright Sensor- \& Vision-based} & DL \& ML & SLR & \parbox{6cm}{Discusses in detail existing methods for SLR and provides multiple tables for quantitative comparison.}  & \parbox{6cm}{This survey analyzes newer approaches for SLR and expands taxonomy and comparison to more application domains, beyond SLR.}\\
            \midrule
            \citep{survey-16-joksimoski2022technological} & 2010-2021 & 80 & Vision-based & DL \&  ML & SLR & \parbox{6cm}{A Natural Language Processing (NLP) toolkit was utilized to filter publications. Insightful statistics are also provided.}  & \parbox{6cm}{Only a few methods are discussed and the current trends are not highlighted. On the contrary, this survey aims to identify trends and challenges.}\\
            \midrule
            \citep{survey-17-sarhan2023unraveling} & 2015-2023 & - & Vision-based & DL & ISLR & \parbox{6cm}{Provides statistical charts and summary tables based on a four-fold taxonomy.}  & \parbox{6cm}{This review introduces additional taxonomy criteria and covers more tasks.}\\
            \midrule
            \citep{survey-4-shin2024methodological} & 2014-2024 & - & \parbox{1.5cm}{\raggedright Sensor- \& Vision-based} & DL \& ML & General & \parbox{6cm}{Considers a variety of modalities and examines methods for each one.}  & \parbox{6cm}{The current work aims to discuss further multi-modality fusion strategies and identify more challenges.}\\
            \midrule
            \citep{survey-8-hashi2024systematic} & 2018-2023 & 256 & \parbox{1.5cm}{\raggedright Sensor- \& Vision-based} & ML & General & \parbox{6cm}{Covers various aspects of HGR, including DAQ techniques, hand segmentation and feature extraction.}  & \parbox{6cm}{Focuses exclusively on traditional ML approaches, which are currently surpassed by DL ones. This survey, searching for recent advancements, focuses on DL approaches.}\\
            \midrule
            \citep{survey-9-al2024advancements} & 2015-2024 & 58 & \parbox{1.5cm}{\raggedright Sensor- \& Vision-based} & DL \& ML & SLR & \parbox{6cm}{Examines various SLR methods and highlights the superiority of DL methods over older ones.} & \parbox{6cm}{It does not identify challenges and future perspectives; this survey aims to cover this, going deeper into details.} \\
            \midrule
            \citep{survey-13-tao2024sign} & 2017-2024 & - & \parbox{1.5cm}{\raggedright Sensor- \& Vision-based} & DL \& ML & SLR & \parbox{6cm}{Groups methods into RNN-based, CNN-based and transformers and discusses well-known benchmarks.}  & \parbox{6cm}{This survey expands the scope to non-SLU papers and seeks to provide methods comparison and future prospects.}\\
            \midrule
            \citep{survey-1-rahman2024comparative} & 2008-2024 & - & \parbox{1.5cm}{\raggedright Sensor- \& Vision-based} & DL \& ML & General & \parbox{6cm}{Covers a wide spectrum of gesture-based interaction and related technologies.}  & \parbox{6cm}{The considered methods are relatively old. This survey aims to identify more recent advancements.}\\
            \midrule
            \citep{linardakis2025_hgr_survey} & 2018-2025 & 125 & Vision-based & DL \& ML & General & \parbox{6cm}{Discusses jointly hand gesture capturing and recognition techniques. Many challenges are highlighted, including data bias and privacy issues.} & \parbox{6cm}{It covers a limited number of datasets for HGR and relatively old methods; this survey complements it by analyzing more datasets and recent advancements.} \\
            \midrule
            \citep{survey-emporio2025continuous} & 2002-2023 & - & \parbox{1.5cm}{\raggedright Sensor- \& Vision-based} & DL \& ML & CDHGR & \parbox{6cm}{Focuses on online gesture recognition, discussing DAQ technologies and evaluation metrics, among others.}  & \parbox{6cm}{This survey covers more topics and provides comparison of state-of-the-art methods.}\\
            \midrule
            \citep{survey-khan2025deep} & 2010-2024 & - & \parbox{1.5cm}{\raggedright Sensor- \& Vision-based} & DL \& ML & CSLR & \parbox{6cm}{Discusses in detail methods and datasets for CSLR.} & \parbox{6cm}{It does not conduct methods comparison and the challenges are general; this survey highlights optimization difficulties in the context of CSLR.} \\
            \bottomrule
        \end{tabular}
        \label{tab:review-papers-comp}
    \end{sidewaystable}
    
    The great importance and wide scope of HGR have led researchers to conduct surveys focusing on different aspects of the field, as shown in Table \ref{tab:review-papers-comp}. Compared to existing review papers, this survey provides an in-depth examination of the recent deep learning (DL) methods (published since 2021), covering the current state-of-the-art (SOTA). In contrast, other existing reviews either focus on older approaches (published since 1998) \citep{survey-6-ojeda2022survey,survey-3-el2022comprehensive,survey-8-hashi2024systematic} or mainly cover traditional machine learning techniques \citep{survey-1-rahman2024comparative,survey-13-tao2024sign,linardakis2025_hgr_survey}. While such works are valuable for understanding the fundamental methods and providing statistical insights, they are less suited for reflecting the recent evolution of the field. The present survey stands out by its broader and more up-to-date scope, including 238 papers published since 2021--an average of nearly 47 publications per year. By comparison, \citep{survey-11-adeyanju2021machine} and \citep{survey-6-ojeda2022survey} analyze 649 and 571 papers, respectively, but span roughly 20 years of research, corresponding to only about 30 publications per year. Finally, the current survey focuses exclusively on VHGR. Including sensor-based approaches would significantly expand the scope and potentially confuse readers. Additionally, such methods generally offer lower user convenience and reduced applicability in real-world scenarios. Nevertheless, the review maintains a broad and inclusive perspective, acknowledging that approaches from different domains may share common principles and can inform each other.
    
    More specifically, the survey focuses on vision-based methods for three key VHGR tasks, which are discussed in detail in the following sections: static HGR (SHGR) \citep{FGDSNet,HAGRIDv2_ARXIV}, isolated dynamic HGR (IDHGR) \citep{SAM_SLR,SignBERT_ICCV,NLA_SLR,CTFB_Hampiholi,RAAR3DNet}, and continuous (dynamic) HGR (CDHGR or CHGR) \citep{VAC-CSLR,NEURIPS2022_6cd3ac24_HEATMAPS,CorrNet}. SHGR involves recognizing gestures from single frames or spatial data only. IDHGR extends this by analyzing short, pre-segmented sequences to capture both spatial and temporal patterns. CDHGR goes a step further by detecting multiple gestures within long, unsegmented video sequences. Each task presents its own unique challenges. For example, CDHGR often relies on weak supervision, making optimization more difficult. At the same time, all tasks share common issues such as interference from irrelevant background movements or variations in lighting, viewpoint, and hand appearance.
    
    VHGR has a long history, beginning with traditional computer vision techniques based on hand-crafted features and evolving into complex deep learning pipelines \citep{NVIDIA_GESTURE_R3DCNN_Molchanov}. Early approaches typically followed a three-stage process: pre-processing, feature extraction, and classification. Pre-processing often involved segmenting the hand region using skin color or other visual cues. Feature extraction relied on descriptors like Histogram of Oriented Gradients (HOG), while classification was usually performed using models such as Support Vector Machines (SVMs) or Hidden Markov Models (HMMs), depending on the task. With the rise of DL, these hand-crafted features were gradually replaced by learned representations, which offer better generalization. Modern deep learning architectures have become increasingly deep and sophisticated, especially for dynamic HGR tasks. These models first capture short-term motion patterns and then aggregate them to understand long-term context \citep{VAC-CSLR,TemporalLiftPooling,SMKD,SEN-CSLR,CorrNet}. Due to their superior performance, this study focuses primarily on DL-based methods.
    
    The main contributions of this study can be summarized as follows:
    \begin{enumerate}[leftmargin=*]
        \item A structured and comprehensive methodology for reviewing the VHGR literature;
        \item A broad overview of VHGR, organized according to multiple taxonomy criteria;
        \item A detailed analysis of recent methods, highlighting their key characteristics and innovations;
        \item A synthesis of current SOTA approaches along with a discussion of the key challenges facing the field.
    \end{enumerate}
    
    In particular, the key research questions that the current study aims to address are the following:
    \begin{enumerate}[label=Q\arabic*,leftmargin=*]
        \item What are the principal aspects of VHGR? Specifically, what are its main tasks, gesture types, general architectures, learning strategies, and application domains?
        \item What are the current SOTA methods for each category and the corresponding recent developments?
        \item What are the theoretical and quantitative insights obtained through the comparative analysis of methods, both per-task and cross-task?
        \item What are the current challenges and future research directions concerning the domain?
    \end{enumerate}
    
    The structure of this paper aims to guide the reader through a comprehensive review of VHGR. Section \ref{sec:literature-metho} presents the literature review methodology, outlining the systematic approach used to collect and evaluate relevant works. Section \ref{sec:taxonomy} introduces a taxonomy of VHGR methods based on diverse criteria, offering a structured classification of existing approaches. Sections \ref{sec:static}, \ref{sec:isolated}, and \ref{sec:continuous} delve into the three main VHGR tasks --static, isolated dynamic, and continuous gesture recognition-- discussing representative methods and their underlying principles. Section \ref{sec:comparison} provides a unified comparison of methods for VHGR regardless of the task. Section \ref{sec:datasets} presents the main datasets used for the evaluation of the presented methods and Section \ref{sec:metrics} summarizes the evaluation metrics commonly used in the field. Finally, Section \ref{sec:challenges} outlines key challenges in VHGR and highlights current solutions, while Section \ref{sec:conclusions} concludes the survey by reflecting on recent trends and emphasizing open challenges that shape the future of VHGR research.

\section{Literature Review Methodology}
    \label{sec:literature-metho}

    The literature review was conducted following a systematic methodology designed to ensure a comprehensive and structured exploration of the VHGR field. The study scope was first defined, guiding the development of inclusion and exclusion criteria for selecting relevant publications. Subsequently, targeted queries were executed across major scientific databases to retrieve works meeting these criteria. Finally, the selected literature was thoroughly examined to extract key innovations, contributions, and methodological insights, which are detailed in the following sections.

    \subsection{Survey Scope}
        
        This work sets out to map the evolving landscape of VHGR. By distilling the most impactful recent contributions, it seeks to clarify the field's current trajectory and highlight its most pressing research needs. The review is structured around five foundational pillars, which are crucial for answering the research questions, as mentioned in the previous section (Section \ref{sec:intro}):
        \begin{itemize}[leftmargin=*]
            \item Advanced deep learning \textbf{methods} introduced in recent literature;
            \item Benchmark \textbf{datasets} employed across various VHGR tasks;
            \item Established \textbf{evaluation metrics} used to measure the effectiveness of the proposed solutions;
            \item Key \textbf{challenges} that continue to shape the research agenda;
            \item Innovative \textbf{solutions} and \textbf{future directions} designed to address current limitations.
        \end{itemize}

    \subsection{Selection Criteria}
    
        To ensure the quality, relevance, and completeness of this review, a well-defined set of inclusion and exclusion criteria was established prior to initiating the literature search. The selection process was guided by three key axes:
        \begin{itemize}[leftmargin=*]
            \item \textbf{Relevance}: Only works that directly address VHGR were considered. This includes methods focused on hand gesture recognition and sign language recognition based solely on visual modalities such as RGB images, infrared (IR), optical flow, motion history, depth maps, and pose representations (e.g., skeleton landmarks and heatmaps derived either from RGB-based pose estimators or visual data acquisition devices). Studies centered on Human Action Recognition (HAR), sensor-based techniques (e.g., using electromyography or radar), or multimodal approaches combining visual and non-visual data were excluded to maintain a focused scope.
            \item \textbf{Up-to-date}: To capture the current state of the field, priority was given to publications from 2021 onward, especially those introducing new methods. Regarding datasets, no strict publication date was enforced; however, only those actively used by the community or released within the past four years were included. All surveyed review papers were published after 2021.
            \item \textbf{Impact}: To further emphasize seminal and influential works, publications were primarily sourced from highly cited venues (i.e., those with an h5-index of 90 or higher on Google Scholar), with a single exception for the IEEE International Conference on Automatic Face \& Gesture Recognition, due to its strong relevance to VHGR. As this h5-index criterion may not be sufficient when applied as a standalone criterion, potentially significant works may be discarded. Therefore, the references of the collected publications were examined to identify additional milestone or seminal works. Their selection was guided by their accuracy on common benchmarks, applicable only to methods, and by the number of citations. Furthermore, additional queries were submitted to identify highly cited publications, ensuring that the search included the majority of important works. Finally, to provide the theoretical context of the study, a small number of milestone papers were referenced, although they were not analyzed in detail in the main methods section.
        \end{itemize}
    
    \subsection{Literature Search}

        To identify publications meeting the predefined criteria, a comprehensive literature search was performed using specialized queries across established scientific databases, including \textit{Scopus} and \textit{Web of Science (WoS)}. \textit{Google Scholar} was additionally consulted to capture potentially overlooked works from major venues, and \textit{arXiv} was also used to find relevant high-impact pre-prints. Queries were constructed according to the inclusion/exclusion criteria, employing specific commands (e.g., \texttt{EXACTSRCTITLE}, \texttt{LIMIT-TO}) and Boolean operators (\texttt{AND}, \texttt{OR}, \texttt{NOT}) to refine results. The search targeted keywords in titles, abstracts, and index terms, including \textit{gesture recognition, gesture classification, sign language recognition, sign language understanding, pointing recognition, human finger pointing, hand gestures, human gestures, gesture, hand gesture, and body gesture}. This process retrieved over 900 publications, of which 210 were relevant to VHGR. Additional queries on Google Scholar and arXiv, with duplicates removed, produced a final set of 278 publications, encompassing review papers, methods, and datasets.
        
    \subsection{Categorizing the Identified Articles}

        After applying the inclusion and exclusion criteria, the remaining publications were carefully reviewed for relevance. Those that did not meet the criteria were excluded from further analysis. The selected papers were then categorized into three distinct groups: \textit{methods}, \textit{datasets}, and \textit{review papers}. Each category was analyzed separately, and the corresponding information was organized into three dedicated tables. Publications relevant to more than one category were included in each table as separate entries. This procedure resulted in 185 articles presenting methods, 72 presenting datasets, and 21 survey articles. Key information for each paper was systematically recorded in three tables with corresponding columns, as detailed below:
    
        \begin{itemize}[leftmargin=*]
            \item \textbf{Methods:} For each article, we recorded the name of the method it proposes, the datasets used for evaluation, the evaluation metrics, a brief description of the architecture, the input modality, and the learning strategy.
            \item \textbf{Datasets:} For each dataset, we recorded its name, the task based on temporal characteristics (e.g., static, isolated dynamic, or continuous), the data modality, the total number of samples, the number of subjects, and, when applicable to CDHGR, the number of gesture instances. For Sign Language Understanding (SLU) datasets, we also noted the language, vocabulary sizes (both sign and text), and total duration (for video-based datasets). Additional attributes include the source of the dataset (e.g., laboratory, television, or web), frame rate, resolution, and the environmental conditions under which the data were recorded.
            \item \textbf{Surveys:} For survey papers, we provide a concise description of the content. Furthermore, the proposed taxonomy, the time range covered, and the scope of the survey were recorded. Although we examine reviews considering both sensor-based and vision-based HGR techniques, we focus only on content related to VHGR.
        \end{itemize}
    
\begin{figure}[t]
        \centering
        \begin{tabular}{cc}
            \begin{subfigure}[t]{0.5\textwidth}
                \centering
                \includegraphics[width=\linewidth]{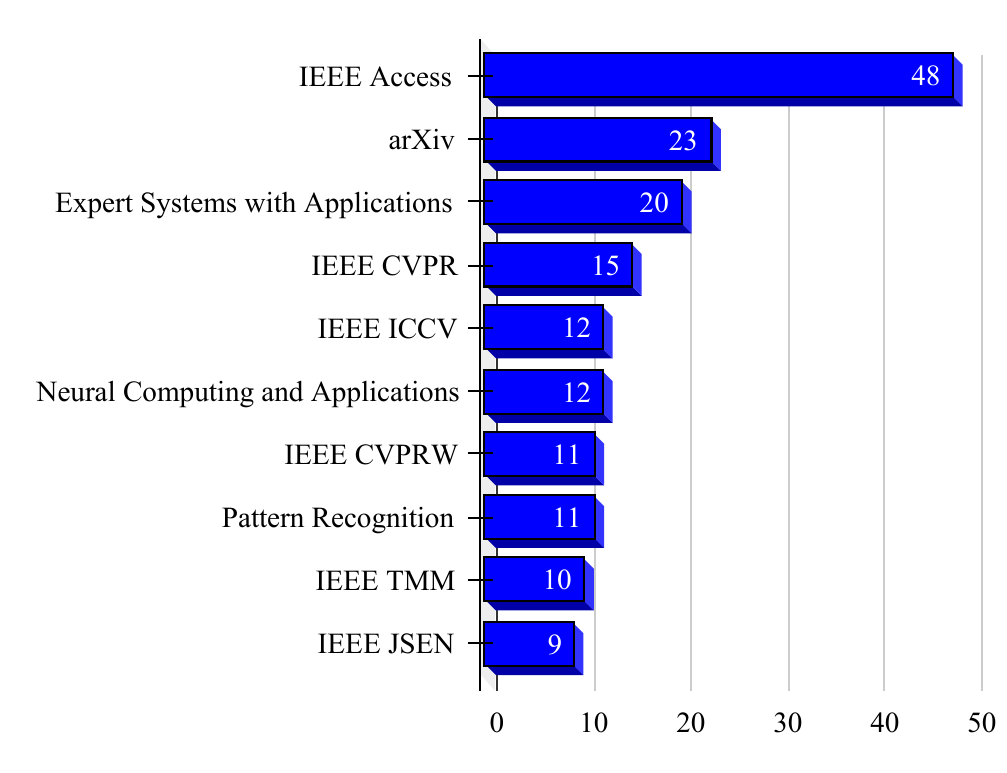}
                \caption{}
            \end{subfigure}&
            \begin{subfigure}[t]{0.38\textwidth}
                \centering
                \includegraphics[width=\linewidth]{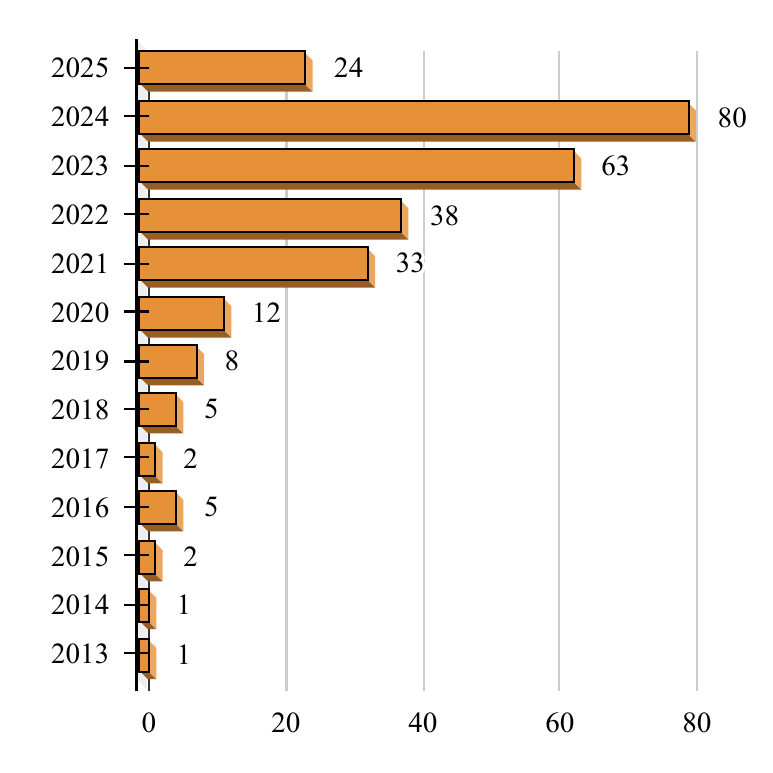}
                \caption{}
            \end{subfigure}\\
            \begin{subfigure}[t]{0.3\textwidth}
                \centering
                \includegraphics[width=\linewidth]{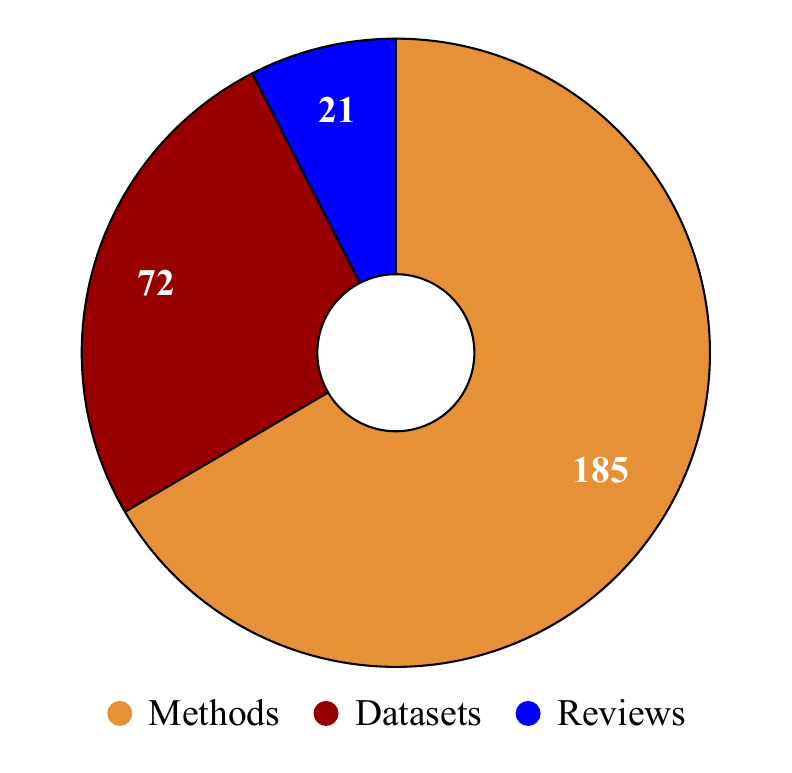}
                \caption{}
            \end{subfigure}&
            \begin{subfigure}{0.3\textwidth}
                \centering
                \begin{tabular}{cc}
                     \toprule
                     \textbf{Papers} & \textbf{Range} \\
                     \midrule
                     Methods & 2021-2025 \\
                     Datasets & 2010-2025 \\
                     Reviews & 2021-2025 \\
                    \bottomrule
                \end{tabular}
                \caption{}
            \end{subfigure}\\
        \end{tabular}
        \caption{Bibliometric statistics. Figure a) shows the distribution of the selected papers on the top-10 venues, in terms of number of publications. Almost 50 of the papers considered in the current survey have been published in IEEE Access, covering the whole spectrum of domains and tasks. Figure b) illustrates the number of publications \& pre-prints per year. As the current survey focuses on recent advancements, most of the selected papers have been published since 2021. Methods and reviews published before 2021 were ignored. The number of methods, datasets and reviews (which may overlap) are shown in figure c). The publication range for each category is reported in table d).}
        \label{fig:bibliometrics}
    \end{figure}

        For each publication, various metadata such as title, authors' names, venue, publication year, citation count, keywords, task, application domain, authors' motivation, challenges, and future directions were recorded. The most recent surveys have already been presented in Table \ref{tab:review-papers-comp} and discussed in the Introduction section, the methods and datasets are presented in an organized manner in the following Sections, and some bibliometrics are provided in Fig.~\ref{fig:bibliometrics}.

\section{Taxonomy of Hand Gesture Recognition Methods}
    \label{sec:taxonomy}

    \begin{figure}
        \centering
        \includegraphics[width=0.9\linewidth]{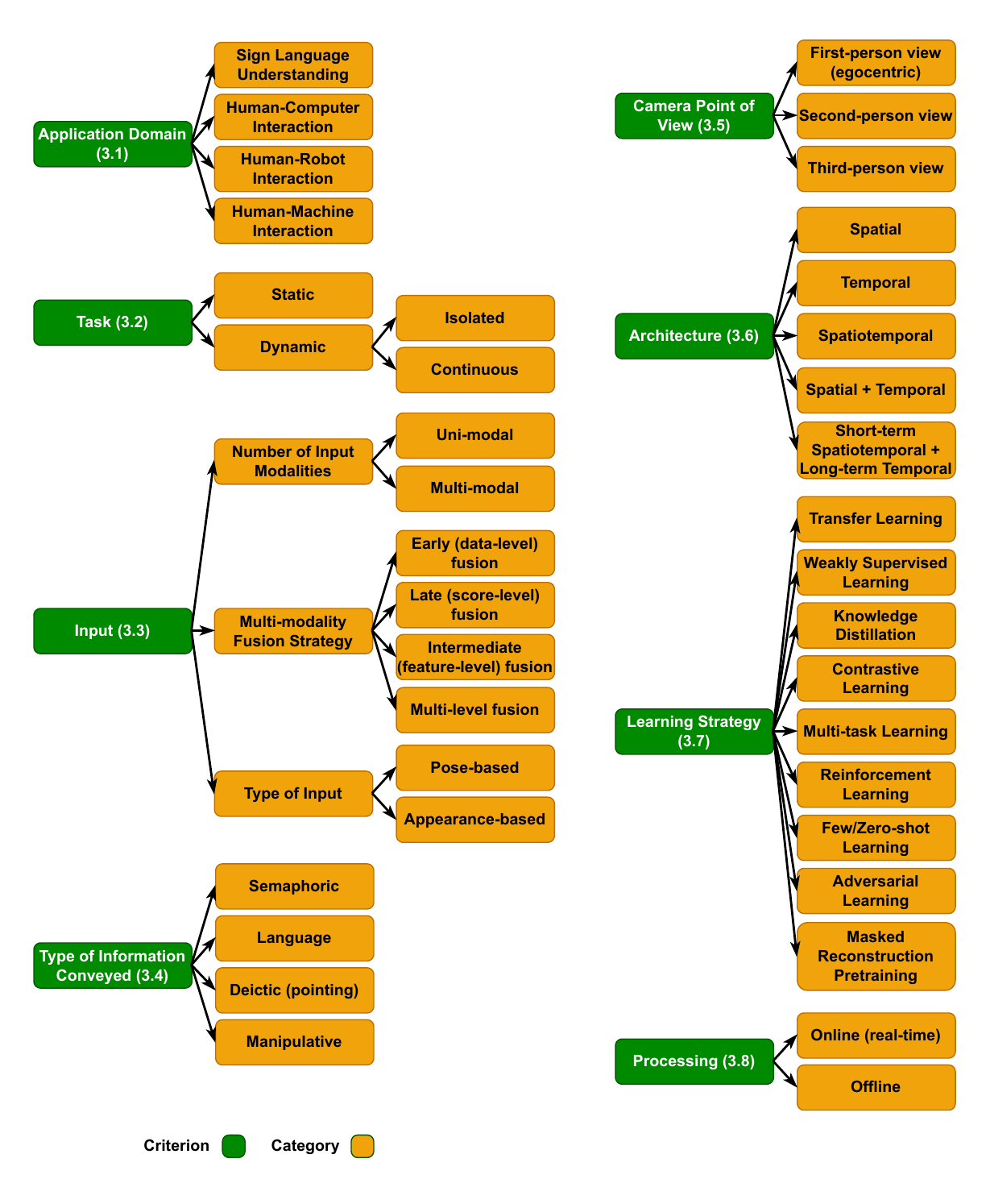}
        \caption{Taxonomy of hand gesture recognition methods. A diverse set of criteria has been proposed by many works. This survey adds a few more, including "Architecture", considering it essential for a method-oriented presentation.}
        \label{fig:taxonomy-chart}
    \end{figure}

    To better understand the diversity and evolution of vision-based hand gesture recognition (VHGR) methods and answer the first research question (see Q1 in Section \ref{sec:intro}), this section introduces a structured taxonomy of existing approaches. It classifies related works according to complementary criteria that capture key aspects of VHGR systems-such as input type, data representation, and application domain-each further divided into subcategories. The overall taxonomy is illustrated in Fig.~\ref{fig:taxonomy-chart} and described in detail below.

    \subsection{Application Domain}
    
        This criterion refers to the various domains in which VHGR solutions have been applied. The following categories are considered:
        
        \begin{itemize}[leftmargin=*]
            \item \textbf{Sign Language Understanding} (SLU): Involves the automatic interpretation of gestures in either a single sign language (monolingual SLU~\citep{UNISIGNli2025,BOBSL-BBC}) or across multiple sign languages (multilingual SLU~\citep{SIGN-muller-etal-2023-findings}).
            \item \textbf{Human-Computer Interaction} (HCI): Encompasses any form of interaction between users and computers. Numerous studies focus on HCI~\citep{HAGRIDv2_ARXIV,ZJUGesture}, with particular emphasis on VR/AR applications~\citep{EgoGesture_DATASET,OO-dMVMT,BOAT-MI_VR}.
            \item \textbf{Human-Robot Interaction} (HRI): Includes gesture-based control of mobile robots such as UGVs~\citep{ULTRA-RANGE-UGV-Bamani,MOBILE_ROBOT_DRCAM} and UAVs~\citep{LONG_RANGE_HGR_SSD,MD-UHGRD_SA-YOLO}, industrial robotic arms~\citep{ROBOTIC_ARM}, assistive robots~\citep{DIALOG_POINTING}, and surgical robots~\citep{TELEOPERATED_SURGICAL_ROBOT}.
            \item \textbf{Human-Machine Interaction} (HMI): Focuses on interfaces between humans and machines, especially in automotive contexts. VHGR enables recognition of traffic gestures (e.g., from police officers)~\citep{FA-STGCN,MG-GCT,CTCX_WANG2022123} and supports the development of more intuitive in-vehicle user interfaces~\citep{NVIDIA_GESTURE_R3DCNN_Molchanov,EgoFormer_METHOD_EgoDriving_DATASET,ConvMixFormer,Briareo_dataset,MVTN_ARXIV}.
        \end{itemize}

    \subsection{Task}
    
        Complementing the application domains, this criterion focuses on the nature of the gestures themselves, specifically their temporal characteristics. Gestures are broadly classified as either \textbf{static} or \textbf{dynamic}, depending on whether they involve a fixed hand posture or a sequence of movements over time. Dynamic gestures typically consist of three temporal phases: preparation, core (or nucleus), and retraction~\citep{NVIDIA_GESTURE_R3DCNN_Molchanov}, which reflect the gesture's evolution from initiation to completion. The recognition of dynamic gestures can be further divided into \textbf{isolated} and \textbf{continuous}, considering only one gesture per data sequence and identifying gestures in a continuous manner, respectively.
        
    \subsection{Input}
        
        In continuation of the previous criteria, which focused on application contexts and gesture types, this criterion examines the nature and variety of visual input data used for VHGR. The choice of \textbf{input modality} directly influences the representation architecture and the deep learning models employed. While RGB frames remain the most commonly used modality, other forms of visual data have also been explored-such as depth maps (including RGB-like representations, such as HHA \citep{SAM_SLR}), infrared (IR) images, skeleton joints (and derived features like bones and motion~\citep{TMS-NET}), pose heatmaps \citep{NEURIPS2022_6cd3ac24_HEATMAPS,NLA_SLR}, optical flow and motion history~\citep{MOTION_HISTORY_Mercanoglu}. 
        
        In terms of the \textbf{number of input modalities}, methods are classified as \textit{unimodal}, when they use a single type of input data, or \textit{multi-modal}, when they integrate two or more modalities. In the case of multiple modalities in the input, a \textbf{multi-modality fusion strategy} is chosen \citep{MMTM,CTFB_Hampiholi}. This could be:
            \begin{itemize}[leftmargin=*]
                \item \textbf{Early (data-level) fusion}: Combines modalities at the input level before processing~\citep{DHGR_3D_POSE_HRI,3STREAM_DISTILLATION}.
                \item \textbf{Late (score-level) fusion}: Fuses final prediction outputs from independent unimodal models~\citep{MOTION_HISTORY_Mercanoglu,JOINTS_MOTION_VECTORS_Improved_ST-GCN,GestFormer}.
                \item \textbf{Intermediate (feature-level) fusion}: Merges features extracted from separate modalities at a single processing level ~\citep{tang2021graph}.
                \item \textbf{Multi-level fusion}: Enables interaction across modalities at multiple stages of processing~\citep{NLA_SLR,CTFB_Hampiholi}.
            \end{itemize}

        Finally, based on the \textbf{type of input}, the methods are broadly categorized into two groups: \textit{pose-based} (or skeleton-based) approaches, which exclusively process skeleton data, and \textit{appearance-based} approaches, which rely on visual cues from RGB, depth, infrared, or similar modalities.

        \begin{figure}[h]
                \centering
                \begin{tabular}{ccc}
                    \multicolumn{3}{c}{
                        \begin{subfigure}[t]{0.7\textwidth}
                        \includegraphics[width=\textwidth]{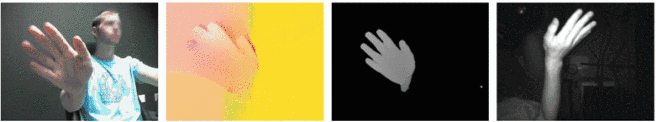}
                        \caption{\centering From left to right: RGB, Optical Flow, Depth and IR~\citep{NVIDIA_GESTURE_R3DCNN_Molchanov}}
                        \end{subfigure}
                    }\\
                        \begin{subfigure}[t]{0.2\textwidth}
                        \includegraphics[width=\textwidth]{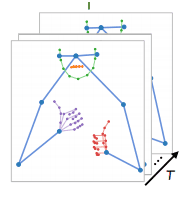}
                        \caption{\centering Skeleton joints~\citep{UNISIGNli2025}}
                        \end{subfigure}
                    &
                        \begin{subfigure}[t]{0.2\textwidth}
                        \includegraphics[width=\textwidth]{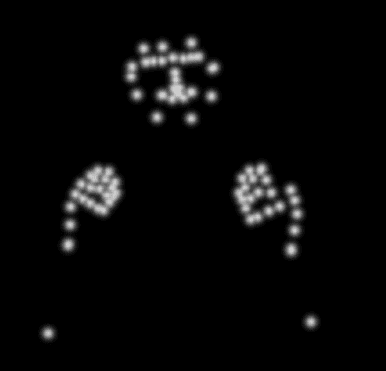}
                        \caption{\centering Pose heatmap~\citep{NEURIPS2022_6cd3ac24_HEATMAPS}}
                        \end{subfigure}
                    &
                        \begin{subfigure}[t]{0.2\textwidth}
                        \includegraphics[width=\textwidth]{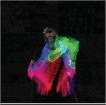}
                        \caption{\centering Motion history~\citep{MOTION_HISTORY_Mercanoglu}}
                        \end{subfigure}
                    \\
                \end{tabular}
                \caption{\centering Modalities that are considered for VHGR.}
                \label{fig:input_modalities}
        \end{figure}

    \subsection{Type of Information Conveyed}
              
\begin{figure}
            \centering
            \begin{tabular}{cc}
                \begin{tabular}{c}
                    \begin{subfigure}[t]{0.3\textwidth}
                    \includegraphics[width=\textwidth]{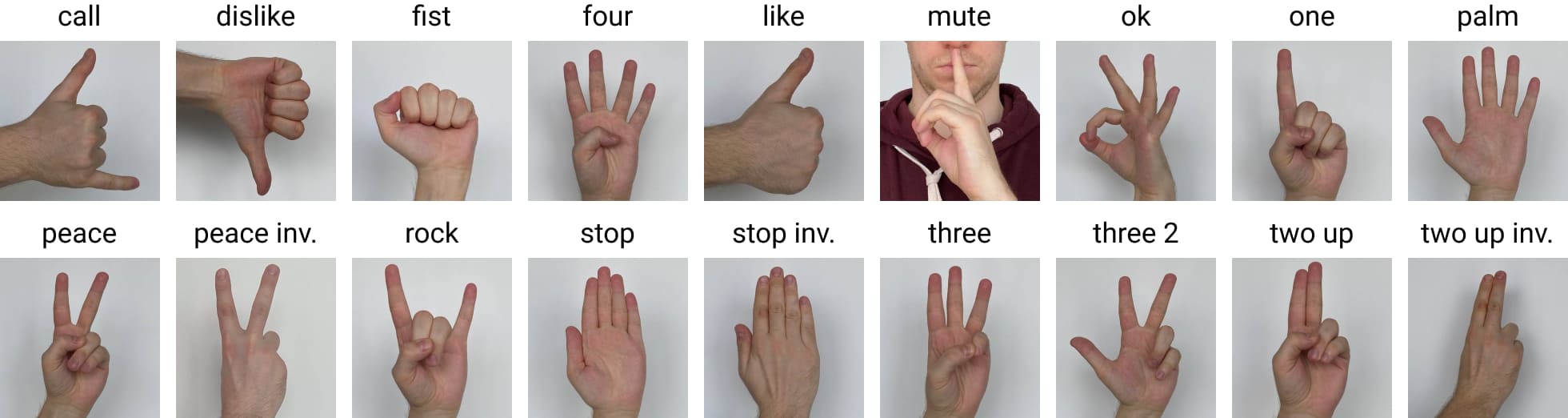}
                    \caption{\centering Semaphoric~\citep{WACV_kapitanov2024hagrid}}
                    \end{subfigure}
                \end{tabular}
                &
                \begin{tabular}{c}
                    \begin{subfigure}[t]{0.3\textwidth}
                    \includegraphics[width=\textwidth]{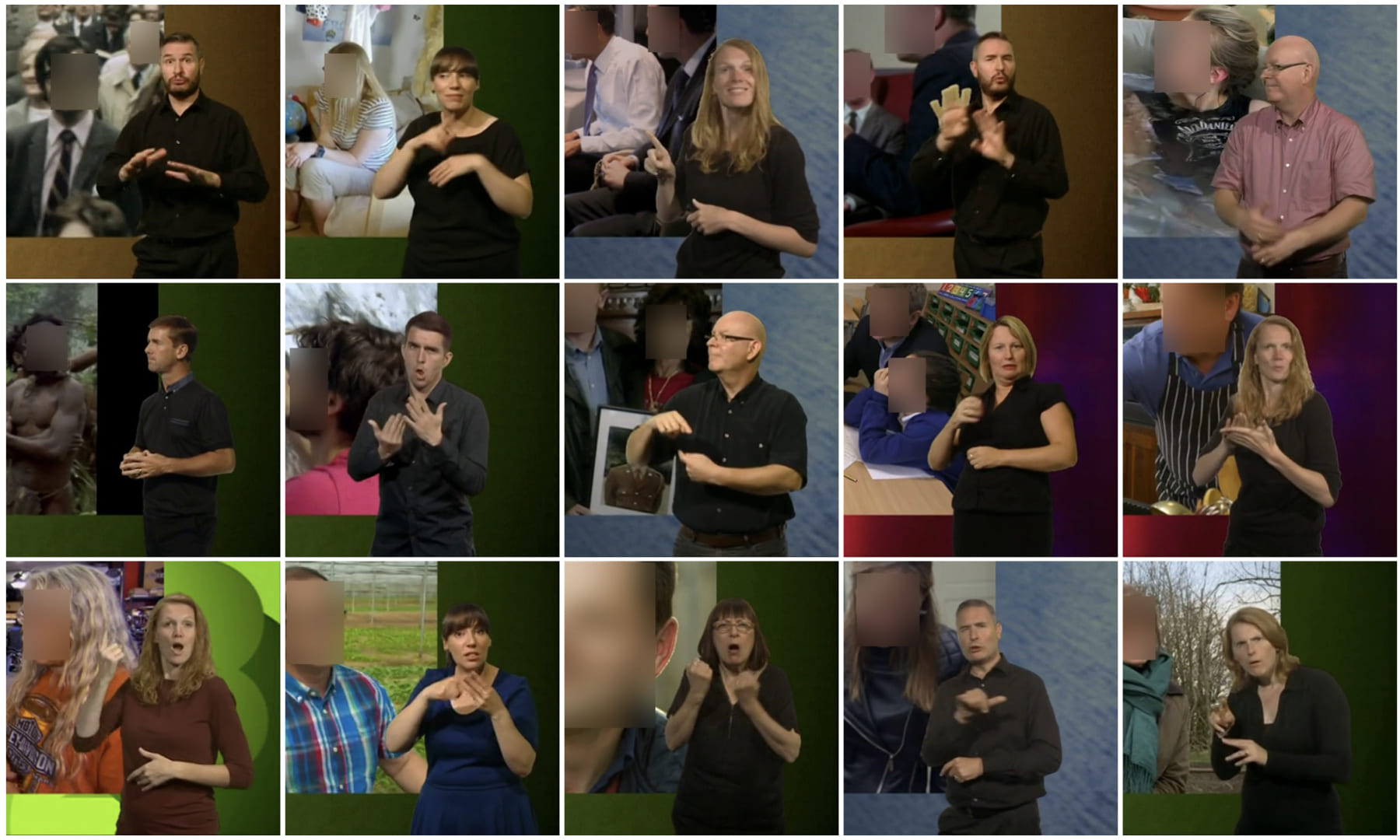}
                    \caption{\centering Language~\citep{BOBSL-BBC}}
                    \end{subfigure}
                \end{tabular}
                \\
                \begin{tabular}{c}
                    \begin{subfigure}[t]{0.3\textwidth}
                    \includegraphics[width=\textwidth]{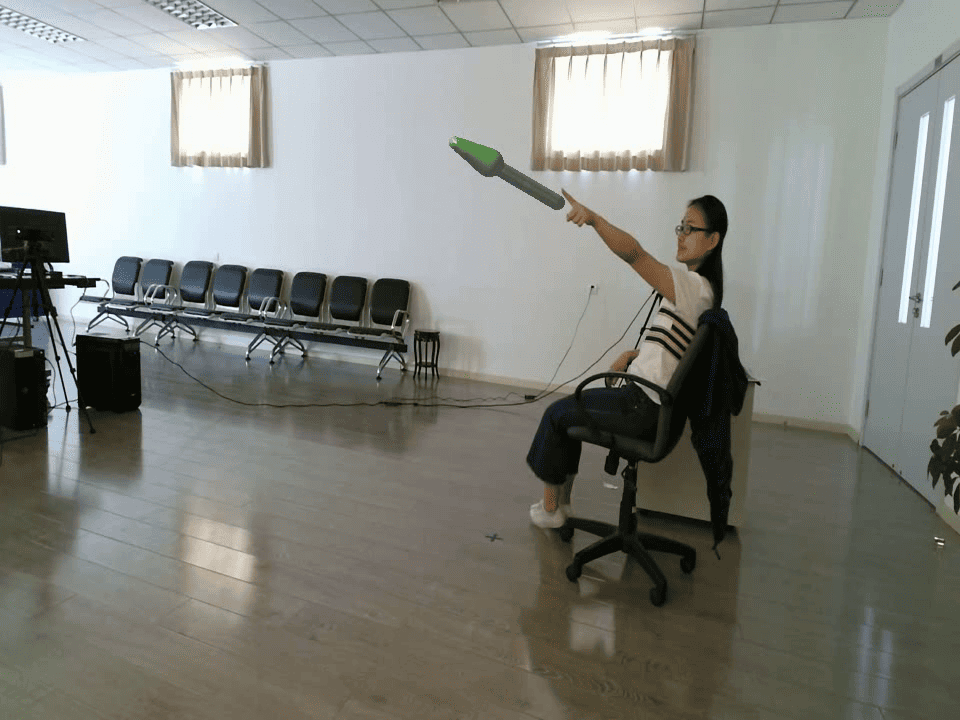}
                    \caption{\centering Pointing~\citep{DeePoint}}
                    \end{subfigure}
                \end{tabular}
                &
                \begin{tabular}{c}
                    \begin{subfigure}[t]{0.3\textwidth}
                    \includegraphics[width=\textwidth]{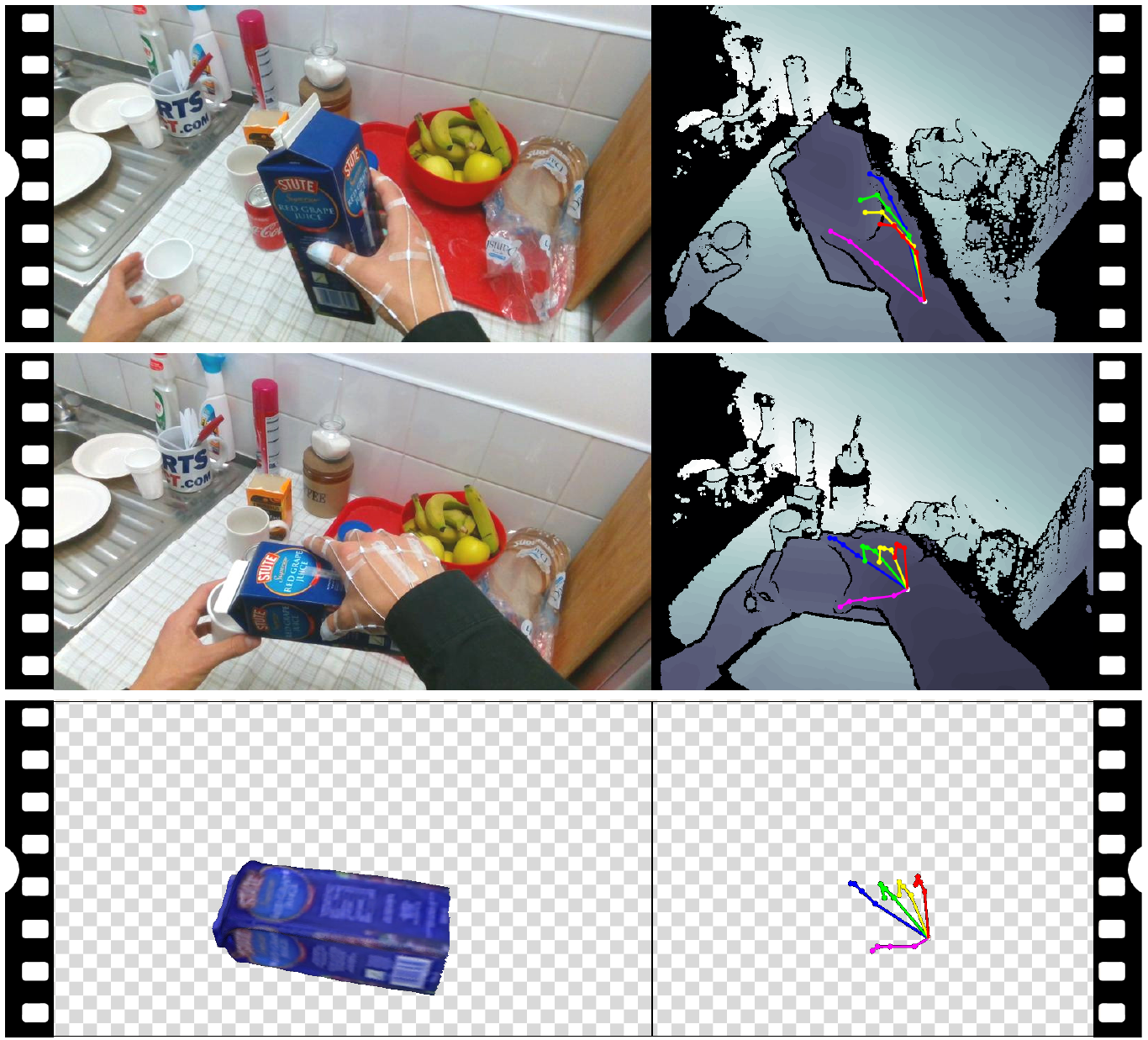}
                    \caption{\centering Manipulative~\citep{FPHA}}
                    \end{subfigure}
                \end{tabular}\\
            \end{tabular}
            \caption{\centering Main categories of hand gestures based on type of information conveyed.}
            \label{fig:transmitted_information}
        \end{figure}

        Building upon the previous criterion, which focused on the nature of the input, this dimension considers the \textbf{meaning} or \textbf{intent} behind the gestures. Specifically, it examines the kind of information gestures are meant to transmit~\citep{TAXONOMY_OF_HAND_GESTURES_2000_ASL_DATASET}, resulting in the following categories (illustrated in Fig.~\ref{fig:transmitted_information}):

        \begin{itemize}[leftmargin=*]
            \item \textbf{Semaphoric} gestures convey specific commands or instructions and are commonly used in gesture-based interfaces~\citep{NVIDIA_GESTURE_R3DCNN_Molchanov,HAGRIDv2_ARXIV,ULTRA-RANGE-UGV-Bamani}.
            \item \textbf{Language} gestures function similarly to semaphoric ones but are exclusively used in the context of sign languages~\citep{BOBSL-BBC,Auslan-Daily}.
            \item \textbf{Deictic (pointing)} gestures are used to indicate objects, directions, or spatial targets~\citep{DeePoint}.
            \item \textbf{Manipulative} gestures aim to interact with or mimic the manipulation of physical objects~\citep{FPHA}.
        \end{itemize}

    \subsection{Camera Point of View} 

\begin{figure}
        \centering
            \begin{tabular}{ccc}
                    \begin{tabular}{c}
                        \begin{subfigure}[t]{0.2\textwidth}
                        \includegraphics[width=\textwidth]{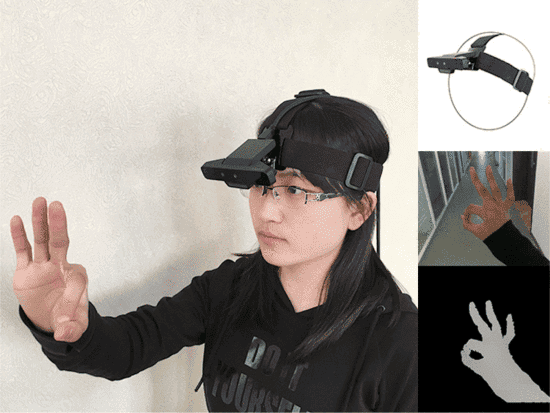}
                        \caption{\centering First-person view (egocentric) \citep{EgoGesture_DATASET}}
                        \end{subfigure}
                    \end{tabular}
                    &
                    \begin{tabular}{c}
                        \begin{subfigure}[t]{0.2\textwidth}
                        \includegraphics[width=\textwidth]{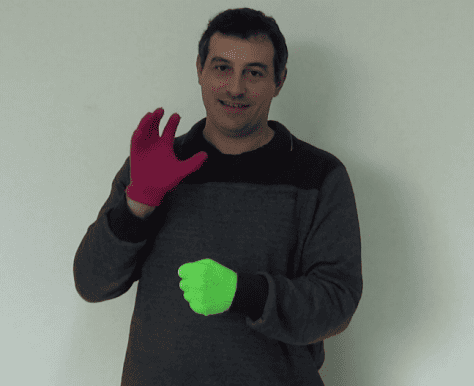}
                        \caption{\centering Second-person view \citep{Lsa64argentiniansignlanguage}}
                        \end{subfigure}
                    \end{tabular}
                    &
                    \begin{tabular}{c}
                            \begin{subfigure}[t]{0.3\textwidth}
                            \includegraphics[width=\textwidth]{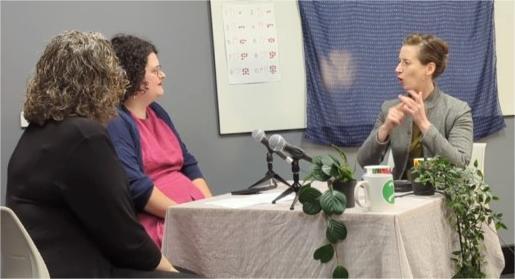}
                            \caption{\centering Third-person view \citep{Auslan-Daily}}
                            \end{subfigure}       
                    \end{tabular}
                    \\
            \end{tabular}
            \caption{\centering Different camera points of view for VHGR.}
            \label{fig:perception_view}
        \end{figure}

        Complementing the previous criteria on gesture content and input modalities, this dimension focuses on the spatial relationship between the camera and the performer~\citep{EgoGesture_DATASET}. The camera's perspective influences both the visual characteristics of the gesture and the recognition model design. Three primary viewpoints are considered (see Fig.~\ref{fig:perception_view}):

        \begin{itemize}[leftmargin=*]
            \item \textbf{First-person view (egocentric)}: The camera is mounted on the performer, capturing the gestures from their own perspective.
            \item \textbf{Second-person view}: The camera acts as the direct recipient of the gestures, typically simulating the perspective of a human observer or interacting agent.
            \item \textbf{Third-person view}: The camera observes the gesture from an external position, detached from the performer.
        \end{itemize}

        \subsection{Architecture}
        Building on the previous criteria, which focused on the nature of input and camera perspective, this criterion concerns the architectural design of the gesture recognition system, regarding the modeling stages performed by the backbone. Specifically, it refers to how spatial and temporal information is processed and integrated to model gesture dynamics. Three main categories are identified:
        \begin{itemize}[leftmargin=*]
            \item \textbf{Spatial} architectures consider only spatial data, such as isolated RGB frames, and feed them into corresponding DNNs (mainly 2DCNNs) to capture spatial information.
            \item \textbf{Temporal} architectures extract only temporal features, utilizing sequential models to process sequences of raw data, such as coordinates of skeleton joints.
            \item \textbf{Spatiotemporal} modeling architectures simultaneously analyze both the spatial and temporal domains, as in 3DCNNs.
            \item \textbf{Spatial + Temporal} pipelines first extract spatial features (e.g., using 2DCNNs) and then pass them through a sequential model (e.g., BiLSTM) to perform temporal modeling.
            \item \textbf{Short-term Spatiotemporal + Long-term Temporal} pipelines aim to capture local context from small parts of the overall sequence and then feed these spatiotemporal features into a temporal model to aggregate partial information and obtain a global vector representation. 
        \end{itemize}

    \subsection{Learning Strategy}
        
        This criterion explores the different training methodologies adopted by researchers to develop VHGR models, often influenced by data availability, task complexity, and desired generalization. The main strategies identified include:
        
        \begin{itemize}[leftmargin=*]
            \item \textbf{Transfer Learning}, where models are pretrained on large-scale datasets (e.g., Jester~\citep{Jester-Dataset}) and subsequently fine-tuned for specific downstream VHGR tasks~\citep{pretrained_on_jester_3dDenseNet_LSTM}.
            \item \textbf{Weakly Supervised Learning}, often used in CDHGR, where temporal annotations are coarse or unaligned, lacking exact gesture boundaries~\citep{Fangyun_Wei_ICCV}.
            \item \textbf{Knowledge Distillation}, in which one part of the network (often a teacher model) transfers knowledge to another (student model) to enhance feature extraction and efficiency~\citep{VAC-CSLR,C2ST,NEURIPS2022_6cd3ac24_HEATMAPS}.
            \item \textbf{Contrastive Learning}, used either in a CLIP-based framework to support zero-shot/few-shot recognition by leveraging textual descriptions~\citep{ZERO-SHOT-SLR,ZS-GR,FEW-SHOT-BILGE2024110374}, or to improve the discriminative capacity of learned representations~\citep{MASA,Weichao-Zhao}.
            \item \textbf{Multi-task Learning}, where a single model is optimized to perform multiple related tasks simultaneously, encouraging knowledge sharing and improved generalization~\citep{UNISIGNli2025}.
            \item \textbf{Reinforcement Learning}, applied to optimize decision-making processes, such as keyframe selection, by rewarding effective representations~\citep{KEYFRAMES_DRL_SKELETON}.
            \item \textbf{Few/Zero-shot Learning}, which enables the model to generalize to new gestures or languages with limited or no annotated examples, often leveraging semantic or textual guidance~\citep{ZERO-SHOT-SLR,ZS-GR,FEW-SHOT-BILGE2024110374}.
            \item \textbf{Adversarial Learning}, used as a data augmentation strategy \citep{alimisis2025advances} to improve robustness, as demonstrated in adversarial video synthesis networks~\citep{ADVERSARIAL_DATA_AUGM}.
            \item \textbf{Masked Reconstruction Pretraining}, inspired by BERT-style architectures, in which models are pretrained to reconstruct masked skeleton sequences, injecting structural prior knowledge (e.g., hand topology)~\citep{BEST,SignBERT+,SignBERT_ICCV}.
        \end{itemize}
        
    \subsection{Processing}
        
        Existing methods for isolated and continuous dynamic HGR can be further categorized based on whether the models have full access to the entire gesture(s) sequence or not. According to \citep{NVIDIA_GESTURE_R3DCNN_Molchanov}, two types of processing strategies apply:
    
        \begin{itemize}[leftmargin=*]
            \item \textbf{Online (real-time) processing}: Online HGR models perform gesture recognition in a stream of input frames. In other words, these models take as input the gesture sequence recorded until a specific time. Thus, online gesture recognition is suitable for real-time applications, as the gesture is recognized before the retraction part, reducing the latency. Sliding windows and gesture detector/classifier pipelines are often employed \citep{IPN-HAND,survey-emporio2025continuous}.
            \item \textbf{Offline}: Offline processing means that the model treats the entire gesture sequence. This approach is limited to applications that do not require real-time recognition. Most of the existing methods for VHGR fall into this category, such as those using BiLSTM for temporal modeling \citep{VAC-CSLR}.
        \end{itemize}

    The presented taxonomy of VHGR methods highlights the diversity of approaches developed to address different aspects of gesture understanding-from isolated dynamic gestures to sign language recognition and multimodal human-computer interaction. It provides a comprehensive foundation for analyzing current trends and identifying research gaps. Building on this overview, the following three sections delve deeper into specific VHGR categories, examining in detail the latest methods for Static (Section \ref{sec:static}), Isolated Dynamic (Section \ref{sec:isolated}), and Continuous Dynamic (Section \ref{sec:continuous}) gesture recognition, respectively.

\section{Methods for Static VHGR}
\label{sec:static}

\begin{figure}[t]
        \centering
        \setlength{\tabcolsep}{5pt}
        \renewcommand{\arraystretch}{5}
        \begin{tabular}{cc}
            \begin{subfigure}[t]{0.60\textwidth}
                \centering
                \includegraphics[width=\linewidth]{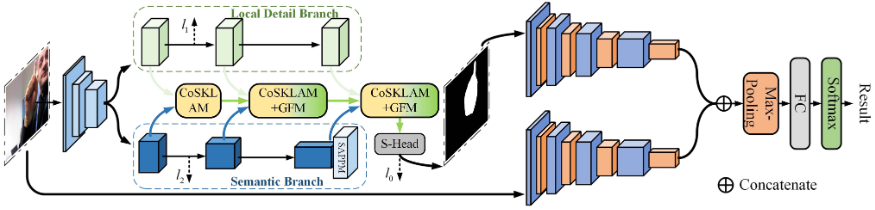}
                \caption{}
            \end{subfigure} &
            \begin{subfigure}[t]{0.30\textwidth}
                \centering
                \includegraphics[width=\linewidth]{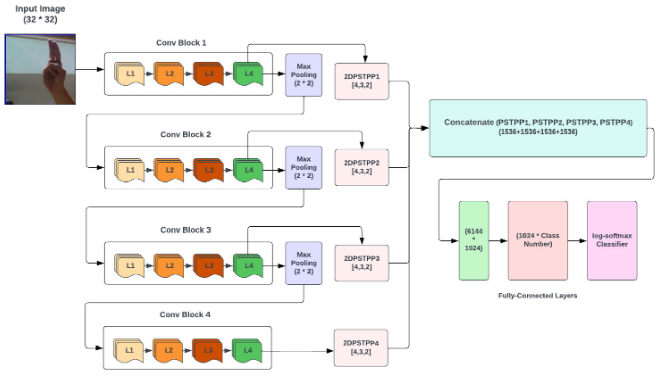}
                \caption{}
            \end{subfigure} \\
        \end{tabular}
        \caption{Representative networks for SHGR: FGDSNet \cite{FGDSNet} (a) proposes a dual-stream hand palm segmentation module followed by a classification model, while 2DPSTPP-Net \cite{2DPSTPP} (b) incorporates a novel pooling mechanism within a custom network with dense connections.}
        \label{fig:composite_static}
    \end{figure}

    Static VHGR (SHGR) aims to identify gestures relying only on spatial data. As this approach neglects temporal information, it is only suitable for recognizing static hand gestures. For gesture classification, fine-tuned variants of well-known 2DCNN architectures are employed, such as VGG and DenseNet, or custom lightweight architectures are trained from scratch. When classification is combined with localization of gestures, variants of YOLO are adopted. Hand segmentation is often applied to eliminate the background, reducing the interference of gesture-irrelevant information. Hand or human detection is also utilized in some cases to normalize the spatial dimensions, especially for long-range recognition. Regarding application domains, SHGR methods are used in SLU, HCI, and HRI. However, a few more general-purpose models have also been proposed. Representative network structures and algorithmic pipelines for SHGR are illustrated in Fig.~\ref{fig:composite_static}. The rest of the section analyzes representative methods in detail, highlights the current SOTA, and compares them, answering Q2 and Q3 (see Section~\ref{sec:intro}).

    \subsection{Sign Language Understanding (SLU)}
    
        For SLU, the authors in \citep{G-CNN-SHARMA2021115657} developed Gesture-CNN, a shallow 2DCNN to recognize Indian Sign Language (ISL) signs and slightly outperformed various versions of VGG on a self-collected dataset. The authors in \citep{STN-ND-Net} proposed the STN-ND-Net, a combination of a Spatial Transformer Network (STN) that improves robustness to spatial transformations, with a 2DCNN backbone that uses deconvolution operations to reduce the pixel-wise and channel-wise correlations and eliminates the information redundancy of colored images. The success of the vision transformer motivated the authors of SIGNFORMER~\citep{SIGNFORMER} to adopt it for ISL recognition, while authors in \citep{Jungpil-Shin} leveraged both skeleton and RGB data in their dual-stream feature extractor and an MLP to perform intermediate fusion. A YOLO variant, namely YOLOv4-CSP~\citep{YOLOv4-CSP} has also been developed, to simultaneously detect and identify Turkish Sign Language gestures.

    \subsection{Human-Computer Interaction (HCI)}
    
        For gesture-based HCI the authors in \citep{LHGR-Net} proposed a lightweight 2DCNN, named LHGR-Net, which combines a Multiscale Structure (MSS) that aims to make the model scale-invariant via atrous convolution with a Lightweight Attention Structure (LAS) based on CBAM. The authors in \citep{2DPSTPP} proposed 2DPSTPP-Net, a 2DCNN with four stacked convolutional blocks whose features are pooled using max pooling and a novel method, then concatenated to form the final representation, tailored for HCI scenarios. The authors in \citep{WEI_HAN_YOLOV5_COPULA} modified the YOLOv5 to make it more lightweight and developed a model that can be used for gesture-based user interfaces in smart sports stadiums, improving the experience of spectators and athletes. In \citep{Mohammad_Mahmudul_static_hgr}, a pipeline for egocentric HGR and fingertip localization uses YOLOv2 for hand detection, followed by VGG16 for feature extraction and gesture classification. In \citep{tuan_linh_pham_dang2023lightweight}, a lightweight HCI pipeline segments the hand using DeepLabV3+ or Unet, then classifies the cropped image with a 2DCNN like MobileNet or EfficientNet. \citep{FGDSNet} introduced FGDSNet, a two-stage approach with dual-stream segmentation (local detail and semantic branches) and multi-level fusion for feature integration. The segmentation mask and original image are fed into a dual-stream classification module, where mid-level fusion combines features for final classification via an MLP.

    \subsection{Human-Robot Interaction (HRI)}

        HRI, particularly gesture-based mobile robot navigation, has attracted research interest. URGR \citep{ULTRA-RANGE-UGV-Bamani} is a three-stage approach using YOLOv3 for human cropping, High-Quality Network (HQ-Net) for super-resolution to address long-range scenarios, and a classifier, Graph-Vision Transformer (GViT), for gesture recognition. The HQ-Net comprises convolutional and deconvolutional pathways. Features extracted by Canny Edge algorithm and a convolutional module followed by self-attention are fused with the latent representation. GViT utilizes two GCN layers that capture fine-grained details followed by a modified ViT. This setting enables effective recognition of gestures at a distance of up to 25m for UGV guidance. Similarly, \citep{DRCAM_TIM} introduced a two-stage method based on depth data: hand segmentation followed by gesture classification using DRCAM, a 2DCNN with stacked Residual Channel Attention Modules (RCAMs) and dense interconnections between the cascaded blocks. Because of the attention mechanisms, the model focuses effectively on the most informative regions of the image.

    \subsection{General Purpose}

        General-purpose models have also been developed. EDenseNet \citep{EDenseNet} extends DenseNet by adding convolutional layers between dense blocks. SpAtNet \citep{SpAtNet} is a lightweight 2DCNN with three streams: two  Multi-scale Attentive Feature Fusion (MAFF) modules for coarse-grained features and one Interleaved Module for high-level representations; their fused features are fed to an MLP for classification. This approach, employing multiple kernel sizes, increases accuracy by capturing various levels of granularity. MMFANet \citep{MMFANet} performs gesture detection and recognition using a 2DCNN backbone followed by the proposed feature aggregation pyramid network (FAPN) and task-specific regressor and classifier branches. The FAPN module aims to extract multiscale features through upsampling/downsampling operations and fusion of adjacent feature maps, while reducing the computational complexity by utilizing novel expanded convolutional blocks with different dilation rates combined with attention mechanisms that eliminate redundant information.

    \subsection{Comparison and Critical Summary of SHGR Methods}

        \begin{table}[t]
          \caption{Static VHGR: comparison of method families and key insights.}
          \label{tab:static_vhgr_methods_summary}
          \centering
          \tiny
          
          \setlength{\tabcolsep}{2pt}
          \renewcommand{\arraystretch}{0.9}
          \newcommand{\caspect}{1.5cm}
          \newcommand{\cslu}{2.8cm}
          \newcommand{\chci}{2.8cm}
          \newcommand{\chri}{2.8cm}
          \newcommand{\cgen}{2.8cm}
        
          \begin{tabular}{
          m{\caspect}
          m{\cslu}
          m{\chci}
          m{\chri}
          m{\cgen}
          }
            \toprule
            \textbf{Aspect} & 
            \textbf{Sign Language Understanding (SLU)} & 
            \textbf{Human-Computer Interaction (HCI)} & 
            \textbf{Human-Robot Interaction (HRI)} & 
            \textbf{General Purpose} \\
            \midrule
        
            Primary Function & 
                Automated recognition of gesture alphabets, numeric signs, and static fingerspelling
            & 
                Gesture-based control for VR/AR, smart environments, and device interaction
            & 
                Intuitive mobile robots (UGVs/UAVs) guidance, robotic arm control, and Robotic Assisted Surgery (RAS)
            & 
                Foundational feature extraction and gesture recognition across various contexts
            \\
            \midrule
            Most suitable scenarios &
            Small- or medium-scale static sign vocabularies and fingerspelling, especially when gestures are captured under relatively controlled conditions
            &
            Real-time command interfaces for VR/AR, smart environments, and edge devices where low latency is more important than large vocabulary coverage
            &
            Long-range robot guidance and robotic-control scenarios where human/hand localization, scale normalization, or super-resolution is required
            &
            Baseline recognition, transfer learning, or general-purpose feature extraction when the application does not impose strong domain-specific constraints
            \\

            \midrule
        
            Architectures & 
            \begin{itemize}[leftmargin=*,nosep]
                \item Shallow 2DCNNs
                \item 2DCNN variants (VGG, DenseNet)
                \item Vision transformers (ViT)
                \item YOLO variants
            \end{itemize} & 
            \begin{itemize}[leftmargin=*,nosep]
                \item Lightweight 2DCNNs
                \item Hand palm segmentation
                \item Multi-scale networks
                \item YOLO variants
            \end{itemize} & 
            \begin{itemize}[leftmargin=*,nosep]
                \item Spatial normalization and super-resolution
                \item Hybrid GCN-ViT classifiers
            \end{itemize} & 
            \begin{itemize}[leftmargin=*,nosep]
                \item Enhanced DenseNet
                \item Multi-stream 2DCNNs
                \item Pyramid feature aggregation
                \item Multiple kernel sizes
            \end{itemize} \\
            \midrule
        
            Strengths & 
            \begin{itemize}[leftmargin=*,nosep]
                \item High discriminative power for fine-grained details
                \item Capturing of global semantic dependencies
                \item Robust to spatial transformations
            \end{itemize} & 
            \begin{itemize}[leftmargin=*,nosep]
                \item Low-latency and real-time inference
                \item Suitable for resource-constrained edge devices
                \item Scale-invariance through (e.g., atrous) convolutions
            \end{itemize} & 
            \begin{itemize}[leftmargin=*,nosep]
                \item Effective ultra-range recognition at long distances (up to 25m)
                \item Robustness to motion blur
            \end{itemize} & 
            \begin{itemize}[leftmargin=*,nosep]
                \item High generalization ability
                \item Capturing of multiple levels of granularity
                \item Rotation and scale invariance
            \end{itemize} \\
            \midrule
        
            Limitations & 
            \begin{itemize}[leftmargin=*,nosep]
                \item High computational complexity for Transformer backbones
                \item Sensitivity to visually similar signs
                \item Requirement for large annotated datasets
            \end{itemize} & 
            \begin{itemize}[leftmargin=*,nosep]
                \item Difficulty with intricate background interference
                \item Limited gesture vocabulary in lightweight models
                \item High susceptibility to lighting
            \end{itemize} & 
            \begin{itemize}[leftmargin=*,nosep]
                \item High complexity of multi-stage pipelines
                \item Dependency on accurate human/hand detection
                \item Increased computational overhead for super-resolution
            \end{itemize} & 
            \begin{itemize}[leftmargin=*,nosep]
                \item Lack of domain-specific optimization
                \item High number of parameters
                \item Prone to overfitting on small, non-diverse datasets
            \end{itemize} \\
            \midrule
        
            Indicative Methods & 
                Gesture-CNN \citep{G-CNN-SHARMA2021115657}, STN-ND-Net \citep{STN-ND-Net}, SIGNFORMER \citep{SIGNFORMER}
            & 
                LHGR-Net \citep{LHGR-Net}, 2DPSTPP-Net \citep{2DPSTPP}, FGDSNet \citep{FGDSNet}
            & 
                URGR \citep{ULTRA-RANGE-UGV-Bamani}, DRCAM \citep{DRCAM_TIM}
            & 
                EDenseNet \citep{EDenseNet}, SpAtNet \citep{SpAtNet}, MMFANet \citep{MMFANet}
            \\
            \bottomrule
          \end{tabular}
          
        \end{table}

        Table~\ref{tab:static_vhgr_methods_summary} summarizes the main qualitative trade-offs among SHGR method families, while Table~\ref{tab:static-accuracy} reports the quantitative performance of representative methods. Most existing methods for SHGR are evaluated on different or custom datasets, which prevents a direct comparison of their performance. Despite their high accuracy and low parameter count -- suggesting that SHGR might be a largely solved problem -- the datasets used (ASL-FS \citep{ASL-FS-2011-CVPRW}, HGR1 \citep{HGR-1-2024-PATTER-RECOGNITION}, NUS-I \citep{NUS-I-2010}, NUS-II \citep{NUS-II-2013}, and OUHANDS \citep{ouhands}) remain relatively small (except for ASL-FS) and lack diversity. Consequently, model performance would likely degrade in real-world, uncontrolled environments. Among the existing methods, LHGR-Net \citep{LHGR-Net} and FGDSNet \citep{FGDSNet} demonstrate strong results on the OUHANDS dataset with lightweight architectures, making them suitable for gesture-based HCI. On the other hand, EDenseNet \citep{EDenseNet} achieves high accuracy on ASL-FS but at a higher computational cost, making it better suited for static sign language recognition.

        Combining Tables~\ref{tab:static_vhgr_methods_summary} and~\ref{tab:static-accuracy}, we can draw the following conclusions (answering partially Q3 in Section~\ref{sec:intro}):
        \begin{itemize}[leftmargin=*]
            \item Direct comparison of recognition accuracy should be interpreted cautiously, since methods are often evaluated on different datasets and not all works report computational cost, preprocessing overhead, or protocol-level details. Therefore, the quantitative comparison in Table~\ref{tab:static-accuracy} is intended mainly as a benchmark-oriented summary rather than a fully controlled comparison across methods.
            \item The RGB modality remains dominant, while depth information has rarely been leveraged \cite{DRCAM_TIM} and has been associated with lower accuracy. Only a few works have used multiple modalities, as standalone RGB-based approaches already achieve sufficient performance. Since these multimodal approaches have been evaluated on custom datasets, they are not shown in the table.
            \item Applying background removal through hand segmentation \cite{FGDSNet,tuan_linh_pham_dang2023lightweight,DRCAM_TIM} does not appear to significantly improve accuracy compared to end-to-end learning approaches.
            \item Slightly modified general-purpose classifiers \cite{tuan_linh_pham_dang2023lightweight,EDenseNet} show competitive performance compared to models specifically tailored to HGR.
            \item More recent works use a smaller number of parameters, revealing a shift toward resource-efficient recognition \cite{FGDSNet,SpAtNet}.
            \item Newer publications also consider multi-stream approaches \cite{FGDSNet,SpAtNet} to enrich the extracted features without substantially increasing computational cost.
        \end{itemize}

        \begin{table}[t]
            \centering
            \tiny
            \setlength{\tabcolsep}{1.5pt}
            \renewcommand{\arraystretch}{1}
            \caption{Performance of representative SHGR methods. Many approaches achieve high accuracy even with small model size, but the limited size and diversity of the datasets used may result in performance reduction during real-world deployment. Methods marked with star (*) perform hand palm segmentation before applying classification. "Architecture" refers to the structure of the classification model and "Params" to the total number of parameters. Results are reported as in the original publications and should be interpreted mainly within each benchmark dataset.}        
            \begin{tabular}{llllllllll}
                \toprule
                \textbf{Method} & \textbf{Year} & \textbf{Modality} & \textbf{Architecture} & \textbf{Params} & \textbf{ASL-FS} & \textbf{HGR1} & \textbf{NUS-I} & \textbf{NUS-II} & \textbf{OUHANDS} \\
                \midrule
                FGDSNet \citep{FGDSNet} * & 2024 &RGB & Dual-stream 2DCNN &1.20M& -&94.80\%&-&99.80\%&97.80\% \\
                SpAtNet \citep{SpAtNet} & 2024&RGB& Three-stream 2DCNN & 3.80M & - & 93.33\% & - & - & - \\
                \citep{tuan_linh_pham_dang2023lightweight} * & 2023&RGB & EfficientNetB0 & 5.34M & - & 98.15\%	& - & -	& 89.20\% \\
                LHGR-Net \citep{LHGR-Net}& 2023 &RGB& 2DCNN+lightweight attention &1.81M& -&93.15\%&-&-&98.57\% \\
                2DPSTPP-Net \citep{2DPSTPP} & 2023 &RGB & 2DCNN & 19.01M &  -	& - & 100\% & 99.61\% & - \\
                DRCAM \citep{DRCAM_TIM} * &2023&Depth & Attention-based 2DCNN & - &  93.44\% & - & - & - & - \\
                EDenseNet \citep{EDenseNet} & 2021 &RGB & Modified DenseNet & 7.77M  & 99.99\% & -	& - & 99.30\% & - \\
                \bottomrule
            \end{tabular}
            \label{tab:static-accuracy}
        \end{table}

\section{Methods for Isolated Dynamic VHGR}
    \label{sec:isolated}

    Unlike SHGR approaches, which fail to capture temporal dynamics and struggle with dynamic gestures, recent advances in \textbf{Isolated Dynamic Vision-based Hand Gesture Recognition (IDHGR)} have focused on recognizing individual gesture instances from spatiotemporal inputs such as RGB video sequences or skeleton joint trajectories.
    
    When gestures are part of a sign language, the task is referred to as Isolated Sign Language Recognition (ISLR). A notable distinction is that ISLR models frequently leverage both manual features (e.g., hand movements) and non-manual features (e.g., facial expressions, upper body motion), which can significantly enhance recognition accuracy. As a result, many ISLR approaches go beyond hand-centric analysis to incorporate richer contextual cues.
    
\begin{figure}[t]
        \centering
        \setlength{\tabcolsep}{5pt}
        \renewcommand{\arraystretch}{5}
        \begin{tabular}{cc}
            \begin{subfigure}[t]{0.35\textwidth}
                \centering
                \includegraphics[width=\linewidth]{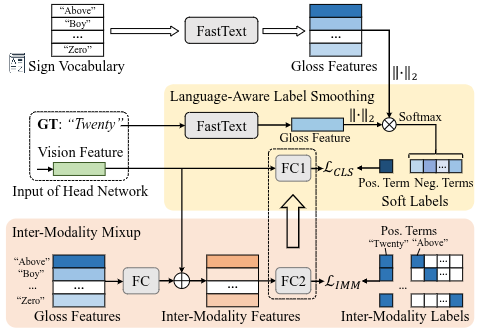}
                \caption{}
            \end{subfigure} &
            \begin{subfigure}[t]{0.55\textwidth}
                \centering
                \includegraphics[width=\linewidth]{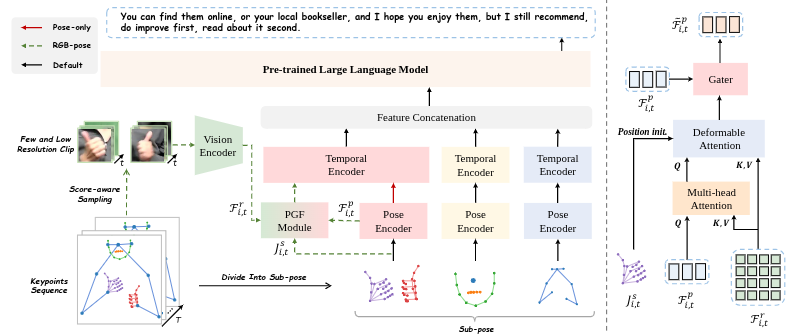}
                \caption{}
            \end{subfigure} \\
            \begin{subfigure}[t]{0.35\textwidth}
                \centering
                \includegraphics[width=\linewidth]{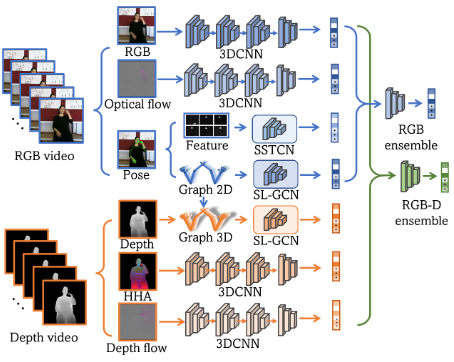}
                \caption{}
            \end{subfigure} &
            \begin{subfigure}[t]{0.55\textwidth}
                \centering
                \includegraphics[width=\linewidth]{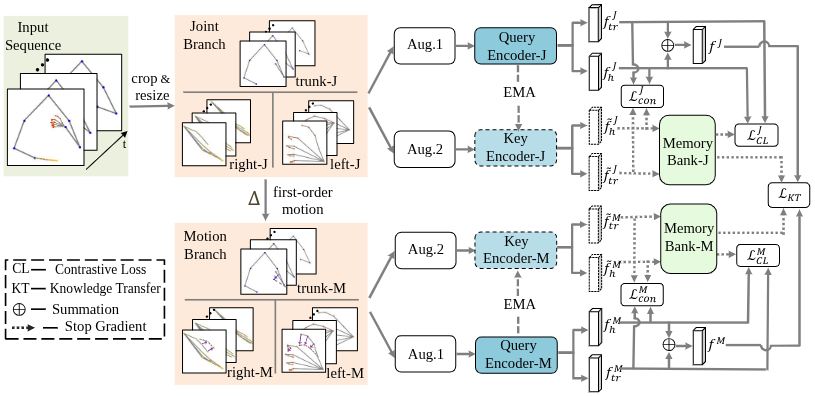}
                \caption{}
            \end{subfigure} \\
        \end{tabular}
        \caption{Representative network structures and algorithmic pipelines for IDHGR: NLA-SLR \cite{NLA_SLR} (a) introduces language-aware label smoothing to facilitate training, Uni-Sign \cite{UNISIGNli2025} (b) proposes a multi-task framework incorporating a language model, SAM-SLRv2 \cite{jiang2021sign-sam-slr-v2} (c) deploys several modalities for accurate recognition, and \cite{Weichao-Zhao} (d) applies consistency constraints in a self-supervised manner.}
        \label{fig:composite_idhgr}
    \end{figure}

    In the following subsections, we present IDHGR methods grouped by input modality as (a) \textbf{unimodal appearance-based}, (b) \textbf{unimodal pose-based}, and (c) \textbf{multimodal} approaches. Then, we conclude with a summary of key innovations and SOTA performance; this aims to answer Q2 and Q3 (see Section~\ref{sec:intro}). Representative network structures and algorithmic pipelines for IDHGR are illustrated in Fig.~\ref{fig:composite_idhgr}.

    \subsection{Unimodal Appearance-based Methods}

        Unimodal appearance-based methods rely on a single input type-such as RGB, optical flow, infrared (IR), motion history, depth, or depth flow-to classify hand gestures. As shown in Fig. \ref{fig:input_modalities}, these representations resemble color images, which allows for the application of similar computer vision techniques. These methods can be further categorized based on the architecture of the feature extractor, which typically follows one of three main pipelines: (1) spatiotemporal, (2) spatial + temporal, and (3) short-term spatiotemporal + long-term temporal backbones. The following paragraphs explain each of these architectures in more detail.
    
        \subsubsection{\textbf{Spatiotemporal}} 
    
        Spatiotemporal backbones perform spatial and temporal modeling jointly, typically using 3DCNNs or (2+1)D CNNs-either custom-built or adapted from existing models.
        
        In \citep{2plus1-D-SLR_and_NCSL}, the authors introduced a custom network, (2+1)D-SLR, composed of five stacked (2+1)D convolutional blocks (i.e., a 2D convolution followed by a 1D convolution). This structure reduces the computational overhead compared to traditional 3D convolutions, making the model more practical for deployment. To enhance temporal modeling, authors in \citep{Snehal_Bharti_et_al} proposed a trainable key frame selection method based on error correction. VGG16 features are used to identify key frames, which are then preprocessed and fed into a custom 3DCNN for classification.
        
        In \citep{ShiqiWang_et_al}, a dual-stage multi-stream pipeline for Chinese ISLR was developed. An adapted EfficientDet model-enhanced with cross-level Bi-FPN connections and a dual-channel spatial attention module (DCSAM)-detects both hands. The full image and cropped hand patches are passed through three parallel 3D-ResNet-18 models (following the P3D scheme), with their features fused and classified via an MLP. Authors in \citep{TS3C-Net} introduced TS3C-Net, a modified 3D ConvNeXt that adds an extra 3×3×3 convolutional branch to each block, enhancing spatiotemporal representation. Finally, \citep{SeST} addressed limitations in fixed-pipeline designs by combining features from both ResC3D and ConvLSTM using the Selective Spatiotemporal (SeST) module. The fused features are passed through fully connected and softmax layers for classification.
            
        \subsubsection{\textbf{Spatial + Temporal}} 
            
        These backbones first extract spatial features from each frame, typically using 2DCNNs or vision transformers, and then model the temporal dynamics with sequential modules such as GRUs or LSTMs. The resulting sequence embeddings are aggregated, and a classifier (e.g., MLP) outputs the final prediction.
        
        In \citep{DAVID_RICHARD_TOM_HAX}, a method tailored for HCI tasks using only depth data was proposed. The method combined Inception-v3 that extracted spatial features with an LSTM for temporal modeling. For ISLR, authors in \citep{GoogleNet-BiLSTM} introduced a pipeline to recognize Indian sign language words used by deaf farmers. They combined GoogleNet for spatial encoding with a BiLSTM for sequence modeling.
        
        To reduce inference time, a lightweight ISLR model, called TIM-SLR has been proposed in \citep{TIM-SLR}. The model uses a 2DCNN for per-frame feature extraction, followed by a Time Interaction Module (TIM)-a zero-parameter unit that fuses frame features into a compact representation. Finally, authors in \citep{Full_Transformer_Network} proposed a two-stage transformer-based model that extracts spatial features per frame, while a causally masked transformer processes the sequence, ensuring that only past and present frame features contribute at each step.
            
        \subsubsection{\textbf{Short-Term Spatiotemporal + Long-Term Temporal}}
    
        The methods in this category follow a two-stage architecture: the first stage captures short-term spatiotemporal patterns from segments of the input sequence, while the second stage aggregates these features to form a final representation of the entire sequence.
        
        EgoFormer \citep{EgoFormer_METHOD_EgoDriving_DATASET} is a model for egocentric dynamic gesture recognition optimized for edge devices, which has been introduced in the context of autonomous driving. EgoFormer combines a custom 3DCNN with a transformer module and an MLP for final classification. For long-range gesture-based UGV control, authors in \citep{SlowFast-Transformer} proposed a pipeline where a lightweight 2DCNN extracts per-frame features, which are used to select key frames. After cropping the human region, the selected frames are passed through a SlowFast network, followed by a transformer and an MLP.
        
        To address data scarcity in sign language understanding, \citep{ZERO-SHOT-SLR} proposed a zero-shot learning framework. Sign videos are represented using BERT-encoded textual descriptions combined with hand-crafted attributes. This representation guides a dual-stream backbone composed of short-term spatiotemporal modules (e.g., I3D) and long-term models (e.g., BiLSTM).
            
        \subsubsection{\textbf{Other}} 
            
        Some appearance-based methods fall outside the previously discussed categories but offer valuable contributions, especially in ISLR. For example, the method proposed in \citep{HAMZAH_LUQMAN} for Indian ISLR compresses an entire frame sequence into a single image, which is then processed by a MobileNet. A second MobileNet extracts per-frame features, which are passed through a two-layer LSTM. The outputs of both streams are concatenated and refined using a custom shallow 2DCNN. A different approach was presented in \citep{Ensemble-InceptionV4}, where the authors introduced a dual-stream ISLR architecture combining Inception V4 with a novel convolutional module called stem. \citep{Tianyu_Liu} proposed a dual-stream model with keyframe selection. A 2DCNN and attention mechanism filter out less informative frames. The remaining keyframes are processed by two branches: the Hand stream, which crops and analyzes hand regions with attention blocks, and the Full stream, which processes the entire frame. The fused features are used for final classification.
        
    \subsection{Unimodal Pose-based Methods}

        Pose-based methods, also known as skeleton-based approaches, rely on structured motion data-such as skeleton graphs and pose heatmaps-to classify gestures. These methods often incorporate various forms of pose information, including joint coordinates, bones, motion trajectories, and joint angles, sometimes in combination \citep{TMS-NET}. They also explore joint subsets related to specific body regions \citep{FDMSE-hierarchicalwindowedgraphattention}, often adopting multi-stream architectures to handle these diverse inputs.
        
        Given the natural graph structure of skeletal data, many methods employ Graph Convolutional Networks (GCNs) originally developed for HAR. In the sections that follow, we group the approaches based on the structure of their backbone, following the same organization as in the previous subsection. Notably, pose information alone is often insufficient for accurate sign recognition \citep{moryossef2021evaluating}. Consequently, many models integrate additional cues-typically from RGB inputs-using 3DCNNs and perform score-level fusion to enhance performance.

        \subsubsection{\textbf{Spatiotemporal}}

        Pose-based methods that incorporate spatiotemporal representations simultaneously process spatial and temporal data to improve SLU and HCI.

        In SLU, pose-based methods model dynamic hand and body movements for gesture recognition. SKIM \citep{SKIM} combines dynamic GCNs and temporal convolutional networks (TCNs) to analyze global and local motion, while MC-LSTM \citep{ozdemir2023multi_MC-LSTM} fuses multiple streams (face, hand, body) via temporal convolutions and GCNs. P3D \citep{P3D_ICCV} employs transformer-based part-wise and whole-body encoders, and HWGAT \citep{FDMSE-hierarchicalwindowedgraphattention} enhances key body-part focus with hierarchical attention. TMS-Net \citep{TMS-NET} introduces a six-stream GCN for skeleton, bone, and motion data with inter-stream fusion. To tackle data limitations, SLGTformer \citep{SLGTFORMER} and VSCT \citep{spoter_vsct} use transformers and few-shot learning, respectively. Dual-stream approaches such as \citep{Jungpil-Shin} and MULTISTAGE-CGAR \citep{MULTISTAGE-CGAR} integrate joint and motion data through temporal convolutions and attention. Other notable advances include SL-TSSI-DenseNet \citep{L-TSSI-DenseNet}, which converts joint coordinates into image-like matrices (termed Tree Structure Skeleton Image, TSSI) for DenseNet processing; STST \citep{STST}, a two-stage transformer for local-global spatiotemporal features; and DSTA-SLR \citep{DSTA_SLR_hu2024dynamic}, which employs dynamic adjacency matrices and multi-stream fusion for improved recognition accuracy. Dual-stream architectures such as STGCN-LSTM \citep{STGCN-LSTM-10965662} combine ST-GCN and self-attention ConvLSTM for capturing global and local features.
        
        In HCI, pose-based methods address challenges such as data overfitting and the need for self-supervised learning frameworks. MAE-based frameworks \citep{omar_ikne} use a masked autoencoder to reconstruct skeleton graphs, focusing on motion and spatial relationships. 3sISTGCN \citep{JOINTS_MOTION_VECTORS_Improved_ST-GCN} proposes a three-stream model to process joints, bones, and motions, applying modified ST-GCN layers and late fusion for improved classification. Another approach, DG-STA \citep{BOAT-MI_VR}, tackles class-incremental learning in HCI by training models on non-overlapping tasks without access to old training data. DSTSA-GCN \citep{DSTSA-CGN_cui2025dstsa} addresses temporal expression limitations by stacking three stages of graph convolutions and multiscale temporal convolutions. In another approach, TD-GCN \citep{TD-GCN} introduces a pose-based method for IDHGR. Unlike traditional methods that use fixed skeleton topologies, TD-GCN employs temporal-sensitive topology learning to enhance feature quality. First, graph convolution is used to extract high-level spatiotemporal representations, which are then used to calculate temporal and channel-independent adjacency matrices. These matrices help model the skeleton, and the features are fused using the proposed TCF module. Finally, authors in \citep{ST-CGNet-hu2025st} proposed ST-CGNet, a dual-stream architecture comprising a 3DCNN and a convolutional LSTM module to capture different levels of information. Attention mechanisms and an advanced fusion strategy are utilized to efficiently incorporate the features of each stream.
        
        \subsubsection{\textbf{Spatial + Temporal}}

        Many models use a two-step approach, where the spatial domain is first processed by a spatial perception module to extract spatial features. These features are then passed into a sequential model for temporal aggregation, resulting in a combined spatial-temporal representation of the input sequence. For instance, GCN-BERT \citep{GCN-BERT} utilizes a two-layer GCN for spatial modeling, followed by BERT for temporal modeling.
        
        Other general-purpose methods adopt a multi-stream, two-stage pose-based approach. In \citep{SBI-DHGR}, three streams, each containing three 1D convolutional layers, are combined with a parallel module of average pooling layers. The extracted features are then concatenated and input into an LSTM, followed by a fully connected layer and softmax for classification. Similarly, STr-GCN \citep{STr-GCN} employs a two-stage model where a GCN spatial feature extractor is first applied to obtain intra-frame features and then passes the features through a transformer encoder for modeling inter-frame temporal relationships. The final classification scores are generated by an MLP.
        
        \subsubsection{\textbf{Short-Term Spatiotemporal + Long-Term Temporal}}

        In this approach, the backbone first extracts features from smaller segments of the skeleton sequence to capture short-term spatial-temporal relationships. Then, long-term temporal modeling is applied to aggregate global context across the entire sequence.
        
        For example, gesture recognition in autonomous driving has become an important research area. In \citep{MG-GCT}, the authors propose a pose-based approach called MG-GCT for recognizing police traffic gestures. The model uses two graph convolution streams: one to process skeleton joints and another to analyze joint motions. The features extracted from both streams are fused and passed through a transformer encoder for temporal modeling, with the final classification performed by an MLP. Authors in \citep{ISLR-STGCN-attention-LSTM-ozdemir2025cross} proposed a dual-stream backbone processing both hand and whole body keypoints using ST-GCN models. Cross-attention is then applied to focus on the most important features, which are subsequently pass through an LSTM to model long-term dependencies.
        
        \subsubsection{\textbf{Temporal}}

        Some pose-based methods bypass spatial modeling and focus directly on sequential processing. Typically, keypoints are slightly preprocessed-often by extracting geometric features such as bone angles-and then passed into sequential models like RNNs or TCNs to produce a vector representation of the input sequence.
        
        In \citep{TELEOPERATED_SURGICAL_ROBOT}, the authors proposed a pose-based method for surgical robot teleoperation. Preprocessed skeleton joints are fed into an LSTM, and the resulting features are passed to an MLP for final classification. The model was evaluated on a custom dataset. Similarly, in \citep{Sabater_alberto}, a cross-domain gesture recognition approach was developed using geometric features (e.g., bone angles), which are processed by a TCN. The resulting features are summarized into a sequence descriptor, and a linear classifier predicts the final class. To improve signer-independence and reduce overfitting, the SwC GR-MMixer \citep{SwC-GR-MMixer} incorporates GRU units for temporal modeling and a shifted window concatenation strategy to enhance sequence representation.
        
        To address data scarcity, the authors in \citep{ADVERSARIAL_DATA_AUGM} introduced AVSN, an adversarial learning-based data augmentation framework. The method jointly optimizes augmentation parameters and the HGR classifier. A BiLSTM serves as the discriminator, while an MLP acts as the generator. The generator produces challenging augmented data that exploits weaknesses in the discriminator, which in turn learns to correctly classify gestures.

        \subsubsection{\textbf{Other}}

        Several works explore alternative architectures or skeleton representations beyond the main categories. In \citep{FS-EFCNN}, a Fourier CNN with BiLSTM extracts spatiotemporal features from 3D joints for ISLR, using a trainable binary mask for feature selection. Similarly, \citep{SCT-DCNN} applies feature thresholding based on variance and importance, followed by deep CNN classification. To overcome color-coded map limitations, \citep{MRCNNSA} introduces the Joint Motion Affinity Map (JMAM) with a multi-stream attention-based 2DCNN for feature extraction and recognition.
        
        Key frame selection is addressed in \citep{KEYFRAMES_DRL_SKELETON} via FS-Net, a reinforcement learning agent that selects informative frames for DD-Net classification.  \citep{Yangke_patcog} proposes a lightweight variant merging attention-based and autoencoder streams. Finally, the Hierarchical Self-Attention Network (HAN) \citep{HAN-liu2025han} employs modular attention across finger joints (J-Att, F-Att, T-Att, Fusion-Att) for efficient spatial-temporal modeling, achieving strong accuracy with just 0.05~GFLOPs.
    
    \subsection{Multimodal Methods}

        Multimodal approaches combine different input types (e.g., RGB, Depth, Skeleton, Text) to improve gesture recognition performance by leveraging complementary information. IDHGR multimodal methods do not seem to follow any trend, likely due to the wide spectrum of alternative options (combinations of streams, fusion techniques, modalities, etc.), so we group them by application domain.

        \subsubsection{\textbf{General Purpose}}

        In \citep{Yunan_Li}, a dual-stream 3DCNN with average fusion was proposed for RGBD gesture recognition, using a memory bank to retain discriminative features ($z$) while filtering redundant ones ($\upsilon$) by maximizing mutual information with class labels. RAAR3DNet \citep{RAAR3DNet} employed Neural Architecture Search (NAS) with Dynamic Static Attention (DSA) modules and a modified I3D backbone to emphasize motion and hand regions, supported by an auxiliary HeatmapNet and late RGBD fusion.

        To better exploit modality interactions, \citep{Benjia_Zhou} introduced a decoupling-recoupling framework combining independent spatial and temporal modeling with mid-level fusion through a shared MLP-AE module. Multi-level fusion was further explored in \citep{CTFB_Hampiholi}, integrating 3DCNNs with a Convolutional Transformer Fusion Block (CTFB) that fuses intermediate cross-stream features.

        To address data scarcity, several works explored zero-shot learning by incorporating textual gesture descriptions encoded by pretrained language models (mainly BERT). In \citep{ZS-GR}, a dual-stream model combined a vision transformer for spatial perception and an LSTM-MLP for temporal modeling, learning visual features aligned with textual embeddings. Similarly, \citep{ZS-DHGR} fused skeleton-derived geometric features with RGB-C3D-LSTM-AE representations into a shared semantic space for classification. To reduce computational cost, \citep{ConvMixFormer} replaced the transformer's self-attention mechanism with convolutional layers and a gated feedforward network, resulting in the lightweight ConvMixFormer architecture.

        Knowledge distillation and contrastive learning were applied in \citep{3STREAM_DISTILLATION}, where three 3DCNNs (RGB, Depth, Optical Flow) were fused with Coordinate Attention, using I3D as teacher and ResNet as student, while integrating CLIP for zero-shot recognition. Finally, \citep{MFST} and \citep{3D_CDC_NAS_TIP_2021} replaced standard convolutions with 3D Central Difference Convolution (3D-CDC), combining hybrid residual blocks, transformers, and NAS-based optimization for enhanced spatiotemporal modeling.

        \subsubsection{\textbf{Sign Language Understanding (SLU)}}

        Recent works on ISLR have focused on multimodal, pose-centric architectures combining RGB and skeleton data. SignBERT \citep{SignBERT_ICCV} employs a spatial GCN with temporal and hand chirality encoding, feeding features into a transformer encoder and a hand-model-aware decoder for self-supervised pretraining, with parallel RGB predictions fused at inference. Its enhanced variant, SignBERT+ \citep{SignBERT+}, adds spatial-temporal position encoding and an improved RGB-pose fusion module. Similarly, MASA \citep{MASA} uses the same GCN+Transformer backbone with contrastive learning on pose reconstruction, while BEST \citep{BEST} embeds pose triplets with BERT and applies late fusion with RGB data.

        To improve feature invariance, \citep{Weichao-Zhao} applies a contrastive approach enforcing consistency among hand, trunk, and motion features, using the same GCN+Transformer backbone with late fusion of pose and RGB predictions. SAM-SLR \citep{SAM_SLR} fuses six streams (RGB, depth, skeleton, optical flow, pose heatmaps, HHA), while SAM-SLR-v2 \citep{jiang2021sign-sam-slr-v2} adds a 3D hand graph stream and learnable fusion weights.
        
        VTN-HC/PF \citep{VTN-HC-PF} processes hand-cropped images via a 2DCNN, transformer, and MLP for final predictions, with PF incorporating pose-based geometrical features. Motion History Images are used in \citep{MOTION_HISTORY_Mercanoglu}, analyzed by a 3DCNN with attention, and in a dual-stream setup combined with 2DCNN RGB features via weighted summation.  

        To tackle data scarcity, ZSSLR \citep{ZSSLR} uses a CLIP-based zero-shot framework, while \citep{FSSLR} proposes a few-shot cross-lingual method combining two 3DCNNs and a pose estimator with a Temporal Feature Aggregation Module. NLA-SLR \citep{NLA_SLR} targets semantically similar signs with a multistream S3D backbone and language-aware label smoothing. MSNN \citep{MSNN} uses five I3D streams plus ST-GCN, and hDNN-SLR \citep{hDNN-SLR} combines a 3DCNN, attention-based BiLSTM, and a VAE-based module. Finally, Uni-Sign \citep{UNISIGNli2025} proposes a unified multimodal architecture with three pose encoders and an RGB encoder fused via attention, followed by temporal encoding and LLM-based pretraining for ISLR and CSLR.

        \subsubsection{\textbf{Human-Computer Interaction (HCI)}}
        
        Motivated by the difficulty of recognizing fine-grained gestures in HCI applications, authors in \citep{HandSense} proposed HandSense, a method relying on depth and RGB data. Two 3DCNNs are deployed to extract features for each modality. These features are fused and fed into an SVM to obtain the final prediction. Authors in \citep{mXception_and_RDDI} proposed an approach relying on RGBD Dynamic Images (RDDIs). These images are obtained by combining RGB and estimated depth data. The resulting RDDI is fed into a modified Xception, resulting from removing the top four Xception modules, in order to extract meaningful features. These features are finally used to perform the classification. 

        \subsubsection{\textbf{Other}}

        Inspired by the success of transformers, GestFormer \citep{GestFormer} introduces a lightweight transformer-based framework for gesture-driven HRI. A 2DCNN extracts per-frame features, processed by PoolFormer layers (replacing attention) and two custom blocks for multiscale pooling/WT and token mixing. Each modality is processed independently, and outputs are averaged. In \citep{DHGR_3D_POSE_HRI}, a hybrid 3DCNN/LSTM model fuses RGB, depth, and skeleton data (from an enhanced pose estimator) via weighted sum. The fused input feeds a spatiotemporal model combining 2D/3D convolutions and ConvLSTM layers. Authors in \citep{MDSI-fang2025mdsi} emphasize two major issues, information redundancy and the difficulty of recognizing similar gestures. Their approach leverages a CLIP encoder to provide qualitative embeddings and enhance the feature extractor through label smoothing as well as a semantic filter to incorporate multimodal data.

    \subsection{Comparison and Critical Summary of IDHGR Methods}
    
        \begin{table}[t]
            \caption{Isolated Dynamic VHGR: comparison of appearance-based, pose-based, and multimodal method families.}
            \label{tab:idvhgr_modality_summary}
            \centering
            \tiny
            \setlength{\tabcolsep}{2pt}
            \renewcommand{\arraystretch}{0.9}
        
            \newcommand{\caspect}{1.7cm}
            \newcommand{\capp}{3.5cm}
            \newcommand{\cpose}{3.5cm}
            \newcommand{\cmult}{3.5cm}
            
            \begin{tabular}{
            m{\caspect}
            m{\capp}
            m{\cpose}
            m{\cmult}
            }
            \toprule
            \textbf{Aspect} & 
            \textbf{Unimodal appearance-based methods} & 
            \textbf{Unimodal pose-based methods} & 
            \textbf{Multimodal methods} \\
            \midrule
            
            Primary function & 
                Direct spatiotemporal feature extraction from raw pixel intensities
            & 
                Geometric modeling of skeletal joint relationships and trajectories
            & 
                Integration of hierarchical features from cross-modal data streams
            \\
            \midrule
        
            Most suitable scenarios &
            Trimmed RGB/depth videos where hand shape, texture, and scene context are informative and the computational budget allows video backbones
            &
            Privacy-sensitive, low-light, or background-cluttered settings where reliable skeletons are available and geometric motion is sufficient
            &
            High-accuracy ISLR or RGB-D HCI/HRI scenarios where complementary cues can justify the added sensing, synchronization, and fusion cost
            \\
        
            \midrule
            
            Input data types & 
            \begin{itemize}[leftmargin=*,nosep]
                \item RGB
                \item Depth
                \item Infrared (IR)
                \item Optical Flow (OF)
                \item Motion history
            \end{itemize} & 
            \begin{itemize}[leftmargin=*,nosep]
                \item 2D/3D skeletal joints
                \item Bones
                \item Joint angles
                \item Motion vectors
            \end{itemize} & 
            \begin{itemize}[leftmargin=*,nosep]
                \item RGB + Skeleton
                \item Depth + Pose
                \item Visual + textual/semantic guidance
            \end{itemize} \\
            \midrule
            
            Architectures & 
            \begin{itemize}[leftmargin=*,nosep]
                \item 3DCNN
                \item (2+1)D CNN
                \item ConvLSTM
                \item 2DCNN+RNN
                \item Video Transformers
            \end{itemize} & 
            \begin{itemize}[leftmargin=*,nosep]
                \item Graph Convolutional Networks (GCNs)
                \item Temporal Convolutional Networks (TCNs)
                \item Transformers
            \end{itemize} & 
            \begin{itemize}[leftmargin=*,nosep]
                \item Multi-stream ensembles
                \item Cross-attention fusion modules
                \item Hybrid GCN-CNNs
            \end{itemize} \\
            \midrule
            
            Learning strategy & 
            \begin{itemize}[leftmargin=*,nosep]
                \item Transfer learning
                \item Contrastive learning
                \item Supervised pretraining
            \end{itemize} & 
            \begin{itemize}[leftmargin=*,nosep]
                \item Masked reconstruction
                \item Self-Supervised Learning (SSL)
                \item Adversarial data augmentation
            \end{itemize} & 
            \begin{itemize}[leftmargin=*,nosep]
                \item Knowledge distillation
                \item Multi-task learning
                \item Zero-shot/Few-shot learning
            \end{itemize} \\
            \midrule
            
            Strengths & 
            \begin{itemize}[leftmargin=*,nosep]
                \item Capturing of fine-grained hand shapes, texture and environmental context
            \end{itemize} & 
            \begin{itemize}[leftmargin=*,nosep]
                \item Computationally efficient
                \item Privacy-preserving
                \item Lighting-invariant
            \end{itemize} & 
            \begin{itemize}[leftmargin=*,nosep]
                \item Highest accuracy
                \item Robust to occlusion
                \item Handling of visually similar gestures
            \end{itemize} \\
            \midrule
            
            Limitations & 
            \begin{itemize}[leftmargin=*,nosep]
                \item High computational cost
                \item Sensitive to background clutter and lighting
            \end{itemize} & 
            \begin{itemize}[leftmargin=*,nosep]
                \item Dependency on pose estimator accuracy
                \item Lack of textural detail
            \end{itemize} & 
            \begin{itemize}[leftmargin=*,nosep]
                \item Extreme complexity
                \item Large parameter footprint
                \item Difficulty in synchronization
            \end{itemize} \\
            \midrule
            
            Indicative methods & 
                (2+1)D-SLR \citep{2plus1-D-SLR_and_NCSL}, TS3C-Net \citep{TS3C-Net}, EgoFormer \citep{EgoFormer_METHOD_EgoDriving_DATASET}, TIM-SLR \citep{TIM-SLR}, \citep{SlowFast-Transformer}
            & 
                TD-GCN \citep{TD-GCN}, STST \citep{STST}, DSTSA-GCN \citep{DSTSA-CGN_cui2025dstsa}
            & 
                SAM-SLR \citep{SAM_SLR}, Uni-Sign \citep{UNISIGNli2025}, NLA-SLR \citep{NLA_SLR}, \citep{Yunan_Li}
            \\
            \bottomrule
            \end{tabular}%
        \end{table}

        Table~\ref{tab:idvhgr_modality_summary} provides the qualitative comparison of the main IDHGR method families, while Table~\ref{tab:idhgr_quantitative} reports their quantitative performance on the most used benchmarks. The performance of IDHGR methods in the most used benchmarks (IsoGD \citep{chalearn-lap}, SHREC \citep{SHREC}, DHG \citep{DHG-CVPRW}, MSASL \citep{MS-ASL}, WLASL \citep{WLASL}, CSL500 \citep{CSL500} and AUTSL \citep{AUTSL}) is shown in Table \ref{tab:idhgr_quantitative}. Since the results are collected from the original publications rather than reproduced under a unified experimental setup, they should be interpreted cautiously. Direct comparison is most meaningful within the same benchmark dataset and reported protocol; across datasets, modalities, and preprocessing pipelines, differences in data difficulty, pose-estimation overhead, additional training data, and fusion strategies may affect the reported performance. For ISLR, Uni-Sign \citep{UNISIGNli2025} surpasses relevant methods in terms of recognition accuracy, but requires powerful hardware to be executed fast, due to its large size (592.1M parameters, most of them contained in the language model). Thus, smaller but less accurate models like NLA-SLR \citep{NLA_SLR}, SignBERT+ \citep{SignBERT_ICCV} and \citep{Weichao-Zhao} are more reasonable solutions for real-world applications. Even though the multimodal SAM-SLR \citep{SAM_SLR} and SAM-SLR-v2 \citep{jiang2021sign-sam-slr-v2} are very accurate models for ISLR, they require a large number of modalities that are not fused effectively (using late fusion), which may be suboptimal. Pose-based methods for HCI and HRI including TD-GCN \citep{TD-GCN}, STST \citep{STST} and DSTSA-GCN \citep{DSTSA-CGN_cui2025dstsa} achieve high recognition rates with relatively low computational cost. Although, the computational complexity of pose estimation and the effect of its incorrect predictions in more challenging environments (e.g. fast movements) may reduce their overall performance. Appearance-based methods are not affected by off-the-shelf pose estimators but they comprise many more parameters (up to 251.4M \citep{3D_CDC_NAS_TIP_2021}).

        Overall, the main insights arising from Tables~\ref{tab:idvhgr_modality_summary} and~\ref{tab:idhgr_quantitative}, resolving Q3 (see Section~\ref{sec:intro}), are the following (note that most of the purely appearance-based methods are evaluated on either custom or lesser-known datasets, so only a few of them are included in the table):
        \begin{itemize}[leftmargin=*]
            \item Keyframe selection \cite{Tianyu_Liu,KEYFRAMES_DRL_SKELETON} does not appear to introduce additional errors due to failures, unlike pose estimators, which may degrade accuracy. However, only a few works have leveraged its potential to reduce computational cost, probably due to the complexity of training this additional component.
            \item A few works consider human body regions and process each of them independently \cite{Tianyu_Liu,FDMSE-hierarchicalwindowedgraphattention,ISLR-STGCN-attention-LSTM-ozdemir2025cross,HAN-liu2025han,MSNN,UNISIGNli2025,Weichao-Zhao}. However, it remains unclear whether the independent processing of different human regions improves accuracy, as observed in some works \cite{Weichao-Zhao,UNISIGNli2025}, or not, as reported in others \cite{MSNN}.
            \item NAS is deployed in older approaches \cite{RAAR3DNet,3D_CDC_NAS_TIP_2021}, surpassing their contemporary counterparts at the time, around 2021. However, although such methods explore a wide spectrum of possible architectures, they use older backbones, such as I3D \citep{RAAR3DNet} instead of more advanced models like S3D \citep{NLA_SLR}, or limited training strategies, such as \citep{3D_CDC_NAS_TIP_2021}. This leads to inferior accuracy compared to more recent methods.
            \item Unimodal pose-based approaches still incorporate derived skeleton modalities, such as motion and bones \cite{SKIM,DSTA_SLR_hu2024dynamic,JOINTS_MOTION_VECTORS_Improved_ST-GCN}, without any significant impact on recognition accuracy.
            \item Having been thoroughly explored, the ST-GCN scheme dominates pose-based IDHGR \cite{JOINTS_MOTION_VECTORS_Improved_ST-GCN,SKIM} and achieves superior performance \cite{DSTSA-CGN_cui2025dstsa,TD-GCN}. However, spatiotemporal transformers are emerging as competitive alternatives in terms of accuracy \citep{STST}.
            \item Researchers seldom rely solely on temporal modeling  \cite{SwC-GR-MMixer}, aiming instead to fully exploit the spatial domain.
            \item In general, pure transformers have not yet been applied for appearance-based feature extraction. Instead, only 3DCNN backbones, either custom models or models originally proposed for action recognition, have been used to extract features from related modalities  \cite{NLA_SLR,SeST,MSNN,jiang2021sign-sam-slr-v2,SAM_SLR}.
            \item Incorporating multiple modalities and cues, such as face and body regions, under a multi-task, high-capacity framework significantly improves accuracy but increases computational cost \cite{UNISIGNli2025}.
            \item Numerous approaches have achieved more than 90\% accuracy on datasets with small corpora, such as SHREC \citep{SHREC} and DHG \citep{DHG-CVPRW}. More complicated datasets, especially for ISLR, that comprise large vocabularies and only a few samples per class remain highly challenging, such as MSASL \citep{MS-ASL} and WLASL \citep{WLASL}.
        \end{itemize}

    \begin{sidewaystable}[]
        
        \newcommand{\cspatiotemporal}{SkyBlue}
        \newcommand{\cspatiotemporaltemporal}{GreenYellow}
        \newcommand{\cspatialandtemporal}{Orchid}
        \newcommand{\ctemporal}{Orange}
        \newcommand{\cother}{Gray}
        
        \centering
        \tiny
        \caption{Performance of IDHGR methods on various datasets. Notably, accuracy depends on dataset difficulty; for example, methods with accuracy near 100\% on AUTSL do not exhibit similar performance on more challenging datasets with thousands of classes, such as MSASL and WLASL. Top-3 accuracies are shown in bold. KS, R and NAS stand for keyframe selection, regions and Neural Architecture Search, respectively.} Colors correspond to architecture type: \cback{\cspatiotemporal}{Spatiotemporal}, \cback{\cspatiotemporaltemporal}{(Short-term) Spatiotemporal + (Long-term) Temporal}, \cback{\cspatialandtemporal}{Spatial + Temporal}, \cback{\ctemporal}{Temporal} and \cback{\cother}{other}. Methods marked with "\dag" have used additional training data. Results are reported as in the original publications; therefore, comparisons should be made primarily within the same benchmark dataset and reported protocol.
        \setlength{\tabcolsep}{1pt}
        \renewcommand{\arraystretch}{1}

        \begin{tabular}{ll|cccll|llllllllllll}
        \toprule
        & & & & & & & & & \multicolumn{2}{l}{SHREC} & \multicolumn{2}{l}{DHG} & \multicolumn{2}{l}{MSASL1000} & \multicolumn{2}{l}{WLASL2000} & & \\
        \multirow{-2}{*}{Method}&\multirow{-2}{*}{Year} & \multirow{-2}{*}{KS} & \multirow{-2}{*}{R} & \multirow{-2}{*}{NAS} & \multirow{-2}{*}{Fusion} & \multirow{-2}{*}{Backbone} & \multirow{-2}{*}{Params} & \multirow{-2}{*}{IsoGD} & 14 & 28 & 14 & 28 & 
        PI & PC & PI & PC &\multirow{-2}{*}{CSL500} & \multirow{-2}{*}{AUTSL} \\
        
        \midrule
        \multicolumn{19}{c}{Unimodal appearance-based} \\
        \midrule
        
        \cback{\cspatiotemporal}{SeST \citep{SeST}} &2021& \xmark & \xmark & \xmark & Mid-level & ResC3D \& ConvLSTM &-&60.27\%&-&-&-&-&-&-&-&-&-&-\\ 
        
        \cback{\cother}{\citep{Tianyu_Liu}}&2024& \cmark & \cmark & \xmark & Mid-level & Other & -&-&-&-&-&-&-&-&-&-&\textbf{98.87\%}&92.53\%\\
        
        \midrule
        \multicolumn{19}{c}{Unimodal pose-based} \\
        \midrule
        
        \cback{\cspatiotemporal}{STGCN-LSTM \citep{STGCN-LSTM-10965662}}&2025& \xmark & \xmark & \xmark & Mid-level & ST-GCN \& ConvLSTM &-&-&-&-&-&-&-&-&-&-&95.2\%&-\\ 
        \cback{\cspatiotemporal}{DSTSA-GCN  \citep{DSTSA-CGN_cui2025dstsa}}&2025& \xmark & \xmark & \xmark & - & GCN/TCN &8.08M&-&\textbf{97.74\%}&\textbf{95.37\%}&\textbf{95.04\%}&\textbf{93.57\%}&-&-&-&-&-&-\\ 
        \cback{\cspatiotemporal}{TMS-Net \citep{TMS-NET}}&2024& \xmark & \xmark & \xmark & Multi-level & GCN/TCN &-&-&-&-&-&-&-&-&56.4\%&-&-&-\\ 
        \cback{\cspatiotemporal}{SKIM \citep{SKIM}}&2024& \xmark & \xmark & \xmark & Late & GCN/TCN &-&-&-&-&-&-&-&-&57.73\%&55.40\%&-&-\\ %
        \cback{\cspatiotemporal}{DSTA-SLR \citep{DSTA_SLR_hu2024dynamic}}&2024& \xmark & \xmark & \xmark & Late & GCN/TCN &7.4M&-&-&-&-&-&65.74\%&62.31\%&53.68\%&51.17\%&98.15\%&-\\ %
        \cback{\cspatiotemporal}{STST \citep{STST}}&2024& \xmark & \xmark & \xmark & - & Transformer &-&-&\textbf{97.62\%}&\textbf{95.83\%}&\textbf{94.82\%}&\textbf{93.18\%}&-&-&-&-&-&-\\ 
        \cback{\cspatiotemporal}{HWGAT \citep{FDMSE-hierarchicalwindowedgraphattention}}&2024& \xmark & \cmark & \xmark & - & Transformer \& GCN &-&-&-&-&-&-&-&-&48.49\%&-&-&\textbf{95.80\%}\\ 
        \cback{\cspatiotemporal}{\citep{omar_ikne}}&2024& \xmark & \xmark & \xmark & - & MAE+ST-GCN &-&-&94.1\%&90.0\%&-&-&-&-&-&-&-&-\\ 
        \cback{\cspatiotemporal}{TD-GCN \citep{TD-GCN}}&2024& \xmark & \xmark & \xmark & Late & GCN/TCN &1.36M&-&\textbf{97.02\%}&\textbf{95.36\%}&93.9\%&91.4\%&-&-&-&-&-&-\\ 
        \cback{\cspatiotemporal}{SL-TSSI-DenseNet \citep{L-TSSI-DenseNet}} &2023& \xmark & \xmark & \xmark & - & TSSI+DenseNet &7.2M&-&-&-&-&-&-&-&-&-&-&93.13\%\\ %
        \cback{\cspatiotemporal}{3sISTGCN \citep{JOINTS_MOTION_VECTORS_Improved_ST-GCN}} &2022& \xmark & \xmark & \xmark & Late & Modified ST-GCN &7.1M&-&96.7\%&94.9\%&93.7\%&91.2\%&-&-&-&-&-&-\\ 
        
        \cback{\cspatiotemporaltemporal}{\citep{ISLR-STGCN-attention-LSTM-ozdemir2025cross}}&2025& \xmark & \cmark & \xmark & Mid-level & ST-GCN+Attention+LSTM &-&-&-&-&-&-&-&-&-&-&-&86.43\%\\ 

        \cback{\cspatialandtemporal}{SBI-DHGR \citep{SBI-DHGR}}&2023& \xmark & \xmark & \xmark & Mid-level & Multistream 1DCNN+LSTM &28M&-&97\%&92.36\%&\textbf{94.64\%}&91.79\%&-&-&-&-&-&-\\ 
        \cback{\cspatialandtemporal}{STr-GCN \citep{STr-GCN}}&2023& \xmark & \xmark & \xmark & - & GCN+Transformer &-&-&93.39\%&89.20\%&-&-&-&-&-&-&-&-\\ 
        
        \cback{\ctemporal}{SwC GR-MMixer \citep{SwC-GR-MMixer}}&2024& \xmark & \xmark & \xmark & - & GRU &-&-&-&-&-&-&-&-&-&-&\textbf{98.54\%}&-\\ 
                
        \cback{\cother}{HAN-2S \citep{HAN-liu2025han}} &2025& \xmark & \cmark & \xmark & Multi-level & Transformer &0.94M&-&95\%&92.86\%&92.71\%&89.15\%&-&-&-&-&-&-\\ 
        \cback{\cother}{\citep{Yangke_patcog}}&2024& \xmark & \xmark & \xmark & Late & AE \& Transformer &0.21M&-&96.9\%&94.17\%&94.21\%&\textbf{92.11\%}&-&-&-&-&-&-\\ 
        \cback{\cother}{\citep{Musa_IEEE_ACCESS}}&2023& \xmark & \xmark & \xmark & Mid-level & GCN \& MLP &-&-&97.01\%&92.78\%&92.00\%&88.78\%&-&-&-&-&-&-\\ 
        \cback{\cother}{DRL-DDNet \citep{KEYFRAMES_DRL_SKELETON}}&2023& \cmark & \xmark & \xmark & - & Other &-&-&96.5\%&93.8\%&-&-&-&-&-&-&-&-\\ 
        
        \midrule
        \multicolumn{19}{c}{Multimodal} \\
        \midrule
        
        \cback{\cspatiotemporal}{MSNN \citep{MSNN}}&2024& \xmark & \cmark & \xmark & Late & I3D \& ST-GCN streams &-&-&-&-&-&-&50.59\%&-&47.26\%&-&-&-\\ 
        \cback{\cspatiotemporal}{NLA-SLR \citep{NLA_SLR}}&2023& \xmark & \xmark & \xmark & Multi-level & Multi-stream S3D variant &-&-&-&-&-&-&\textbf{73.80\%}&\textbf{70.95\%}&\textbf{61.26\%}&\textbf{58.31\%}&-&-\\ 
        \cback{\cspatiotemporal}{\citep{Yunan_Li}}&2023& \xmark & \xmark & \xmark & Late & Multi-stream Res3D &-&\textbf{74.08\%}&-&-&-&-&-&-&-&-&-&-\\ 
        \cback{\cspatiotemporal}{CTFB \citep{CTFB_Hampiholi}}&2023& \xmark & \xmark & \xmark & Multi-level & Dual-stream I3D &-&\textbf{69.02\%}&-&-&-&-&-&-&-&-&-&-\\ 
        \cback{\cspatiotemporal}{SlowFast+CDC \citep{3D_CDC_NAS_TIP_2021}} &2021& \xmark & \xmark & \cmark & Multi-level & Multi-stream SlowFast &251.4M&66.23\%&-&-&-&-&-&-&-&-&-&-\\ 
        \cback{\cspatiotemporal}{\citep{Hu_Zhou_Li_2021}}&2021& \xmark & \xmark & \xmark & Late & ST-GCN \& Res3D &-&-&-&-&-&-&69.39\%&66.54\%&51.39\%&48.75\%&98.3\%&-\\ 
        \cback{\cspatiotemporal}{RAAR3DNet \citep{RAAR3DNet}}&2021& \xmark & \xmark & \cmark & Late & I3D with attention &-&66.62\%&-&-&-&-&-&-&-&-&-&-\\ 
        \cback{\cspatiotemporal}{SAM-SLR-v2 \citep{jiang2021sign-sam-slr-v2}}&2021& \xmark & \xmark & \xmark & Late & 3DCNN \& GCN/TCN &-&-&-&-&-&-&-&-&\textbf{59.39\%}&\textbf{56.63\%}&\textbf{99.00\%}&\textbf{98.53\%}\\ 
        \cback{\cspatiotemporal}{SAM-SLR \citep{SAM_SLR}}&2021& \xmark & \xmark & \xmark & Late & 3DCNN \& GCN/TCN &19.2M&-&-&-&-&-&-&-&58.73\%&55.93\%&-&\textbf{98.53\%}\\ 
        
        \cback{\cother}{Uni-Sign \citep{UNISIGNli2025}}\dag &2025& \xmark & \cmark & \xmark & Mid-level & Various encoders + LLM &592.1M&-&-&-&-&-&\textbf{78.16\%}&\textbf{76.97\%}&\textbf{63.52\%}&\textbf{61.32\%}&-&-\\ %
        \cback{\cother}{MASA \citep{MASA}}&2024& \xmark & \xmark & \xmark & Late & GCN+Transformer \& I3D &-&-&-&-&-&-&-&-&55.77\%&53.13\%&-&-\\ 
        \cback{\cother}{\citep{Weichao-Zhao}}&2024& \xmark & \cmark & \xmark & Late & GCN+Transformer \& I3D &-&-&-&-&-&-&\textbf{74.18\%}&\textbf{72.18\%}&58.06\%&55.66\%&97.8\%&-\\ 
        \cback{\cother}{BEST \citep{BEST}}&2023& \xmark & \xmark & \xmark & Late & GCN+Transformer \& I3D &-&-&-&-&-&-&71.21\%&68.24\%&54.59\%&52.12\%&97.7\%&-\\ 
        \cback{\cother}{SignBERT+ \citep{SignBERT+}}&2023& \xmark & \xmark & \xmark & Late & GCN+Transformer \& I3D &-&-&-&-&-&-&73.71\%&70.77\%&55.59\%&53.33\%&97.8\%&-\\ 
        \cback{\cother}{De+Recouple \citep{Benjia_Zhou}}&2022& \xmark & \xmark & \xmark & Multi-level & Other &-&\textbf{66.79\%}&-&-&-&-&-&-&-&-&-\\ 
        \cback{\cother}{SignBERT \citep{SignBERT_ICCV}}&2021& \xmark & \xmark & \xmark & Late & GCN+Transformer \& I3D &-&-&-&-&-&-&71.24\%&67.96\%&54.69\%&52.08\%&97.6\%&-\\ 
        
        \bottomrule
        \end{tabular}
        \label{tab:idhgr_quantitative}
    \end{sidewaystable}

\section{Methods for Continuous Dynamic VHGR}
\label{sec:continuous}

    In contrast to IDHGR, where each sample contains at most one gesture, Continuous Dynamic Hand Gesture Recognition (CDHGR) deals with unsegmented gesture sequences. The goal is to recognize gestures from untrimmed videos and generate the corresponding gesture sequence. CDHGR is typically treated as a \textbf{weakly supervised task}, as the precise temporal boundaries of gestures are not provided; the model learns to identify them implicitly, often using Connectionist Temporal Classification (CTC) loss \citep{Fangyun_Wei_ICCV, CTC_graves2006connectionist}.
    
    While CDHGR has been explored in domains like autonomous driving \citep{FA-STGCN, CTCX_WANG2022123}, most research focuses on Sign Language Understanding (SLU), where it is commonly referred to as \textbf{Continuous Sign Language Recognition (CSLR)} or sign2gloss. CSLR aims to produce a sequence of glosses\footnote{Glosses are atomic lexical units in sign language, analogous to words in a spoken language.} from video input. Due to the importance of non-manual features (e.g., facial expressions and upper-body movements), many approaches also incorporate these regions to improve recognition accuracy. Some CSLR methods are further extended to support gloss-based Sign Language Translation (SLT) by adding a gloss-to-text module \citep{MSKA_ELSEVIER, zuo2024onlinecontinuoussignlanguage, NEURIPS2022_6cd3ac24_HEATMAPS}.
    
\begin{figure}[t]
        \centering
        \small
        \setlength{\tabcolsep}{5pt}
        \renewcommand{\arraystretch}{5}
        \newcommand{\width}{0.45}
        \begin{tabular}{cc}
            \begin{subfigure}[t]{\width\textwidth}
                \centering
                \includegraphics[width=\linewidth]{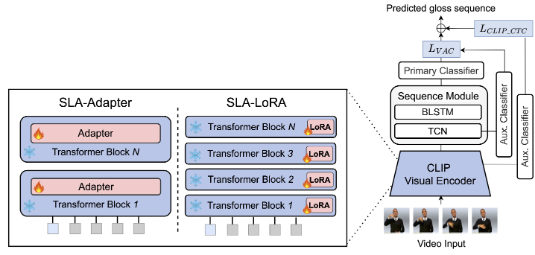}
                \caption{}
            \end{subfigure} &
            \begin{subfigure}[t]{\width\textwidth}
                \centering
                \includegraphics[width=\linewidth]{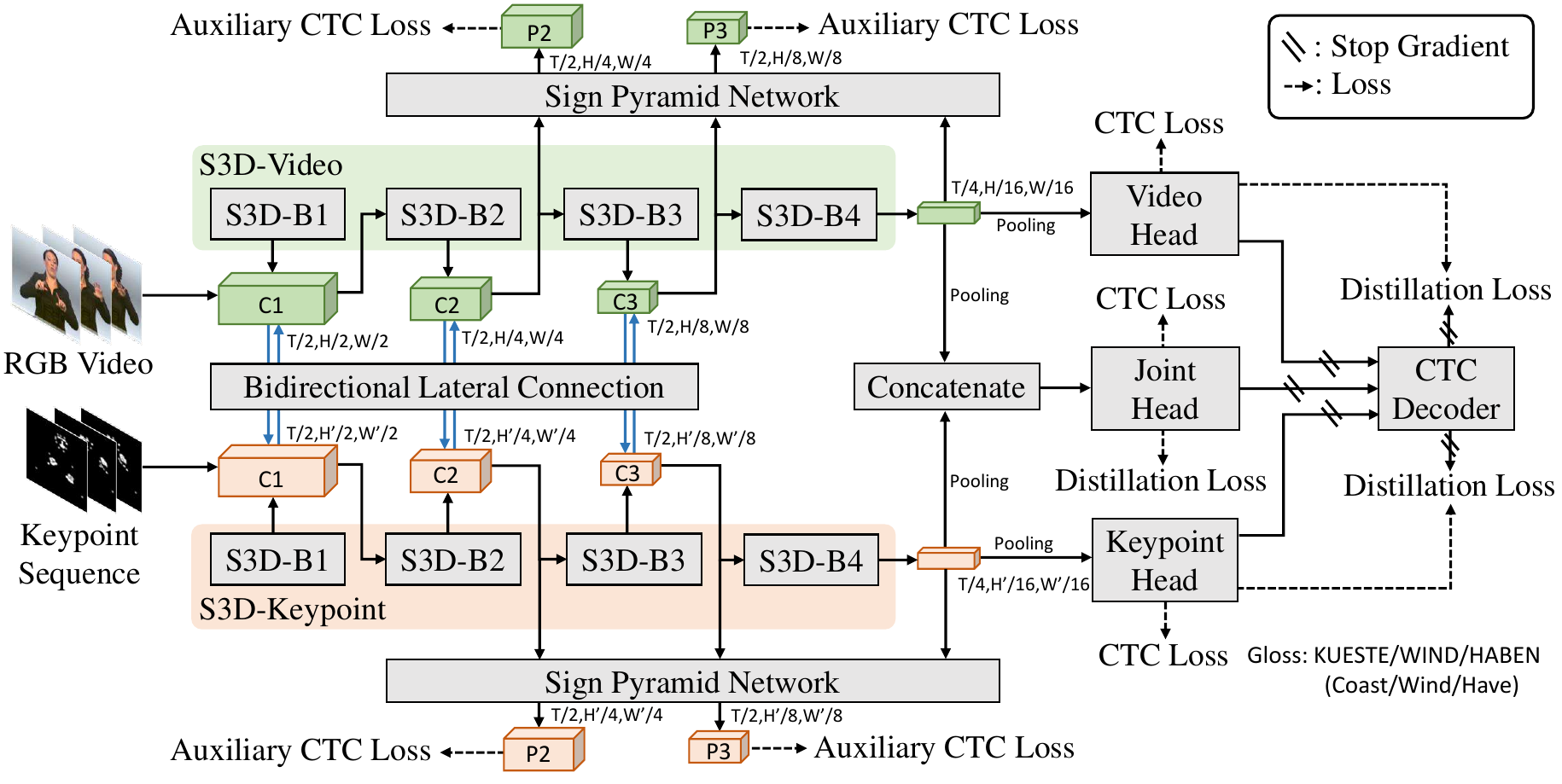}
                \caption{}
            \end{subfigure} \\
            \begin{subfigure}[t]{\width\textwidth}
                \centering
                \includegraphics[width=\linewidth]{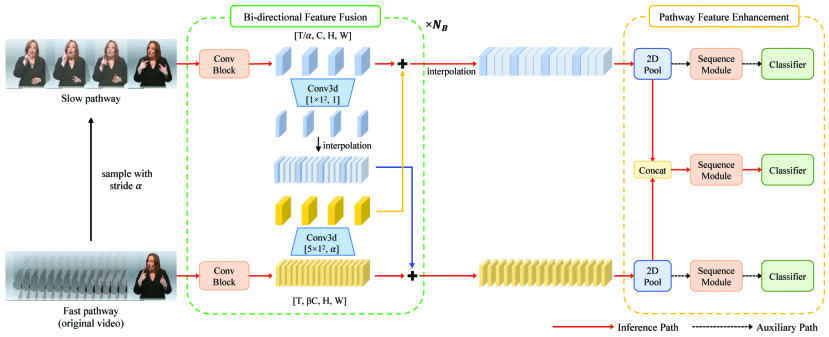}
                \caption{}
            \end{subfigure} &
            \begin{subfigure}[t]{\width\textwidth}
                \centering
                \includegraphics[width=\linewidth]{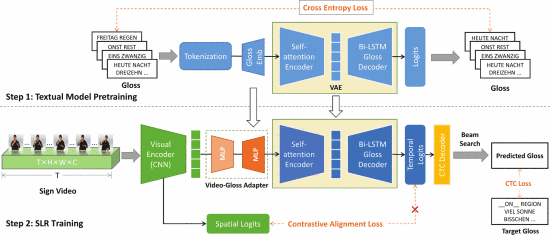}
                \caption{}
            \end{subfigure} \\
        \end{tabular}
        \caption{Representative network structures and algorithmic pipelines for CSLR: CLIP-SLA \cite{CLIP_SLA_alyami2025clip_CSLR_2025} (a) enhances spatial perception using a foundation model, TwoStream-SLR \cite{NEURIPS2022_6cd3ac24_HEATMAPS} (b) proposes an end-to-end 3DCNN for multimodal CSLR with an advanced supervision scheme, SlowFastSign \cite{SlowFastSign} (c) considers various frame rates to increase performance, and CVT-SLR \cite{CVT-SLR} (d) uses a generative model for producing glosses instead of the standard decoder.}
        \label{fig:composite_cslr}
    \end{figure}

    In the following, we review recent advances in CDHGR, grouped by input modality. We highlight key architectures, learning strategies, and benchmark results as required by Q2 and Q3 (see Section~\ref{sec:intro}). Representative network structures and algorithmic pipelines for CSLR are illustrated in Fig.~\ref{fig:composite_cslr}.
    
    \subsection{RGB-based}

        Most SOTA RGB-based CDHGR models follow a \textbf{three-stage pipeline} architecture: (1) a spatial model (commonly a 2DCNN or a transformer) to extract frame-wise features; (2) a short-term temporal model (mainly 1DCNN or TCN) for local temporal modeling and gloss-level feature extraction; and (3) a contextual module (e.g., BiLSTM) for long-term sequence modeling and final gesture sequence prediction \citep{TemporalLiftPooling, VAC-CSLR, SMKD, CorrNet, C2ST, HezhenHu_TMM, SEN-CSLR, HezhenHu-JunfuPu, corrnetplus_hu2024corrnetsignlanguagerecognition}. This standard architecture was introduced in \citep{first_2dcnn_tcn_bilstm_TMM_2019}. This decoupled spatial-temporal structure has been favored over unified spatiotemporal models like 3DCNNs due to the latter's inability to define clear temporal boundaries \citep{ADADOGLOU}.
        
        Many RGB‐based CDHGR methods build on the canonical 2DCNN+1DCNN+BiLSTM architecture by introducing auxiliary losses and multi‐task objectives. For example, VAC‐CSLR \citep{VAC-CSLR} augments the visual and temporal modules with a Visual Enhancement (VE) loss and a Visual Alignment (VA) distillation loss to accelerate convergence and improve representation quality. Similarly, SMKD \citep{SMKD} adds two auxiliary classifiers-one after the 1DCNN and one after the BiLSTM-to perform synchronous gloss segmentation alongside recognition, using a three‐stage training schedule to mitigate gradient "spikes". C\textsuperscript{2}SLR \citep{C2SLR} enforces Spatial Attention Consistency (SAC) by focusing the visual module on pose‐informed regions, and Sentence Embedding Consistency (SEC) to align the visual and sequential representations of the same gloss.
    
        To better capture local temporal patterns, several works replace or enhance the standard 1DCNN. Temporal Lift Pooling (TLP)~\citep{TemporalLiftPooling} adopts a learned pooling scheme inspired by the lifting transform, eliminating handcrafted kernels. mLTSF‐Net~\citep{mLTSF-Net} couples content‐aware feature selection with position‐aware convolutions to fuse adjacent‐frame similarities at multiple scales. HS‐I3D~\citep{HS-I3D} extends I3D by extracting intermediate embeddings and deploying a U‐Net-style decoder that upsamples temporally, enabling simultaneous sign detection and boundary localization. 
        
        Other methods inject lightweight attention and correlation blocks into the spatial encoder. CorrNet~\citep{CorrNet} adds parallel correlation and identification modules after each 2DCNN stage to model inter‐frame trajectories and highlight informative regions; CorrNet+~\citep{corrnetplus_hu2024corrnetsignlanguagerecognition} further replaces its correlation operator with a more efficient variant and inserts an attention layer. SEN‐CSLR~\citep{SEN-CSLR} introduces self‐emphasizing spatial (SSEN) and temporal (TSEN) modules to reweight feature maps and keyframes with minimal overhead. STNet \citep{STNet-wang2025continuous} applied frame-wise loss to the 2DCNN backbone output to enhance feature extraction and inserted multiple Spatial Resonance Modules to facilitate inter-frame modeling.
    
        Many recent approaches replace 2DCNNs with transformer-based or hybrid CNN/attention modules. CLIP-SLA variants~\citep{CLIP_SLA_alyami2025clip_CSLR_2025} use a frozen CLIP visual encoder with LoRA or MLP adapters and an auxiliary classifier for early supervision. Swin-MSTP \citep{SWIN_MSTP_Neurocomputing_2025_CSLR} combines a Swin transformer with an MS-TCN using TLP and dilated convolutions, followed by a BiLSTM for global context.
    
        TCNet \citep{TCNet_Lu_Salah_Poppe_2024} augments a CNN backbone with trajectory and correlation modules, while SlowFastSign \citep{SlowFastSign} modifies Slow-Fast-101 with bidirectional fusion and auxiliary temporal classifiers. To better model spatial correlations, SignGraph \citep{SignGraph_CVPR} treats frame patches as nodes in a dynamic graph, alternating between Local Sign Graph (LSG) and Temporal Sign Graph (TSG) modules; its multiscale variant, MultiSignGraph, further improves spatiotemporal context.
    
        CORE \citep{CORE-CSLR-zhang2025core} proposed a different approach aiming to leverage the relations between regions within an input frame, deploying an off-the-shelf pose estimator to detect fine-grained and contextual cues; the corresponding frame regions are fed into a 2DCNN and the extracted features are fused using a GNN to obtain the final embedding. These per-frame representations pass through a BiLSTM to perform global modeling. SD-Net \citep{SD-Net-CSLR-gao2025structure} enhanced temporal modeling by using three parallel branches. Each branch consists of a two-stage temporal context module that captures both frame-wise and gloss-wise features. Contrastive loss is applied between two of the branches, based on the proposed Entity and Template scheme. Authors in \citep{diffusion-CSLR-geng2026no} formulated CSLR as a cross-modal alignment task using a diffusion model to generate the glosses aiming to mitigate the weak supervision, achieving superior performance. DSTEN-CSLR \citep{DSTEN-CSLR-yu2025dsten} replaced the BiLSTM with a custom contextual model, consisting of memory and feed-forward blocks. Furthermore, the 2DCNN backbone was enhanced by adding a Spatial-Channel Attention (SCA) mechanism, supervised by ground-truth heatmaps, provided by an off-the-shelf pose estimator.
    
        Some models integrate linguistic embeddings with visual features to leverage gloss co-occurrence and language structure. SignBERT-HKSL \citep{SignBERT-HKSL} combines a (3+2+1)D ResNet with BERT and BiLSTM, pretraining on masked inputs and fine-tuning iteratively on masked outputs. C\textsuperscript{2}ST \citep{C2ST} replaces the CTC decoder with a Conditional Gloss Decoder (CGD) to model gloss dependencies and aligns visual and gloss embeddings via cross-modal match loss. GPGN \citep{GPGN} similarly applies a cross-modal match loss between BERT-derived gloss features and PDC-TFE visual embeddings, followed by a BiGRU for sequence modeling.
    
        CVT-SLR \citep{CVT-SLR} pretrains an asymmetric VAE on a pseudo gloss-to-gloss task, then connects it to a visual encoder with a video-gloss adapter and contrastive alignment loss. Fully transformer-based PiSLTRc \citep{PiSLTRc} uses content-aware neighborhood gathering and position-informed attention for end-to-end CSLR, while \citep{HezhenHu-JunfuPu} introduces prior-aware cross-modality augmentation, training a transformer to reconstruct masked gloss sequences and generate pseudo video-text pairs via guided gloss substitutions and insertions.
    
        Finally, data augmentation and cross‐lingual strategies help alleviate limited annotations. Fangyun Wei et al.~\citep{Fangyun_Wei_ICCV} build a “dictionary” by using a pretrained CDHGR model to automatically segment continuous videos into isolated signs, then mapping German and Chinese sign corpora into a shared embedding space. CSGC~\citep{CSGC_Rao_Sun_Wang_Wang_Zhang_2024} maintains a Gloss Prototype memory that stores class prototypes updated during training; a gloss contrastive loss then enforces cross‐sentence consistency.

    \subsection{Pose-based}

        Pose-based (or skeleton-based) approaches for CDHGR rely exclusively on body pose information, typically in the form of 2D/3D skeleton joints or pose heatmaps. These methods aim to model human motion while discarding appearance-based noise. However, due to the rapid motions (especially in the CSLR context) pose estimators often fail, making solely pose-based methods frequently inaccurate.
        
        Several works exploit structured keypoint information by dividing the body into semantic regions such as the face, hands, and upper body. MSKA-SLR~\citep{MSKA_ELSEVIER} processes each region independently via attention-enhanced temporal convolutions across four branches. The resulting features are fused to decode gloss sequences with CTC loss supervision, complemented by frame-level self-distillation. Similarly, CoSign~\citep{CoSign} employs group-specific GCNs for each subset and introduces regularization to preserve inter-stream consistency. Their dual-stream variant integrates joint positions and motion features using multi-level fusion to capture both static and dynamic cues.
        
        Temporal localization of gestures has also been explored in the literature. TY-Net~\citep{TY-Net} draws inspiration from YOLO to detect gesture instances in time by predicting their boundaries and centers, analogous to spatial bounding boxes.
        
        Beyond joint coordinates, pose heatmaps serve as an alternative input modality. The multi-cue framework in~\citep{MULTI-CUE} combines spatial and temporal reasoning by localizing facial and hand regions via a pose estimator and subsequently encoding them with a temporal module, before decoding gloss sequences with a CTC head. The FA-STGCN model~\citep{FA-STGCN} further improves robustness to motion scale and frequency variability. It normalizes skeleton inputs and applies alternating spatial and temporal GCNs. Spatial modules learn latent joint relationships, while temporal processing is split into global and local paths-extracting coarse-grained and fine-grained motion patterns, respectively.
        
        Finally, in the context of autonomous driving, Wang et al.~\citep{CTCX_WANG2022123} developed a two-stage pose-based system to classify road intersection gestures. The first stage extracts geometric features and encodes them via LSTM layers. The second stage uses a dual-stream setup, where one branch processes geometric descriptors and the other uses a ConvLSTM to handle co-occurrence matrices. Features are fused with directional orientation indicators for final classification.

    \subsection{Multimodal}

        Multimodal CDHGR approaches combine inputs such as RGB, pose heatmaps, and depth to overcome unimodal limitations and capture complementary information.
    
        A key example is TwoStream-SLR~\citep{NEURIPS2022_6cd3ac24_HEATMAPS}, a dual-stream architecture for CSLR and SLT. It processes RGB and pose heatmaps in parallel via the first four S3D blocks in each branch, with bidirectional lateral connections for multi-level information exchange. Sign Pyramid Networks and auxiliary heads (video, keypoint, joint) provide additional supervision, while self-distillation minimizes KL-divergence across outputs. The same framework has been applied to ISLR~\citep{zuo2024onlinecontinuoussignlanguage} using a sliding window to extract sign boundaries and construct training dictionaries.
        
        Other CSLR models have adapted multimodal ISLR backbones. Uni-Sign~\citep{UNISIGNli2025} can be used directly for CSLR, while SignBERT+~\citep{SignBERT+} can be extended by appending temporal pooling and a CTC decoder to generate gloss sequences.
        
        Beyond sign language, general-purpose CDHGR systems have explored multimodality across broader visual sensor inputs. For example,~\citep{pretrained_on_jester_3dDenseNet_LSTM} proposes a framework that segments RGB, depth, infrared, and optical flow inputs using a shared U-Net and processes them via separate 3D DenseNet-43 backbones. Extracted features are fused through Canonical Correlation Analysis to align shared representations across modalities.

    \subsection{Comparison and Critical Summary of CDHGR Methods}
    
        \begin{table}[t]
        \caption{Continuous Dynamic VHGR: horizontal comparison of RGB-based, pose-based, and multimodal method families.}
            \label{tab:cdhgr_methods_summary}
            \centering
            \tiny
          
            \setlength{\tabcolsep}{2pt}
            \renewcommand{\arraystretch}{0.9}

            \newcommand{\caspect}{1.7cm}
            \newcommand{\capp}{3.7cm}
            \newcommand{\cpose}{3.7cm}
            \newcommand{\cmult}{3.7cm}
        
            \begin{tabular}{
            m{\caspect}
            m{\capp}
            m{\cpose}
            m{\cmult}
            }
            \toprule
            \textbf{Aspect} & 
            \textbf{RGB-Based Methods} & 
            \textbf{Pose-Based Methods} & 
            \textbf{Multimodal Methods} \\
            \midrule
        
            Primary function & 
              Mapping of unsegmented pixel sequences directly to gloss/command ones
            & 
              Conversion of skeletal joint coordinate sequences into labels
            & 
              Fusion of complementary signals (e.g., RGB + pose) to resolve ambiguities
            \\
            \midrule
            Most suitable scenarios &
            CSLR and continuous command recognition when RGB video is available and the model must exploit hand shape, facial cues, body context, and scene information
            &
            Resource-constrained or privacy-sensitive continuous recognition where pose extraction is reliable and appearance information is less critical
            &
            High-performance CSLR/SLT settings where RGB and pose/depth cues can be fused to handle co-articulation, ambiguous signs, and weak temporal supervision
            \\
        \midrule
        
            Data representation & 
            \begin{itemize}[leftmargin=*,nosep]
              \item Raw pixel values
            \end{itemize} & 
            \begin{itemize}[leftmargin=*,nosep]
              \item 2D/3D joints
              \item Bones
              \item Motion vectors
              \item Pose heatmaps
            \end{itemize} & 
            \begin{itemize}[leftmargin=*,nosep]
              \item Heterogeneous data embeddings (e.g., RGB + skeleton + depth)
            \end{itemize} \\
            \midrule
        
            Temporal modeling & 
            \begin{itemize}[leftmargin=*,nosep]
              \item 1D-TCN
              \item BiLSTM
              \item Multi-scale temporal perception modules
            \end{itemize} & 
            \begin{itemize}[leftmargin=*,nosep]
              \item ST-GCNs
              \item Temporal decoupling
              \item Transformers
            \end{itemize} & 
            \begin{itemize}[leftmargin=*,nosep]
              \item Cross-modal attention
              \item Lateral connections
              \item Distillation across streams
            \end{itemize} \\
            \midrule
        
            Neural network architectures & 
            \begin{itemize}[leftmargin=*,nosep]
              \item 2DCNN+1DCNN+BiLSTM
              \item 3DCNNs (I3D, Res3D)
              \item Transformers (Swin-MSTP, CLIP-SLA)
            \end{itemize} & 
            \begin{itemize}[leftmargin=*,nosep]
              \item Graph Convolutional Networks (ST-GCN, CoSign)
              \item 1DCNN/TCN + BiLSTM
            \end{itemize} & 
            \begin{itemize}[leftmargin=*,nosep]
              \item Dual-stream encoders (TwoStream-SLR)
              \item Unified frameworks with LLM decoders (Uni-Sign)
            \end{itemize} \\
            \midrule
        
            Strengths & 
            \begin{itemize}[leftmargin=*,nosep]
              \item Rich visual context
              \item Capturing of facial/body cues
              \item No need for off-the-shelf estimators
            \end{itemize} & 
            \begin{itemize}[leftmargin=*,nosep]
              \item Invariant to illumination and background
              \item Low computational cost
              \item Privacy-safe approaches
            \end{itemize} & 
            \begin{itemize}[leftmargin=*,nosep]
              \item Mitigation of individual sensor noise
              \item Superior accuracy
              \item Robust to co-articulation
            \end{itemize} \\
            \midrule
        
            Limitations & 
            \begin{itemize}[leftmargin=*,nosep]
              \item Susceptible to motion blur, occlusions, and cluttered backgrounds
              \item High GFLOP rate
            \end{itemize} & 
            \begin{itemize}[leftmargin=*,nosep]
              \item Reliance entirely on pose estimation quality
              \item Ignoring of appearance cues
            \end{itemize} & 
            \begin{itemize}[leftmargin=*,nosep]
              \item Complex training
              \item High number of network parameters
              \item Inference latency
            \end{itemize} \\
            \midrule
        
            Indicative methods & 
              SlowFastSign \citep{SlowFastSign}, CorrNet+ \citep{corrnetplus_hu2024corrnetsignlanguagerecognition}, CLIP-SLA \citep{CLIP_SLA_alyami2025clip_CSLR_2025}, \citep{diffusion-CSLR-geng2026no}
            & 
              CoSign \citep{CoSign}, FA-STGCN \citep{FA-STGCN}, MSKA-SLR \citep{MSKA_ELSEVIER}
            & 
              TwoStream-SLR \citep{NEURIPS2022_6cd3ac24_HEATMAPS}, Uni-Sign \citep{UNISIGNli2025}, SignBERT+ \citep{SignBERT+}, \citep{Fangyun_Wei_ICCV}
            \\
            \bottomrule
          \end{tabular}%
        \end{table}
    
        Table~\ref{tab:cdhgr_methods_summary} provides a qualitative horizontal comparison of CDHGR method families, while Table~\ref{tab:cslr_performance} reports the computational cost and Word Error Rate (WER) of CSLR methods on three benchmark datasets. Since these results are compiled from the original publications, comparisons should be interpreted mainly within the same benchmark dataset and reported protocol. Comparing other CDHGR methods (e.g., for HRI) is infeasible, as they are evaluated on different datasets and use other metrics. Leveraging Diffusion Models (DMs) as in \cite{diffusion-CSLR-geng2026no} yields a substantial performance gain without dramatically increasing parameters. SlowFastSign \citep{SlowFastSign} achieves strong accuracy with a moderate number of parameters, while TwoStream-SLR \citep{NEURIPS2022_6cd3ac24_HEATMAPS} has slightly higher WER, requires twice as many parameters, and involves a more complex training setup, though WER improves with additional training data \citep{Fangyun_Wei_ICCV}. C\textsuperscript{2}ST \citep{C2ST} is accurate but contains 187.8M parameters, which may limit real-world deployment. For resource-constrained scenarios, the pose-based CoSign-2s \citep{CoSign} requires only 30.1 GFLOPs, though pose estimation cost is not included. DSTEN-CSLR \citep{DSTEN-CSLR-yu2025dsten} is a lightweight RGB-based model but less accurate. Overall, a trade-off between accuracy and computational cost persists.
    
        Summarizing, the key insights that could be derived from Tables~\ref{tab:cdhgr_methods_summary} and~\ref{tab:cslr_performance}, resolving Q3 (see Section~\ref{sec:intro}), are the following:
        \begin{itemize}[leftmargin=*]
            \item Only a few works, most of them pose-based, process regions corresponding to human body parts independently by applying separate networks to each region \citep{MSKA_ELSEVIER, CoSign, MULTI-CUE, CORE-CSLR-zhang2025core}, as this usually requires pose estimation. This technique does not seem to increase performance in RGB-based CSLR \citep{CORE-CSLR-zhang2025core} and may introduce additional errors due to pose-estimation failures.
            \item The three-stage paradigm, namely the 2DCNN+1DCNN+BiLSTM pipeline \citep{VAC-CSLR}, remains dominant even with modifications. However, such methods are often outperformed by specific single-stage approaches \citep{NEURIPS2022_6cd3ac24_HEATMAPS, Fangyun_Wei_ICCV}, which achieve lower WER due to advanced training strategies and the integration of multiple modalities.
            \item ResNet \citep{VAC-CSLR} is the most commonly used spatial-perception baseline.
            \item Considering spatiotemporal modeling in the early stages instead of frame-wise feature extraction may offer slight performance gains for RGB-based CSLR \citep{SlowFastSign}.
            \item RGB-based CSLR can also benefit from adopting visual foundation models for spatial perception \citep{CLIP_SLA_alyami2025clip_CSLR_2025}, owing to their large capacity and extensive pretraining.
            \item The two-stage scheme for temporal modeling, namely short-term followed by long-term modeling, is rarely replaced by generative models \citep{diffusion-CSLR-geng2026no, CVT-SLR, UNISIGNli2025} or single sequential modules \citep{TCNet_Lu_Salah_Poppe_2024,MSKA_ELSEVIER}.
            \item Diffusion models for gloss generation are examined in a single work \citep{diffusion-CSLR-geng2026no}, which achieves superior performance and overcomes some limitations of the standard weakly supervised paradigm.
        \end{itemize}

    \begin{table}[t]
        \centering
        \tiny
        \setlength{\tabcolsep}{0.9pt}
        \renewcommand{\arraystretch}{0.9}
        \caption{Comparison of CSLR methods in terms of computational cost and accuracy (Word Error Rate, WER) on three benchmarks: CSL-Daily \citep{Back-Translation-CVPR}, Phoenix \citep{PHOENIX-2014}, and Phoenix T \citep{NEURAL-SIGN-LANGUAGE-TRANSLATION-CVPR-CAMGOZ}. "Params" denotes the number of \textit{all} parameters, both trainable and frozen. Methods marked with "\dag" have utilized additional training data and "\ddag" enhance the vanilla spatial perception model with custom modules (e.g. correlation modules \cite{CorrNet,corrnetplus_hu2024corrnetsignlanguagerecognition} and CBAM \cite{C2SLR}). "Regions" refers to whether the methods consider the body regions (e.g. grouping keypoints accordingly \cite{MSKA_ELSEVIER}) or not. Reported GFLOPs and parameters follow the original publications; preprocessing overhead, such as pose estimation, may not be included unless explicitly stated. The top-3 metrics are shown with bold.}
        \begin{tabular}{ll|c|c|l|l|ll|lll}
        \toprule
        & & \multicolumn{4}{c|}{Architectural Elements} & \multicolumn{5}{c}{Performance} \\
        \midrule
        Method & Year & Regions & Stages & Backbone & Decoder & GFLOPs & Params & CSL-Daily & Phoenix & Phoenix-T \\
        \midrule

        \multicolumn{11}{c}{RGB-based} \\
        \midrule
        
        \citep{diffusion-CSLR-geng2026no} &2025 & \xmark & 2 & ResNet & DM & - 
        & 86.64M & \textbf{23.7\%} & \textbf{16.8\%} & \textbf{17.2\%} \\
        CORE \citep{CORE-CSLR-zhang2025core}&2025 & \cmark & 3 & ResNet & GNN+BiLSTM & - & - & 31.1\% & 21.7\% & 21.6\% \\
        SD-Net \citep{SD-Net-CSLR-gao2025structure} &2025 & \xmark & - & ResNet & Hybrid & - & - & 27.9\% & - & 20.3\% \\
        STNet \citep{STNet-wang2025continuous}&2025 & \xmark & 3 & ResNet & 1DCNN+BiLSTM & - & - & 27.2\% & 19.3\% & 19.8\% \\
        CSGC \citep{CSGC_Rao_Sun_Wang_Wang_Zhang_2024} &2024 & \xmark & 3 & ResNet & 1DCNN+BiLSTM & - & - & 26.7\% & 19\% & 19.5\% \\
        GPGN \citep{GPGN} &2024 & \xmark & 3 & ResNet & TCN+BiGRU & - & - & 30\% & 20.4\% & 20.5\% \\
        CTCA \citep{CTCA} &2023& \xmark & 2 & ResNet & TCN \& BiLSTM & - & - & 29.4\% & 20.1\% & 20.3\% \\
        CVT-SLR \citep{CVT-SLR} &2023 & \xmark & 2 & ResNet & VAE & - & - & - & 20.1\% & 20.3\% \\
        TLP \citep{TemporalLiftPooling} &2022 & \xmark & 3 & ResNet & TLP+BiLSTM & 2950.5 & 48M & - & 20.8\% & 21.2\% \\
        SMKD \citep{SMKD} &2021 & \xmark & 3 & ResNet & 1DCNN+BiLSTM & 1032.3 & 20.3M & - & 21\% & 22.4\% \\
        VAC-CSLR \citep{VAC-CSLR} &2021 & \xmark & 3 & ResNet & 1DCNN+BiLSTM & 1120.4 & 22.3M & - & 22.3\% & - \\
        DSTEN-CSLR \citep{DSTEN-CSLR-yu2025dsten}\ddag &2025 & \xmark & 3 & ResNet & TCN+FFN \& Memory & 366.3 & 30.19M & 28.1\% & 20.0\% & 20.3\% \\
        TCNet \citep{TCNet_Lu_Salah_Poppe_2024}\ddag &2024& \xmark & 2 & ResNet & BiLSTM & 932.1 & 19.3M & 29.3\% & 18.9\% & 19.4\% \\ 
        CorrNet+ \citep{corrnetplus_hu2024corrnetsignlanguagerecognition}\ddag &2024& \xmark & 3 & ResNet & 1DCNN+BiLSTM & - & - & 28.2\% & 18.2\% & 19.1\% \\
        CorrNet \citep{CorrNet}\ddag &2023 & \xmark & 3 & ResNet & 1DCNN+BiLSTM & 1035.4 & 20.5M & 30.1\% & 19.4\% & 20.5\% \\
        SEN-CSLR \citep{SEN-CSLR}\ddag &2023& \xmark & 3 & ResNet & 1DCNN+BiLSTM & 1144.2 & 23.1M & 30.7\% & 21\% & 20.7\% \\
        C\textsuperscript{2}SLR \citep{C2SLR}\ddag &2024& \xmark & 3 & VGG & TCN+Transformer & 1124.6 & 32.6M & - & 20.4\% & 20.4\% \\
        SLA-Adapter \citep{CLIP_SLA_alyami2025clip_CSLR_2025} &2025 & \xmark & 3 & CLIP ViT & TCN+BiLSTM & - & - & 25.8\% & 18.8\% & 19.5\% \\
        SLA-LoRA \citep{CLIP_SLA_alyami2025clip_CSLR_2025} &2025 & \xmark & 3 & CLIP ViT & TCN+BiLSTM & -&  - & 25.8\% & 19.3\% & 19.4\% \\
        Swin-MSTP \citep{SWIN_MSTP_Neurocomputing_2025_CSLR} &2025 & \xmark & 3 & SwinT & TLP+BiLSTM & - & - & 27.1\% & 18.7\% & 19.7\% \\
        C\textsuperscript{2}ST \citep{C2ST} &2023 & \xmark & 3 & SwinT & 1DCNN+BiLSTM & 1635 & 187.8M & 25.8\% & \textbf{17.7\%} & 18.9\% \\
        SlowFastSign \citep{SlowFastSign} &2024 & \xmark & 3 & SlowFast & 1DCNN+BiLSTM & - & 52.5M & \textbf{24.9\%} & 18.3\% & \textbf{18.7\%} \\
        MultiSignGraph \citep{SignGraph_CVPR} &2024 & \xmark & 3 & ST-GCN & 1DCNN+BiLSTM & - & - & 26.4\% & 19.1\% & 19.1\% \\

        \midrule
        \multicolumn{11}{c}{Pose-based} \\
        \midrule
        
        MSKA-SLR \citep{MSKA_ELSEVIER} & 2025 & \cmark & 2 & MHA & TCN & - & - & 27.1\% & 21.2\% & 19.8\% \\ 
        CoSign-2s \citep{CoSign} &2023& \cmark & 3 & GNN & 1DCNN+BiLSTM & 30.1 & 28.2M & 27.2\% & 20.1\% & 20.1\% \\ 
        STMC \citep{MULTI-CUE} &2022 & \cmark & 3 & VGG & TCN+BiLSTM & 1742.1 & 40.71M & - & 20.7\% & 21\% \\ 

        \midrule
        \multicolumn{11}{c}{Multimodal} \\
        \midrule
        
        Uni-Sign \citep{UNISIGNli2025}\dag &2025 & \cmark & 3 & Various & Temporal encoders+LLM & - & 592.1M & 26\% & - & - \\ 
        \citep{Fangyun_Wei_ICCV}\dag &2023& \xmark & 1 & \multicolumn{2}{c|}{ -- Dual-stream S3D --} & - & 105.2M & - & \textbf{16.7\%} & \textbf{18.5\%} \\ 
        TwoStream-SLR \citep{NEURIPS2022_6cd3ac24_HEATMAPS} &2022& \xmark & 1 & \multicolumn{2}{c|}{ -- Dual-stream S3D --} & - & 105.2M & \textbf{25.3\%} & 18.8\% & 19.3\% \\ 
        
        \bottomrule
        
        \end{tabular}
        \label{tab:cslr_performance}
    \end{table}

\section{Summary of SOTA Methods}
\label{sec:comparison}

    After reviewing numerous recent works, this section focuses on SOTA methods across tasks, summarizing the field and comparing key methods in terms of architecture, deployment, strengths, and limitations in order to address Q2 and Q3 (see Section \ref{sec:intro}). Table \ref{tab:cross-task-comparison} highlights each method's task, application domain, input modality, architecture, and major strengths and weaknesses.
    
    One insight from the table is that SLU dominates other fields, likely due to its complexity, which requires advanced visual cues (including facial expressions), extensive temporal modeling for dynamic gestures, and classifiers capable of distinguishing highly similar gestures. Consequently, few SOTA methods exist for static SLR, as they struggle to capture hand movements. General-purpose IDHGR methods, used as baselines for tasks like SLU, HRI, and HCI, have also been widely studied; the similarity of gestures in HRI and HCI allows the same methods to serve both domains.
    
    Depth, pose, and other non-RGB modalities are used less frequently and typically in combination with RGB streams. This is because they either cannot represent fine-grained gestures (depth, optical flow) or perform poorly in dynamic recognition (pose estimators often fail on blurry images). RGB-based models---even older or simpler ones---already achieve strong results on SHGR datasets \citep{HAGRIDv2_ARXIV}, so incorporating pose can incur significant computational overhead without substantial accuracy gains. In CSLR, rapid hand movements degrade pose estimator performance, making pose usable only alongside RGB \citep{NEURIPS2022_6cd3ac24_HEATMAPS} and increasing complexity. Thus, pose-based approaches are mostly limited to IDHGR, with some exceptions \citep{CoSign}, where incorporating multiple modalities can notably improve performance \citep{SAM_SLR,jiang2021sign-sam-slr-v2}, even with late fusion.
    
    Regarding architecture, SHGR's simplicity allows even small 2DCNNs \citep{EDenseNet,G-CNN-SHARMA2021115657} to achieve very high accuracy ($\ge$95\%). However, the limited size and diversity of datasets such as those proposed in \citep{NUS-I-2010,NUS-II-2013,ouhands,triesch2001system} can lead to overfitting, reducing real-world performance. In contrast, IDHGR and especially ISLR rely on more complex models. ST-GCN variants \citep{JOINTS_MOTION_VECTORS_Improved_ST-GCN,TD-GCN} and GCN+Transformer pipelines \citep{SignBERT_ICCV,SignBERT+,BEST,Weichao-Zhao} mainly process skeleton data, often combined with RGB 3DCNNs (e.g., I3D, Res3D) to compensate for pose estimation errors. While many studies focus on improving pose-based streams, RGB modalities are less explored. SAM-SLR \citep{SAM_SLR} and SAM-SLRv2 \citep{jiang2021sign-sam-slr-v2} propose advanced separate processing for each modality, representing current SOTA for IDHGR. RGBD models \citep{RAAR3DNet,CTFB_Hampiholi,Yunan_Li} typically use dual-stream 3DCNNs, increasing computational cost. CSLR demands even more complex architectures to capture short-term (gloss-wise) and long-term (global) context. The 2DCNN+1DCNN+BiLSTM pipeline remains common, with relatively fewer parameters than SHGR models (e.g., Gesture-CNN \citep{G-CNN-SHARMA2021115657} 67M vs. large CSLR models 48M \citep{TemporalLiftPooling}). Vision transformers increasingly replace 2DCNN backbones \citep{SWIN_MSTP_Neurocomputing_2025_CSLR,C2ST,CLIP_SLA_alyami2025clip_CSLR_2025} for their attention-based focus on critical regions. TwoStream-SLR \citep{NEURIPS2022_6cd3ac24_HEATMAPS} employs a dual-stream 3DCNN with multimodal fusion and pose heatmaps (instead of skeleton joints) alongside RGB frames, achieving high accuracy.
    
    As mentioned earlier, SHGR is a relatively easy task, so researchers have not proposed any notable learning strategy. Pose-based methods for ISLR often \citep{SignBERT_ICCV,SignBERT+,BEST,Weichao-Zhao} insert hand topology prior through SSL. NAS is also utilized to determine the optimal structure of 3DCNNs. Models for CSLR face serious optimization difficulties compared to other methods due to their complexity. Thus, extra supervision is provided via additional losses attached to various layers of the models \citep{VAC-CSLR,CLIP_SLA_alyami2025clip_CSLR_2025}. Leveraging additional training data combined with accurate architectures during training \citep{Fangyun_Wei_ICCV} increases performance; therefore, methods marked as using additional data should not be compared one-to-one with methods trained only on the original benchmark data.
    
    In summary, LHGR-Net \citep{LHGR-Net} and stand-alone pose-based methods \citep{Yangke_patcog, TD-GCN, CoSign} are suitable for resource-constrained use cases (assuming that the cost of pose estimation is negligible, which is true especially when the keypoints are extracted by the deployed camera), balancing the computational complexity with recognition accuracy. TwoStream-SLR \citep{NEURIPS2022_6cd3ac24_HEATMAPS}, C\textsuperscript{2}ST \citep{C2ST} and  \citep{diffusion-CSLR-geng2026no} achieve superior performance for CSLR but come with complex training settings and large computational cost. NLA-SLR \citep{NLA_SLR} and SAM-SLRv2 \citep{jiang2021sign-sam-slr-v2} surpass most of the related methods, but they require substantial computational resources. Uni-Sign \citep{UNISIGNli2025} introduced a novel approach by combining various SLU tasks into a single sequence generation task. Although it achieves significant performance and unifies a variety of tasks providing a shared framework, the computational complexity makes real-world deployment difficult, and the requirement of training data with detailed annotations prevents its adoption in other sign languages.

\begin{sidewaystable}[]
    \setlength{\tabcolsep}{1pt}
    \renewcommand{\arraystretch}{2}
    \caption{Comparison of current SOTA methods. They are compared in terms of task, application domain (referred to as "Domain" for the sake of brevity), input modality (or modalities) and general architecture. The strengths of each method or the key innovations introduced as well as some notable limitations of them, are also mentioned.}
    \tiny
    \begin{tabular}{llllllll}
    
                \toprule
                \parbox{2cm}{\textbf{Method}} & 
                \parbox{1cm}{\textbf{Task}} & 
                \parbox{1cm}{\textbf{Domain}} & 
                \parbox{1cm}{\textbf{Modality}} & 
                \parbox{4cm}{\textbf{Architecture}} & 
                \parbox{6cm}{\textbf{Strengths or Key Innovation}} & 
                \parbox{4cm}{\textbf{Limitations}} \\
                \midrule
                
                \parbox{2cm}{FGDSNet \citep{FGDSNet}} & 
                \parbox{1cm}{SHGR} & 
                \parbox{1cm}{HRI} & 
                \parbox{1cm}{RGB} &
                \parbox{4cm}{Dual-stream hand segmentation model + dual-stream 2DCNN + mid-level fusion} &
                \parbox{6cm}{The dual-stream classifier addresses any failures of the task-specific segmentation model.} &
                \parbox{4cm}{The segmentation model and the 2DCNN backbones must be trained separately.} \\

                \parbox{2cm}{LHGR-Net \citep{LHGR-Net}} & 
                \parbox{1cm}{SHGR} & 
                \parbox{1cm}{HCI} & 
                \parbox{1cm}{RGB} &
                \parbox{4cm}{Atrous convolutional layers followed by depth separable convolutions and attention.} &
                \parbox{6cm}{Suitable for resource-constrained deployment.} &
                \parbox{4cm}{Very small capacity.} \\

                \parbox{2cm}{2DPSTPP-Net \citep{2DPSTPP}} & 
                \parbox{1cm}{SHGR} & 
                \parbox{1cm}{HCI} & 
                \parbox{1cm}{RGB} &
                \parbox{4cm}{2DCNN with custom pooling layers and residual connections} &
                \parbox{6cm}{Spatial pyramid pooling and residual connections obtain robust representations.} &
                \parbox{4cm}{The model is relatively large for SHGR.} \\

                \parbox{2cm}{Gesture-CNN \citep{G-CNN-SHARMA2021115657}} & 
                \parbox{1cm}{SHGR} & 
                \parbox{1cm}{SLU} & 
                \parbox{1cm}{RGB} &
                \parbox{4cm}{2DCNN consisting of four convolutional layers} &
                \parbox{6cm}{The simplicity of the architecture enables easy implementation and training.} &
                \parbox{4cm}{Relatively large (67M parameters).} \\
                \midrule

                \parbox{2cm}{SAM-SLR \citep{SAM_SLR}, SAM-SLR-v2 \citep{jiang2021sign-sam-slr-v2}} & 
                \parbox{1cm}{IDHGR} & 
                \parbox{1cm}{SLU} & 
                \parbox{1cm}{RGBD, Pose, HHA, OF, DF} &
                \parbox{4cm}{Six/seven streams followed by late fusion} &
                \parbox{6cm}{Incorporated a large number of modalities, achieving very high accuracy.} &
                \parbox{4cm}{Insufficient fusion and computationally expensive.} \\
                \midrule

                \parbox{2cm}{SignBERT \citep{SignBERT_ICCV}, SignBERT+ \citep{SignBERT+} (ISLR)} & 
                \parbox{1cm}{IDHGR} & 
                \parbox{1cm}{SLU} & 
                \parbox{1cm}{RGB, Pose} &
                \parbox{4cm}{Spatial GCN + Transformer \& 3DCNN + late fusion} &
                \multirow{3}{*}{\parbox{6cm}{The model gains hand topology prior through SSL pretraining (masked hand tokens reconstruction, inspired by BERT)}} &
                \multirow{3}{*}{\parbox{4cm}{Needs extensive SSL pretraining, the processing and incorporation of the RGB modality may be suboptimal}} \\

                \parbox{2cm}{SignBERT+ \citep{SignBERT+} (CSLR)} & 
                \parbox{1cm}{CDHGR} & 
                \parbox{1cm}{SLU} & 
                \parbox{1cm}{RGB, Pose} &
                \parbox{4cm}{Spatial GCN + Transformer \& 3DCNN + intermediate fusion + BiLSTM} &
                & & \\

                \parbox{2cm}{BEST \citep{BEST}} & 
                \parbox{1cm}{IDHGR} & 
                \parbox{1cm}{SLU} & 
                \parbox{1cm}{RGB, Pose} &
                \parbox{4cm}{VAE + Spatial GCN + Transformer \& 3DCNN + late fusion} &
                & & \\
                \midrule

                \parbox{2cm}{NLA-SLR \citep{NLA_SLR}} & 
                \parbox{1cm}{IDHGR} & 
                \parbox{1cm}{SLU} & 
                \parbox{1cm}{RGB, Pose} &
                \parbox{4cm}{Four-stream network based on S3D with two frame rates and multi-level fusion} &
                \parbox{6cm}{Novel architecture performing multimodality fusion efficiently. Language-aware label smoothing facilitates training. Robust to visually similar signs recognition.} &
                \parbox{4cm}{Is affected by the performance of word encoders} \\
                \midrule

                \parbox{2cm}{Uni-Sign \citep{UNISIGNli2025}} & 
                \parbox{1cm}{IDHGR, CDHGR} & 
                \parbox{1cm}{SLU} & 
                \parbox{1cm}{RGB, Pose} &
                \parbox{4cm}{A pretrained LLM process features extracted by pose, visual and temporal encoders} &
                \parbox{6cm}{Provides a unified framework for multiple SLU tasks, increasing the recognition accuracy.} &
                \parbox{4cm}{Requires additional training data} \\
                \midrule

                \parbox{2cm}{TD-GCN \citep{TD-GCN}} & 
                \parbox{1cm}{IDHGR} & 
                \parbox{1cm}{-} & 
                \parbox{1cm}{Pose} &
                \parbox{4cm}{ST-GCN variant} &
                \parbox{6cm}{Facilitates temporal modeling through non-fixed time-dependent topology} &
                \parbox{4cm}{Depends on the pose estimation performance.} \\

                \parbox{2cm}{SBI-DHGR \citep{SBI-DHGR}} & 
                \parbox{1cm}{IDHGR} & 
                \parbox{1cm}{-} & 
                \parbox{1cm}{Pose} &
                \parbox{4cm}{Multi-stream 1DCNN + BiLSTM followed by late fusion} &
                \parbox{6cm}{Utilizes features of various levels of granularity.} &
                \parbox{4cm}{Very large compared to other pose-based methods} \\

                \parbox{2cm}{DSTSA-GCN \citep{DSTSA-CGN_cui2025dstsa} (4s)} & 
                \parbox{1cm}{IDHGR} & 
                \parbox{1cm}{-} & 
                \parbox{1cm}{Pose} &
                \parbox{4cm}{Four stream GCN + TCN} &
                \parbox{6cm}{Group channel-wise and temporal-wise GCNs with dynamic graph topologies improve the performance.} &
                \parbox{4cm}{The static part of the hand topology is initialized randomly, lacking prior.} \\
                \midrule

                \parbox{2cm}{SlowFast+NAS \citep{3D_CDC_NAS_TIP_2021}} & 
                \parbox{1cm}{IDHGR} & 
                \parbox{1cm}{-} & 
                \parbox{1cm}{RGBD, OF} &
                \parbox{4cm}{Multi-stream 3DCNN with 3D-CDC and multi-level fusion} &
                \parbox{6cm}{Enhanced temporal modeling through central difference convolution and optimal hyperparameter selection through NAS.} &
                \parbox{4cm}{Needs substantial computational resources (251.4M params, 232.1GFLOPs) compared to similar methods} \\
                \midrule

                \parbox{2cm}{RAAR3DNet \citep{RAAR3DNet}} & 
                \parbox{1cm}{IDHGR} & 
                \parbox{1cm}{-} & 
                \parbox{1cm}{RGBD} &
                \parbox{4cm}{Dual stream I3D with adoptable cells, spatial and temporal attention} &
                \parbox{6cm}{NAS improves the architecture. The spatial attention mechanism is supervised by an off-the-shelf pose estimator.} &
                \parbox{4cm}{It relies on the correctness of the pose estimator} \\

                \parbox{2cm}{CTFB \citep{CTFB_Hampiholi}} & 
                \parbox{1cm}{IDHGR} & 
                \parbox{1cm}{-} & 
                \parbox{1cm}{RGBD} &
                \parbox{4cm}{Dual-stream 3DCNN + Multi-level fusion with transformer-based modules} &
                \parbox{6cm}{Performs efficient multimodality fusion via multiple attention modules placed in different layers.} &
                \parbox{4cm}{The scaling to more than two modalities is not straightforward} \\

                \midrule

                \parbox{2cm}{\citep{diffusion-CSLR-geng2026no}} & 
                \parbox{1cm}{CDHGR} & 
                \parbox{1cm}{SLU} & 
                \parbox{1cm}{RGB} &
                \parbox{4cm}{Visual encoder followed by a Diffusion Model (DM).} &
                \parbox{6cm}{Using a DM for gloss generation makes it very accurate.} &
                \parbox{4cm}{Increased complexity.} \\

                \parbox{2cm}{CLIP-SLA \citep{CLIP_SLA_alyami2025clip_CSLR_2025}} & 
                \parbox{1cm}{CDHGR} & 
                \parbox{1cm}{SLU} & 
                \parbox{1cm}{RGB} &
                \parbox{4cm}{CLIP visual encoder (with MLP adapters or LoRA) + TCN + BiLSTM} &
                \parbox{6cm}{Early supervision by adding an extra classifier accelerates training, while the utilization of a foundation model enhances the feature extraction.} &
                \parbox{4cm}{The utilization of CLIP implies high computational cost.} \\

                \parbox{2cm}{Swin-MSTP \citep{SWIN_MSTP_Neurocomputing_2025_CSLR}} & 
                \parbox{1cm}{CDHGR} & 
                \parbox{1cm}{SLU} & 
                \parbox{1cm}{RGB} &
                \parbox{4cm}{SwinT+1DCNN/TLP} &
                \parbox{6cm}{Deploying a multiscale TCN, the model is able to recognize continuous signs of varying time durations.} &
                \parbox{4cm}{Computationally expensive.} \\
                \midrule

                \parbox{2cm}{C2ST \citep{C2ST}} & 
                \parbox{1cm}{CDHGR} & 
                \parbox{1cm}{SLU} & 
                \parbox{1cm}{RGB} &
                \parbox{4cm}{SwinT+1DCNN+BiLSTM} &
                \parbox{6cm}{CGD loss (that avoids the conditional independence assumption of CTC) and self-distillation increase the performance.} &
                \parbox{4cm}{The vision transformer as well as the language model utilized during training increase the computational complexity.} \\
                \midrule

                \parbox{2cm}{CoSign-2s \citep{CoSign}} & 
                \parbox{1cm}{CDHGR} & 
                \parbox{1cm}{SLU} & 
                \parbox{1cm}{Pose} &
                \parbox{4cm}{ST-GCN variant} &
                \parbox{6cm}{The computational cost is negligible, thus CoSign-2s perfectly balances efficient execution and accurate recognition.} &
                \parbox{4cm}{The performance depends on the attributes of the pose estimator.} \\
                \midrule

                \parbox{2cm}{\citep{Fangyun_Wei_ICCV}} & 
                \parbox{1cm}{CDHGR} & 
                \parbox{1cm}{SLU} & 
                \parbox{1cm}{RGB, Pose} &
                \parbox{4cm}{TwoStream-SLR} &
                \parbox{6cm}{Utilization of extra training data via dictionary construction.} &
                \parbox{4cm}{Needs additional training data.} \\

                \parbox{2cm}{TwoStream-SLR \citep{NEURIPS2022_6cd3ac24_HEATMAPS}} & 
                \parbox{1cm}{CDHGR} & 
                \parbox{1cm}{SLU} & 
                \parbox{1cm}{RGB, Pose} &
                \parbox{4cm}{Dual-stream 3DCNN with bidirectional lateral connections} &
                \parbox{6cm}{Extra supervision by auxiliary losses and self-distillation applied to additional prediction heads facilitate training. The multi-level fusion incorporates efficiently the two modalities.} &
                \parbox{4cm}{Increased complexity due to the use of two streams and an off-the-shelf pose estimator. Very large.} \\

                \bottomrule
                
    \end{tabular}
    \label{tab:cross-task-comparison}
\end{sidewaystable}

\section{Gesture-based Interaction Paradigms}
\label{sec:interaction-paradigms}

    Having investigated the different algorithmic perspectives of Visual Hand Gesture Recognition (VHGR) methods and their particular characteristics in Sections \ref{sec:taxonomy}-\ref{sec:comparison}, this section outlines how automatically recognized gestures are practically used in real-world interactive systems. In particular, the principal and most widely used gesture-based interaction paradigms, enabled by the use of VHGR methods, fall into the following closely related areas, which are described below: a) Human-Computer Interaction (HCI), with particular emphasis on immersive Virtual and Augmented Reality (VR/AR), b) Human-Robot Interaction (HRI), and c) Human-Machine Interaction (HMI). The following figure (Fig.~\ref{fig:composite_interaction}) illustrates indicative applications of gesture-based interaction.
    
    \begin{figure}[t]
        \centering
        \small
        \setlength{\tabcolsep}{5pt}
        \renewcommand{\arraystretch}{5}
        \newcommand{\width}{0.45}
        \begin{tabular}{cc}
            \begin{subfigure}[t]{\width\textwidth}
                \centering
                \includegraphics[trim={0 0.1cm 0.13cm 0},clip,width=\linewidth]{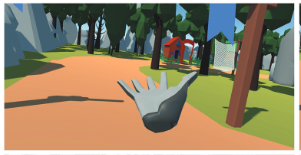}
                \caption{}
            \end{subfigure} &
            \begin{subfigure}[t]{\width\textwidth}
                \centering
                \includegraphics[width=\linewidth]{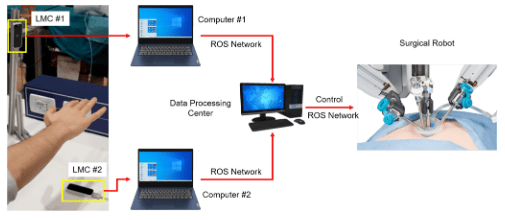}
                \caption{}
            \end{subfigure} \\
            \begin{subfigure}[t]{0.4\textwidth}
                \centering
                \includegraphics[width=\linewidth]{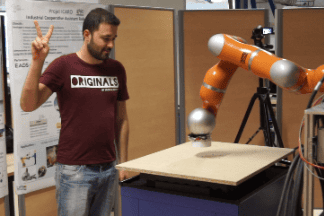}
                \caption{}
            \end{subfigure} &
            \begin{subfigure}[t]{\width\textwidth}
                \centering
                \includegraphics[width=\linewidth]{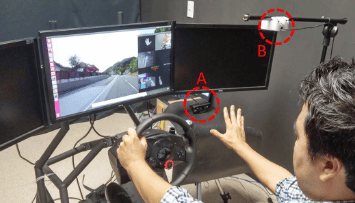}
                \caption{}
            \end{subfigure} \\
        \end{tabular}
        \caption{Indicative works for gesture-based interaction schemes: a) locomotion within a virtual environment by identifying the palm orientation \citep{INTERACTION-LOCMOTION-PALM-INDEX-caggianese2020freehand}, b) surgical robots teleoperation via gestures \citep{TELEOPERATED_SURGICAL_ROBOT}, c) robotic arm control in a teach-by-demonstration paradigm \citep{INTERACTION-ROBOTIC-ARM-mazhar2019real} and d) car manipulation \citep{NVIDIA_GESTURE_R3DCNN_Molchanov}.}
        \label{fig:composite_interaction}
    \end{figure}

\subsection{Gesture-Based Interaction in HCI: VR/AR}
\label{ssec:paradigms-vrar}

Gesture-based interfaces have been expanded in immersive VR/AR applications, replacing controllers and more cumbersome devices to enable more convenient, user-friendly, and intuitive interaction.
In this context, the main gesture-based interaction paradigms are as follows:
\begin{itemize}[leftmargin=*]
    \item \textbf{Selection and manipulation}: Authors in  \cite{bowman1997evaluation} introduce the so-called "Go-Go" and "HOMER" techniques for arm-extension-based virtual object selection.
    Additionally, \cite{mine1997movingobjects} show that body-relative interaction techniques exploiting proprioception can improve both speed and accuracy compared to world-relative ones, a principle that motivates the prevalence of pinch and grab metaphors in current Head-Mounted Displays (HMDs). The survey \cite{argelaguet2013survey} provides an overview of 3D selection techniques, while \cite{mendes2019survey} performs a similar task for direct object manipulation, jointly establishing an evaluation framework for gesture-based interfaces. Authors in \citep{INTERACTION-DirectionQ-kang2022directionq} propose a framework termed DirectionQ for multiple object selection, considering three gesture commands: pinch to start the operation, release to stop the selection and hand movement in the desired direction to indicate one of the selected objects.
    
    \item \textbf{Mid-air gesture vocabularies for VR/AR}: Authors in \cite{piumsomboon2013userdefined} perform an elicitation study for HMD-AR, deriving a user-defined gesture set 
    and concluding that one-handed (or unimanual) semaphoric commands are strongly preferred over two-handed (also referred to as bimanual) ones. Authors in \cite{buchmann2004fingartips} 
    propose FingARtips, which comprises a gesture-based direct-manipulation AR interface based on a fingertip-tracked pinch metaphor, and \cite{reifinger2007static} show that combining static HGR with HMM-based dynamic HGR is sufficient to drive AR menu systems.
    Interaction paradigms for immersive VR/AR on consumer hardware (Microsoft HoloLens 2, Meta Quest) are summarized in \citep{hertel2021taxonomy}, reporting that practitioners converge on a small core gesture vocabulary (pinch, point, grab and air-tap) providing a minimum set of classes that models should support.
    
    \item \textbf{Locomotion and bimanual interaction}:
    Locomotion refers to navigation through the virtual environment.
    An indicative work is \citep{INTERACTION-LOCMOTION-PALM-INDEX-caggianese2020freehand}, which examines three settings for gesture-based locomotion: palm orientation, index-finger direction, and gaze steering. User feedback showed that these techniques are comparable to conventional interfaces, such as controllers, in terms of preference.
    Additionally, \cite{cutler1997twohanded} shows that bimanual direct manipulation on the Responsive Workbench substantially outperforms unimanual alternatives, anticipating the bimanual interaction techniques prevalent in consumer HMDs.
    
    \item \textbf{Ergonomics and the gorilla-arm problem}: A persistent concern in VR/AR is the so-called "gorilla arm" effect that arises when users hold their arms aloft to perform mid-air gestures. Authors in \cite{hincapie2014consumed} introduce the "Consumed Endurance" metric to quantify shoulder fatigue and show that subtle changes in elbow elevation drastically extend the comfortable interaction time. Consumed Endurance is widely used in evaluations of new gesture interfaces and motivates a class of micro-gesture interaction styles performed close to the body (e.g., \cite{sridhar2017watchsense} and \cite{chan2016microgestures}).

    \item \textbf{Multimodal gesture-gaze coupling}: An active research trend involves the combination of gestures with gaze. In particular, authors in \cite{pfeuffer2017gazepinch} introduce the gaze + pinch paradigm, in which gaze performs targeting and a discrete pinch gesture confirms the action; the combination greatly mitigates the Heisenberg effect and the precision cost of mid-air pointing, and is the default interaction style of products, such as the Apple Vision Pro.
    
\end{itemize}

\subsection{Gesture-Based Interaction in HRI}
\label{ssec:paradigms-hri}

In the context of HRI, gestures serve a dual purpose, comprising both explicit commands issued to a robot and implicit social cues that ground joint activity. Indicative gesture-based HRI paradigms are described below:

\begin{itemize}[leftmargin=*]
    \item \textbf{Industrial and collaborative robotics}: Authors in \cite{liu2018gesture} summarize gesture-based HRI in manufacturing and articulate the trade-off between static-pose commands (low cognitive load, reliable recognition) and dynamic-trajectory gestures (greater expressivity, higher recognition cost and latency). Furthermore, authors in \cite{sheikholeslami2017cooperative} study the complementary problem of robot-emitted instructional gestures, identifying the hand configurations that humans most readily decode.
    Authors in \citep{INTERACTION-ROBOTIC-ARM-mazhar2019real} propose static sign language gestures for defining the trajectory of a robotic arm in a teach-by-demonstration scheme.
    
    \item \textbf{Egocentric and shared-perspective gestures}: 
    Other approaches propose on-board egocentric HGR systems to facilitate collaboration.
    In particular, human-robot handovers \cite{strabala2013seamless} show that the success of these short interactions depends less on grip force or trajectory accuracy than on the timely interpretation of subtle hand cues observed from the robot's first-person viewpoint. 
    From the recognition side, egocentric datasets and pipelines, such as EgoHands \cite{bambach2015lending} and recent in-the-wild hand-object reconstruction methods \cite{cao2021reconstructing}, provide the visual foundation that these scenarios require.
    
    \item \textbf{Legibility and social acceptability}: Recent findings suggest that gesture sets used in public-facing HRI should map onto culturally established symbolic actions, both to ease user learning and to make the interaction legible to bystanders. In particular, authors in \cite{dragan2013legibility} formalize legibility as the property of motion that allows an observer to infer the actor's intent quickly and confidently, explicitly distinguishing it from raw predictability and arguing that legibility drives the perceived naturalness of HRI. In parallel, the social acceptability literature \cite{rico2010social}, \cite{ahlstrom2014acceptance} and \cite{koelle2017acceptability} consistently shows that users self-censor gestures they perceive as conspicuous, regardless of how accurately the system recognizes them.

\end{itemize}

\subsection{Gesture-Based Interaction in HMI}
\label{ssec:paradigms-hmi}

Beyond computers and robots, gesture interfaces increasingly mediate interaction with everyday machines, including vehicles, smart appliances, Internet of Things (IoT), public displays and wearable devices. In this context, the main gesture-based interaction paradigms are analyzed below:

\begin{itemize}[leftmargin=*]
    \item \textbf{Automotive in-cabin interaction}: In human-car interaction, mid-air gestures enable drivers to control infotainment, increasing safety and convenience. In particular, authors in \cite{may2017multimodal} report that gesture commands combined with auditory feedback reduce visual demand compared with touch screens, while authors in \cite{pickering2007hand} characterize the practical recognition challenges (variable lighting, partial occlusion, restricted gesture amplitudes) that in-cabin VHGR systems must address. Various datasets have been proposed for in-vehicle gesture interaction, including Briareo \citep{Briareo_dataset}, nvGesture \citep{NVIDIA_GESTURE_R3DCNN_Molchanov}, and EgoDriving \citep{EgoFormer_METHOD_EgoDriving_DATASET}.
    
    \item \textbf{Smart-home, IoT and public displays}: For consumer electronics and smart-home control, works on imaginary interfaces \cite{gustafson2010imaginary}, Skinput \cite{harrison2010skinput} and ShoeSense \cite{bailly2012shoesense} show that gestures can be sensed on, around and through the body without dedicated displays, enabling pervasive control of IoT devices. For public displays, authors in \cite{walter2013strikeapose} study how passers-by discover that mid-air gestures are accepted, identifying the gesture immediacy problem and showing that visible cues can dramatically increase the rate of spontaneous interaction.
    
    \item \textbf{Wearable computing}:
    Authors in \cite{chan2016microgestures} and \cite{sridhar2017watchsense} demonstrate that subtle, unimanual micro-gestures performed in the user's resting hand posture are well suited to socially acceptable always-on interaction.
    
    \item \textbf{Discoverability and feedback}: Across HMI scenarios, mid-air gestures suffer from being invisible affordances. Strategies developed, such as dynamic guides \cite{bau2008octopocus} and progressive disclosure, are increasingly adopted as overlays in AR/HMI to scaffold gesture learning.

\end{itemize}

\section{Datasets}
\label{sec:datasets}

    The performance and generalizability of gesture recognition models are heavily influenced by the quality, diversity, and scale of the datasets used during training and evaluation. Over the years, a wide variety of datasets have been developed to support different HGR tasks, ranging from static and isolated gestures to continuous sign language recognition in real-world settings. These datasets vary widely in terms of the number of gestures or glosses, the types of modalities captured (e.g., RGB, depth, pose, IR), the presence of signer diversity, and the complexity of the sequences. In the following subsections, we categorize and review the most prominent datasets according to the specific recognition task they target: (i) Isolated and Continuous Sign Language Recognition (IDSLR and CSLR, respectively), (ii) Semaphoric Gestures (both static and dynamic), and (iii) Other Gesture Recognition tasks. This organization aims to provide a clear overview of the resources available for each task type and facilitate their appropriate use in future research. Indicative RGB samples from selected representative datasets are shown in Fig.~\ref{fig:datasets-samples}, covering sign language recognition and semaphoric HGR datasets. Some frames have been slightly cropped for visualization purposes.

    \newcommand\width{0.12}
    
    \begin{figure}[]
        \centering
        \small
        \setlength{\tabcolsep}{2pt}
        \renewcommand{\arraystretch}{0.1}
        \begin{tabular}{ccccccc}
            \toprule
    
            \multicolumn{7}{c}{Sign Language Recognition} \\
            \midrule
            
            \begin{subfigure}[t]{\width\textwidth}
                \centering
                \includegraphics[width=\linewidth]{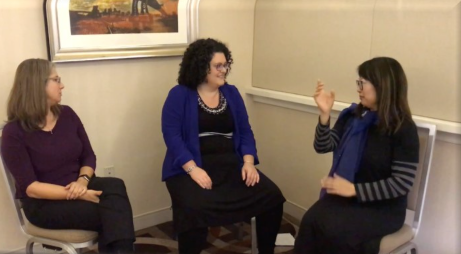}
                \caption*{AUSLAN}
            \end{subfigure} &
             \begin{subfigure}[t]{\width\textwidth}
                \centering
                \includegraphics[width=\linewidth]{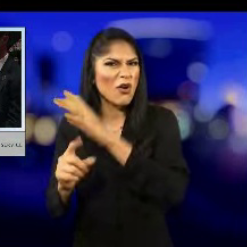}
                \caption*{OpenASL}
            \end{subfigure} &
            \begin{subfigure}[t]{\width\textwidth}
                \centering
                \includegraphics[width=\linewidth]{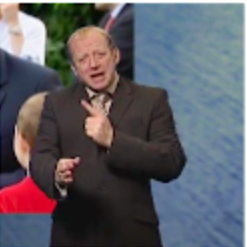}
                \caption*{BOBSL}
            \end{subfigure} &
            \begin{subfigure}[t]{\width\textwidth}
                \centering
                \includegraphics[width=\linewidth]{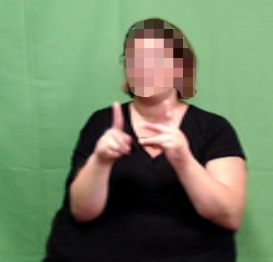}
                \caption*{How2Sign}
            \end{subfigure} &
            \begin{subfigure}[t]{\width\textwidth}
                \centering
                \includegraphics[width=\linewidth]{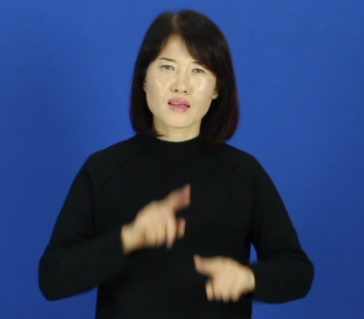}
                \caption*{KETI}
            \end{subfigure} &
            \begin{subfigure}[t]{\width\textwidth}
                \centering
                \includegraphics[width=\linewidth]{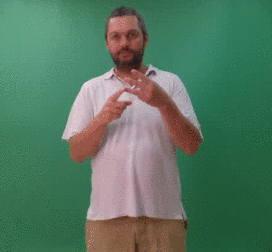}
                \caption*{GSL}
            \end{subfigure} &
            \begin{subfigure}[t]{\width\textwidth}
                \centering
                \includegraphics[trim={0 1cm 0 0},clip,width=\linewidth]{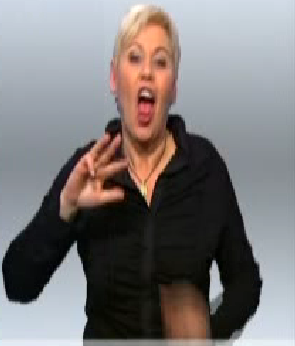}
                \caption*{PHOENIX}
            \end{subfigure} \\
            \begin{subfigure}[t]{\width\textwidth}
                \centering
                \includegraphics[trim={0 2.5cm 0 0},clip,width=\linewidth]{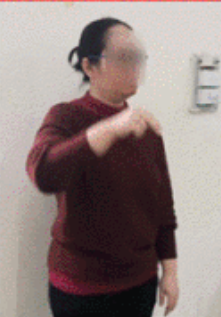}
                \caption*{Multi-VSL}
            \end{subfigure} &
            \begin{subfigure}[t]{\width\textwidth}
                \centering
                \includegraphics[width=\linewidth]{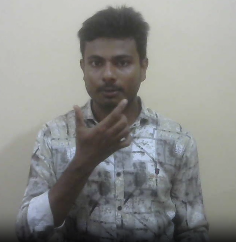}
                \caption*{BdSLW60}
            \end{subfigure} &
            \begin{subfigure}[t]{\width\textwidth}
                \centering
                \includegraphics[width=\linewidth]{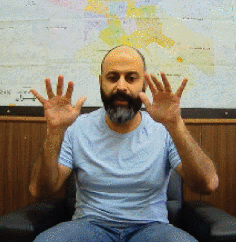}
                \caption*{ISLR101}
            \end{subfigure} &
            \begin{subfigure}[t]{\width\textwidth}
                \centering
                \includegraphics[width=\linewidth]{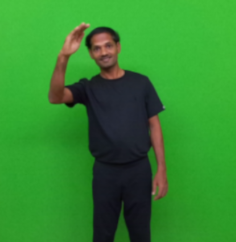}
                \caption*{FDMSE-ISL}
            \end{subfigure} &
            \begin{subfigure}[t]{\width\textwidth}
                \centering
                \includegraphics[width=\linewidth]{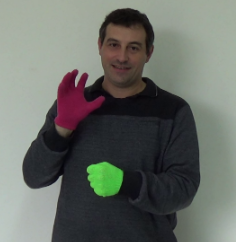}
                \caption*{LSA64}
            \end{subfigure} &
            \begin{subfigure}[t]{\width\textwidth}
                \centering
                \includegraphics[width=\linewidth]{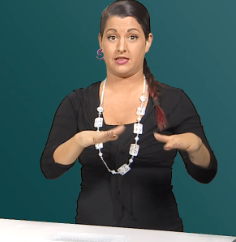}
                \caption*{WMT}
            \end{subfigure} &
            \begin{subfigure}[t]{\width\textwidth}
                \centering
                \includegraphics[width=\linewidth]{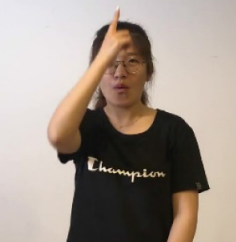}
                \caption*{NMFs-CSL}
            \end{subfigure} \\
            \begin{subfigure}[t]{\width\textwidth}
                \centering
                \includegraphics[trim={0 3cm 0 0},clip,width=\linewidth]{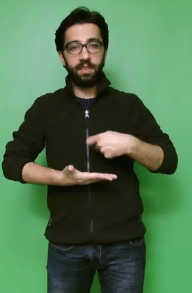}
                \caption*{\centering BosphorusSign22k}
            \end{subfigure} &
             \begin{subfigure}[t]{\width\textwidth}
                \centering
                \includegraphics[width=\linewidth]{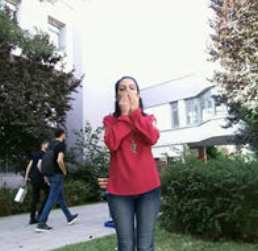}
                \caption*{AUTSL}
            \end{subfigure} &
            \begin{subfigure}[t]{\width\textwidth}
                \centering
                \includegraphics[width=\linewidth]{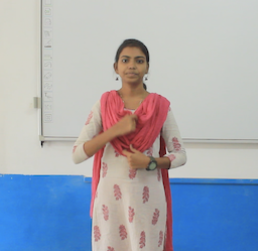}
                \caption*{INCLUDE}
            \end{subfigure} &
            \begin{subfigure}[t]{\width\textwidth}
                \centering
                \includegraphics[width=\linewidth]{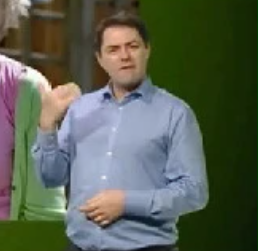}
                \caption*{BSL-1K}
            \end{subfigure} &
            \begin{subfigure}[t]{\width\textwidth}
                \centering
                \includegraphics[width=\linewidth]{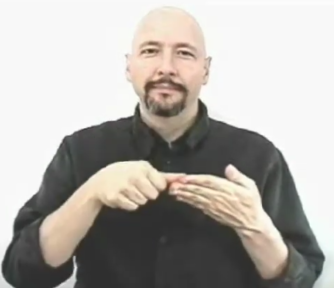}
                \caption*{MS-ASL}
            \end{subfigure} &
            \begin{subfigure}[t]{\width\textwidth}
                \centering
                \includegraphics[width=\linewidth]{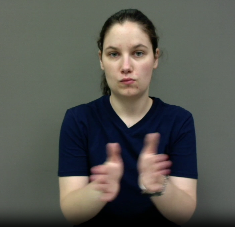}
                \caption*{WLASL}
            \end{subfigure} &
            \begin{subfigure}[t]{\width\textwidth}
                \centering
                \includegraphics[width=\linewidth]{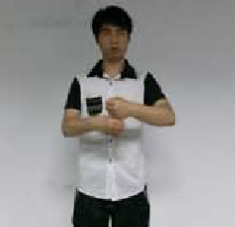}
                \caption*{CSL}
            \end{subfigure} \\
            
            \midrule
    
    
            \multicolumn{7}{c}{Semaphoric HGR} \\
            \midrule
            
            \begin{subfigure}[t]{\width\textwidth}
                \centering
                \includegraphics[width=\linewidth]{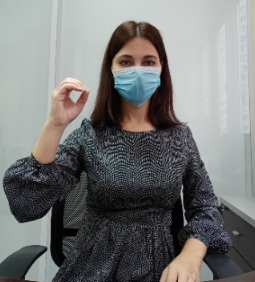}
                \caption*{HaGRID}
            \end{subfigure} &
            \begin{subfigure}[t]{\width\textwidth}
                \centering
                \includegraphics[width=\linewidth]{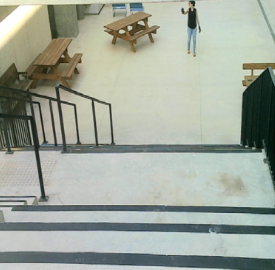}
                \caption*{URGR}
            \end{subfigure} &
            \begin{subfigure}[t]{\width\textwidth}
                \centering
                \includegraphics[width=\linewidth]{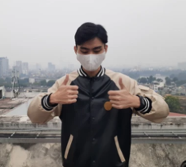}
                \caption*{SHAPE}
            \end{subfigure} &
            \begin{subfigure}[t]{\width\textwidth}
                \centering
                \includegraphics[width=\linewidth]{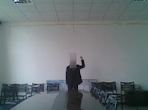}
                \caption*{LRHG}
            \end{subfigure} &
            \begin{subfigure}[t]{\width\textwidth}
                \centering
                \includegraphics[width=\linewidth]{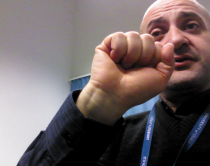}
                \caption*{OUHANDS}
            \end{subfigure} &
            \begin{subfigure}[t]{\width\textwidth}
                \centering
                \includegraphics[width=\linewidth]{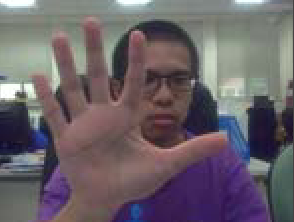}
                \caption*{LaRED}
            \end{subfigure} &
            \begin{subfigure}[t]{\width\textwidth}
                \centering
                \includegraphics[width=\linewidth]{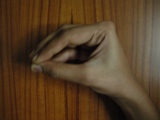}
                \caption*{NUS-II}
            \end{subfigure} \\
            
            \begin{subfigure}[t]{\width\textwidth}
                \centering
                \includegraphics[width=\linewidth]{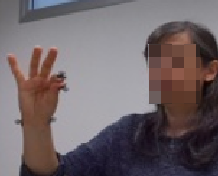}
                \caption*{\centering EHWGesture}
            \end{subfigure} &
            \begin{subfigure}[t]{\width\textwidth}
                \centering
                \includegraphics[width=\linewidth]{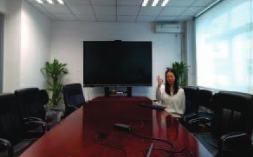}
                \caption*{LD-ConGR}
            \end{subfigure} &
            \begin{subfigure}[t]{\width\textwidth}
                \centering
                \includegraphics[width=\linewidth]{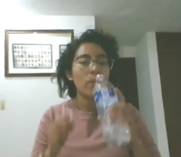}
                \caption*{IPN-Hand}
            \end{subfigure} &
            \begin{subfigure}[t]{\width\textwidth}
                \centering
                \includegraphics[width=\linewidth]{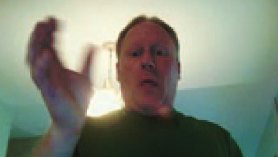}
                \caption*{Jester}
            \end{subfigure} &
            \begin{subfigure}[t]{\width\textwidth}
                \centering
                \includegraphics[width=\linewidth]{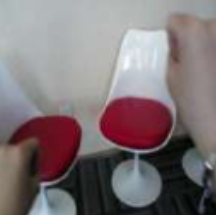}
                \caption*{EgoGesture}
            \end{subfigure} &
            \begin{subfigure}[t]{\width\textwidth}
                \centering
                \includegraphics[width=\linewidth]{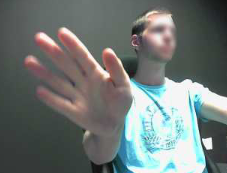}
                \caption*{NVGesture}
            \end{subfigure} &
            \begin{subfigure}[t]{\width\textwidth}
                \centering
                \includegraphics[width=\linewidth]{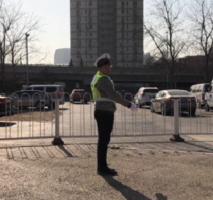}
                \caption*{Police Gesture Dataset}
            \end{subfigure} \\
    
            \bottomrule
        \end{tabular}
        \caption{Indicative RGB samples from selected representative datasets for sign language recognition and semaphoric HGR.}
        \label{fig:datasets-samples}
    \end{figure}

    \subsection{Isolated and Continuous SLR}

    The datasets that have been developed to support video-based Sign Language Recognition (SLR) can be broadly categorized into those designed for IDSLR and CSLR. The former datasets typically focus on individual signs, recorded in controlled lab environments with a fixed vocabulary, making them well-suited for classification-based models. In contrast, continuous datasets present real-world sequences of signs, often extracted from broadcast material or spontaneous signing, requiring temporal segmentation and sequence modeling approaches. Additionally, some datasets support Sign Language Translation (SLT) by providing sentence-level gloss annotations and parallel spoken language transcriptions.
    
    The RWTH-PHOENIX-Weather 2014 \citep{PHOENIX-2014} and its extended version PHOENIX-2014-T1 \citep{NEURAL-SIGN-LANGUAGE-TRANSLATION-CVPR-CAMGOZ} are benchmark datasets offering sentence-level annotations in German Sign Language (DGS) captured from television weather forecasts. A semi-automated annotation protocol was proposed, leveraging existing methods for speech recognition and accelerating the process. The signers perform gestures in a constrained environment (i.e. wearing dark clothes in front of a static gray background), probably limiting the generalization of models trained to it. On the contrary, more recent efforts such as BOBSL \citep{BOBSL-BBC} and OpenASL \citep{OpenASL} have significantly expanded the signer base and vocabulary, with hours of British (BSL) and American Sign Language (ASL) footage collected from TV and web sources, respectively. The more unconstrained environment and the increased number of signers improve the real-world performance of HGR models.
    The lab-recorded KETI \citep{KETI_MDPI} contains samples captured from two camera angles (i.e. front and side), increasing diversity. The aim of the dataset is to interpret Korean sign language gestures that are performed in emergency situations using skeleton and RGB data, reducing the need of professionals. Another multimodal dataset collected in a laboratory is the How2Sign \citep{How2Sign-CVPR} for ASL recognition and translation. The entire dataset was captured in the termed Green Screen studio with a single RGB camera and depth sensor, while a subset of it in the Panoptic studio using multiple cameras from various viewpoints, enabling accurate 3D pose estimation. Pose data and text transcripts are also considered in TVB-HKSL-News \citep{TVB-HKSL-News}. The samples were collected from TV news with subtitles and Optical Character Recognition was applied to extract the text sequence, while an off-the-shelf pose estimator was utilized to obtain the human keypoints. The RGB-based CSL-Daily \citep{Back-Translation-CVPR} also provides glosses and textual transcriptions, aiding both CSLR and SLT tasks. A sign dictionary (referred to as SignDict) was also constructed to be used for other tasks, including sign spotting and segmentation. The Auslan Daily Dataset \citep{Auslan-Daily} consists of Australian Sign Language (AUSLAN) gestures performed by humans in a television series, TV news and web videos. These samples may contain multiple persons interacting with each other, which makes the dataset suitable for in-the-wild scenarios.

    Other datasets for CSLR do not provide text translations. ChicagoFSWild+ \citep{ChicagoFSWild-plus-ICCV} was collected from web videos, including ASL signs performed by 260 signers making this dataset the largest in terms of number of subjects. HKSL \citep{SignBERT-HKSL} was captured in a laboratory environment, providing RGBD data.

    Datasets for IDSLR are larger than datasets for CSLR, as they do not require complex annotation, making data collection easier. Many of these datasets, such as CSL \citep{CSL500}, GSL iso. \citep{ADADOGLOU}, BosphorusSign22k \citep{bosphorussign22k}, AUTSL \citep{AUTSL}, KArSL \citep{KArSL} and FDMSE-ISL \citep{FDMSE-hierarchicalwindowedgraphattention}, include pose or depth modalities, enriching multimodal learning. All of them, except for the AUTSL, were collected in a laboratory with fixed background. Datasets that do not include such modalities can be extended to them by using off-the-shelf DL pose and depth estimators. The ASL Dataset \citep{TAXONOMY_OF_HAND_GESTURES_2000_ASL_DATASET} is the only one consisting of pose sequences exclusively. MS-ASL \citep{MS-ASL} and WLASL \citep{WLASL} are two RGB-based datasets for isolated ASL recognition collected from publicly available web videos of the deaf community, including educational resources and ASL tutorials and performed by 222 and 119 subjects respectively, increasing the diversity. However, they contain relatively few samples compared to large-scale datasets such as BSL-1K \citep{bsl-1K}, CSLD \citep{tim-slr_csld_dataseta} and NCSL \citep{2plus1-D-SLR_and_NCSL}. For isolated Irish Sign Language (ISL) recognition, only laboratory datasets have been proposed (e.g. INCLUDE \citep{INCLUDE}, IISL2020 \citep{IISL2020_MDPI} and FDMSE-ISL \citep{FDMSE-hierarchicalwindowedgraphattention}), making the collection of an in-the-wild dataset necessary. For Vietnamese SLR, authors in \citep{Multi-VSL-dinh2025sign} collected a large-scale multi-view dataset, considering three different points of view, enhancing diversity. Other sign languages have also been considered including Argentinian (e.g. LSA64 \citep{Lsa64argentiniansignlanguage}), Persian (e.g. RKS-PERSIANSIGN \citep{persian_sl_dataset}) and  Iranian (e.g. ISLR101 \citep{ISLR101-IRANIAN}). The rarity of some sign languages, such as those mentioned above, poses obstacles to finding relevant annotators, exacerbating the data scarcity challenge, as it will be discussed later.

    \begin{table}[h!]
        \centering
        \tiny
        \setlength{\tabcolsep}{1pt}
        \renewcommand{\arraystretch}{1.2}
        \caption{Overview of the characteristics of each dataset related to SLU. The sign vocabulary (Sign Voc.) corresponds to the number of distinct classes included in each dataset.}
        \begin{tabular}{*{10}{l}
        }
            \toprule
            \textbf{Name}&\textbf{Year}&\textbf{Task}&\textbf{Modality}&\textbf{Samples}&\textbf{Signers}&\textbf{Sign Voc.}&\textbf{Language}&\textbf{Duration}&\textbf{Source}\\
            \midrule
            TVB-HKSL-News \citep{TVB-HKSL-News}&2024&CSLR/SLT&RGB, Pose&7000&2&6515&HKSL&16.07h&TV\\
            Auslan-Daily \citep{Auslan-Daily}&2023&CSLR/SLT&RGB&25000&67&-&AUSLAN&-&TV, Web\\
            OpenASL \citep{OpenASL}&2022&CSLR/SLT&RGB&98417&220&-&ASL&288h&Web\\
            CSL-Daily \citep{Back-Translation-CVPR}&2021&CSLR/SLT&RGB&20654&10&2000&CSL&23.27h&Lab\\
            BOBSL \citep{BOBSL-BBC}&2021&CSLR/SLT&RGB&1962&39&2281&BSL&1467h&TV\\
            How2Sign (green studio) \citep{How2Sign-CVPR}&2021&CSLR/SLT&RGBD, Pose&2529&11&-&ASL&79.12h&Lab\\
            Corpus FinSL \citep{Corpus-Finnish}&2020&CSLR/SLT&RGB&343&21&-&FinSL&14.37h&-\\
            KETI \citep{KETI_MDPI}&2019&CSLR/SLT&RGB, Pose&14672&14&524&KSL&27.99h&Lab\\
            GSL SD/SI \citep{ADADOGLOU}&2019&CSLR/SLT&RGBD&10295&7&310&GSL&9.59h&Lab\\
            PHOENIX-2014-T1 \citep{NEURAL-SIGN-LANGUAGE-TRANSLATION-CVPR-CAMGOZ}&2018&CSLR/SLT&RGB&8257&9&1066&DGS&-&TV\\
            PHOENIX-2014 \citep{PHOENIX-2014}&2014&CSLR/SLT&RGB&-&9&1558&DGS&10.73h&TV\\
            \midrule
            HKSL \citep{SignBERT-HKSL}&2021&CSLR&RGBD&2400&6&55&HKSL&-&Lab\\
            ChicagoFSWild+ \citep{ChicagoFSWild-plus-ICCV}&2019&CSLR&RGB&55232&260&26&ASL&-&Web\\
            \midrule
            Multi-VSL \citep{Multi-VSL-dinh2025sign}&2025&IDSLR&RGB, Pose&84764&30&1000&Vietnamese&-&Lab\\
            BdSLW60 \citep{BdSLW60-rubaiyeat2025bdslw60}&2025&IDSLR&RGB&9307&18&60&Bangla&-&Lab\\
            ISLR101 \citep{ISLR101-IRANIAN}&2025&IDSLR&RGB, Pose&4614&10&101&Iranian&-&-\\
            ISW-1000 \citep{ISW-1000-ren2025multi}&2025&IDSLR&RGB, Pose, OF&10000&10&1000&CSL&-&Lab\\
            FDMSE-ISL \citep{FDMSE-hierarchicalwindowedgraphattention}&2024&IDSLR&RGBD&40033&20&2002&ISL&36h&Lab\\
            LSA64 \citep{Lsa64argentiniansignlanguage}&2023&IDSLR&RGB&3200&10&64&LSA&1.9h&Lab\\
            Signsuisse Lexicon~\citep{SIGN-muller-etal-2023-findings}&2023&IDSLR&RGB&18221&-&-&3 (Mult)&-&Web\\
            CSLD \citep{tim-slr_csld_dataseta}&2022&IDSLR&RGB&120000&30&400&CSL&66.8h&Lab\\
            NCSL \citep{2plus1-D-SLR_and_NCSL}&2022&IDSLR&RGB&90000&30&300&CSL&50.1h&Lab\\
            KArSL \citep{KArSL}&2021&IDSLR&RGBD, Pose&75300&3&502&ArSL&-&Lab\\
            NMFs-CSL \citep{NMFs-CSL}&2021&IDSLR&RGB&32010&10&1.067&CSL&24.4h&Lab\\
            BosphorusSign22k \citep{bosphorussign22k}&2020&IDSLR&RGBD, Pose&22542&6&744&TSL&19h&Lab\\
            AUTSL \citep{AUTSL}&2020&IDSLR&RGBD, Pose&38336&43&226&TSL&21h&Lab\\
            INCLUDE \citep{INCLUDE}&2020&IDSLR&RGB&4287&7&263&ISL&3h&Lab\\
            IISL2020 \citep{IISL2020_MDPI}&2020&IDSLR&RGB&12100&16&11&ISL&7h&Lab\\
            RKS-PERSIANSIGN \citep{persian_sl_dataset}&2020&IDSLR&RGB&10000&10&100&Persian&-&Lab\\
            BSL-1K \citep{bsl-1K}&2020&IDSLR&RGB&280000&40&1064&BSL&1060h&TV\\
            MSASL \citep{MS-ASL}&2019&IDSLR&RGB&25513&222&1000&ASL&24.65h&Web\\
            WLASL \citep{WLASL}&2019&IDSLR&RGB&21083&119&2000&ASL&14.11h&Web\\
            GSL isol. \citep{ADADOGLOU}&2019&IDSLR&RGBD&40785&7&310&GSL&6.44h&Lab\\
            SMILE \citep{SMILE}&2018&IDSLR&RGB&-&30&100&DSGS&-&Lab\\
            ASL Dataset \citep{TAXONOMY_OF_HAND_GESTURES_2000_ASL_DATASET}&2018&IDSLR&Pose&1200&20&30&ASL&-&Lab\\
            CSL \citep{CSL500}&2018&IDSLR&RGBD, Pose&125000&50&500&CSL&-&Lab\\
            \bottomrule
        \end{tabular}
        \label{tab:slu-datasets-both}
    \end{table}
    
    \subsection{Semaphoric HGR}

    Semaphoric HGR research is supported by a diverse range of datasets categorized into three main types based on task formulation: SHGR, IDHGR and CDHGR. These datasets vary in their application domains (e.g., HCI, HMI, HRI), modalities (e.g., RGB, depth, pose, infrared), and in the number of gesture classes, subjects, and gesture samples they include. The statistics of datasets for semaphoric HGR are provided in tables \ref{tab:static_datasets},\ref{tab:isolated_dynamic_datasets} and \ref{tab:continuous_datasets}, regarding the task for which they are used (i.e. static, isolated dynamic and continuous HGR, respectively).

    SHGR datasets target gesture recognition from single frames, making them easier to process due to the lack of temporal complexity. Older datasets like HGR-1 \citep{HGR-1-2024-PATTER-RECOGNITION}, NUS-I \citep{NUS-I-2010}, NUS-II \citep{NUS-II-2013}, and OUHANDS \citep{ouhands} contain few samples, limiting deep learning training, yet remain benchmarks for HCI due to their simplicity. LaRED \citep{LaRED} offers 243K RGBD samples across 81 gestures, but only from 10 subjects, restricting subject-independent recognition. HaGRID \citep{WACV_kapitanov2024hagrid} is much larger and diverse, with nearly twice as many full-resolution RGB frames from ~35K subjects in varied scenes, including bounding boxes for hand detection. HaGRIDv2 \citep{HAGRIDv2_ARXIV} expands this with 15 additional classes, ~66K subjects, and over 1M samples. SHAPE \citep{SHAPE_DATASET} has a similar number of gestures (all two-handed) but fewer samples. LRHG \citep{LONG_RANGE_HGR_SSD} and MD-UHGRD \citep{MD-UHGRD_SA-YOLO} target UAV-based gesture recognition, with MD-UHGRD offering more samples but fewer classes, captured from varying distances and angles to enhance viewpoint robustness. URGR \citep{ULTRA-RANGE-UGV-Bamani} focuses on ultra-range gestures for UGVs, with 6 robot-command gestures performed by 16 signers at distances from 2 to 25 meters.
    
    IDHGR datasets contain gestures performed in isolation, with start and end points explicitly segmented. Most of the existing datasets (e.g. SKIG \citep{SKIG}, NVIDIA (aka nvGesture) \citep{NVIDIA_GESTURE_R3DCNN_Molchanov}, DHG \citep{DHG-CVPRW} and SHREC17 \citep{SHREC}) for IDHGR are relatively small with a few thousand samples and support multimodal settings by providing depth maps, IR frames, optical flow or skeleton joints, in addition to the RGB modality, aiming to enhance the accuracy of the models. Jester-20bn \citep{Jester-Dataset}, proposed for gesture-based HCI, is the largest dataset for (non-SLU) IDHGR containing \~150K videos performed by 1376 subjects. The remarkable diversity and size of this dataset make it ideal for pretraining DL models. Other datasets for HCI include IPN-Hand \citep{IPN-HAND}, ZJUGesture \citep{ZJUGesture} and DATE \citep{DATE}, comprising 4218, 9892 and 13500 samples respectively. LD-ConGR \citep{LD-ConGR}, ChaLearn IsoGD \citep{chalearn-lap} and EgoGesture \citep{EgoGesture_DATASET} contain a sufficient amount of data, constituting adequate baselines for IDHGR. EgoDriving \citep{EgoFormer_METHOD_EgoDriving_DATASET}, Briareo \citep{Briareo_dataset} and nvGesture \citep{NVIDIA_GESTURE_R3DCNN_Molchanov} have been proposed for human-car interaction, mainly in a multimodal setting, including gestures that correspond to car functionalities (e.g. CW rotation and audio manipulation\footnote{https://youtu.be/NJmk1DUyyB8}). TCG \citep{TCG} was also proposed for human-car interaction, but for recognizing traffic police gestures in autonomous driving scenarios relying only on pose data. EHWGesture \citep{EHWGesture-amprimo2025ehwgesture} is the first dataset for recognizing clinical gestures and providing annotations for many other tasks, including gesture detection \& triggering.
    
    CDHGR datasets challenge models to identify and segment gestures from continuous video streams. Many of the datasets mentioned earlier (e.g. ChaLearn \citep{chalearn-lap}, IPN-Hand \citep{IPN-HAND}, LD-ConGR \citep{LD-ConGR}, EgoGesture \citep{EgoGesture_DATASET} and ZJUGesture \citep{ZJUGesture}) have been obtained by annotating the temporal boundaries of gestures within continuous sequences, thus these datasets have protocols for both IDHGR and CDHGR. Therefore, we will only discuss the Police Gesture Dataset \citep{Police-Gesture-Dataset} as the previous analysis could also be applied here. This dataset was introduced for recognizing police traffic gestures, similarly with TCG \citep{TCG}, containing a small number of samples (21 RGB videos) and 3354 gesture instances.
    
    Each dataset supports specific sub-tasks within HGR research: static datasets aid in gesture classification, isolated dynamic datasets support temporal gesture modeling and classification, and continuous datasets allow for joint gesture spotting and classification-crucial for real-time and interactive applications. This diversity in dataset structure, scale, and modality reflects the evolving demands and challenges of semaphoric gesture recognition across interactive domains. 

    \begin{table}[h]
        \centering
        \footnotesize
        \setlength{\tabcolsep}{1pt}
        \renewcommand{\arraystretch}{1.5}
        \caption{Datasets for semaphoric SHGR.}
        \begin{tabular}{*{9}{l}}
        \toprule
        \textbf{Name}&\textbf{Year}&\textbf{Domain}&\textbf{Modality}&\textbf{Samples}&\textbf{Subjects}&\textbf{Classes}&\textbf{Resolution}&\textbf{Distance}\\
        \midrule
            HaGRID~\citep{WACV_kapitanov2024hagrid}&2024&HCI&RGB&552992&34730&19&1920x1080&0.5-4m\\
            HaGRIDv2~\citep{HAGRIDv2_ARXIV}&2024&HCI&RGB&1086158&65977&34&1920x1080&0.5-4m\\
            MD-UHGRD~\citep{MD-UHGRD_SA-YOLO}&2024&HRI&RGB&20000&-&19&-&3-10m\\
            URGR~\citep{ULTRA-RANGE-UGV-Bamani}&2024&HRI&RGB&347483&16&6&640x480&1-25m\\
            SHAPE~\citep{SHAPE_DATASET}&2022&-&RGB&34226&20&32&4128x3096&2-5m\\
            LRHG~\citep{LONG_RANGE_HGR_SSD}&2021&HRI&RGB&4320&8&10&640x480&1-7m\\
            OUHANDS~\citep{ouhands}&2016&HCI&RGBD&3150&23&10&640x480&$\leq$ 1m\\
            HGR-1~\citep{HGR-1-2024-PATTER-RECOGNITION}&2014&-&RGB&899&12&27&174x131-640x480&$\leq$ 1m\\
            LaRED~\citep{LaRED}&2014&-&RGBD&243000&10&81&640x480&-\\
            NUS-II~\citep{NUS-II-2013}&2013&-&RGB&2000&40&10&160x120&$\leq$ 1m\\
            ASL-FS~\citep{ASL-FS-2011-CVPRW}&2011&-&RGBD&48K&4&24&-&-\\
            NUS-I~\citep{NUS-I-2010}&2010&-&RGB&240&-&10&160x120&$\leq$ 1m\\
        \bottomrule
        \end{tabular}
        \label{tab:static_datasets}
    \end{table}
    
    \begin{table}[h]
        \centering
        \footnotesize
        \setlength{\tabcolsep}{1.2pt}
        \renewcommand{\arraystretch}{1.5}
        \caption{Datasets for semaphoric IDHGR.}
        \begin{tabular}{llllllll}
        \toprule
        \textbf{Name}&\textbf{Year}&\textbf{Domain}&\textbf{Modality}&\textbf{Samples}&\textbf{Subjects}&\textbf{Classes}&\textbf{Frames}\\\midrule
            EHWGesture \citep{EHWGesture-amprimo2025ehwgesture}&2025&-&RGBD&1100&25&11&3.3M\\
            EgoDriving~\citep{EgoFormer_METHOD_EgoDriving_DATASET}&2024&HMI&RGB&400&-&29&70276\\
            DATE~\citep{DATE}&2024&HCI&RGB&13500&22&27&-\\
            ZJUGesture (isol.)~\citep{ZJUGesture}&2023&HCI&RGB&9892&60&9&-\\
            LD-ConGR (isol.)~\citep{LD-ConGR}&2022&HCI&RGBD&44887&30&10&-\\
            IPN Hand (isol.)~\citep{IPN-HAND}&2021&HCI&RGB&4218&50&13&-\\
            TCG~\citep{TCG}&2020&HMI&Pose&250&5&4&839350\\
            Jester~\citep{Jester-Dataset}&2019&HCI&RGB&148092&1376&27&5331312\\
            Briareo~\citep{Briareo_dataset}&2019&HMI&RGBD, Pose, IR&1440&40&12&-\\
            EgoGesture (isol.)~\citep{EgoGesture_DATASET}&2018&HCI&RGBD&24161&50&83&2953224\\
            SHREC17-14/28~\citep{SHREC}&2017&-&Depth, Pose&2800&28&14/28&-\\
            nvGesture~\citep{NVIDIA_GESTURE_R3DCNN_Molchanov}&2016&HMI&RGBD, IR, OF&1532&20&25&-\\
            ChaLearn IsoGD~\citep{chalearn-lap}&2016&-&RGBD&47933&21&249&-\\
            DHG-14/28~\citep{DHG-CVPRW}&2016&-&Depth, Pose&2800&20&14/28&-\\
            SKIG~\citep{SKIG}&2013&-&RGBD&1080&6&10&-\\
            \bottomrule
        \end{tabular}
        \label{tab:isolated_dynamic_datasets}
    \end{table}
        
    \begin{table}[h!]
        \centering
        \footnotesize
        \setlength{\tabcolsep}{1.2pt}
        \renewcommand{\arraystretch}{1.5}
        \caption{Datasets for semaphoric CDHGR.}
        \begin{tabular}{llllllll}
        \toprule        
        \textbf{Name}&\textbf{Year}&\textbf{Domain}&\textbf{Modality}&\textbf{Samples}&\textbf{Instances}&\textbf{Subjects}&\textbf{Classes}
        \\
        \midrule
            ZJUGesture (cont.)~\citep{ZJUGesture}&2023&HCI&RGB&-&9892&60&9\\
            LD-ConGR (cont.)~\citep{LD-ConGR}&2022&HCI&RGBD&542&44887&30&10\\
            IPN Hand (cont.)~\citep{IPN-HAND}&2021&HCI&RGB&200&4218&50&13\\
            Police Gesture Dataset~\citep{Police-Gesture-Dataset}&2020&HMI&RGB&21&3354&-&8\\
            EgoGesture (cont.)~\citep{EgoGesture_DATASET}&2018&HCI&RGBD&2081&24161&50&83\\
            ChaLearn ConGD~\citep{chalearn-lap}&2016&-&RGBD&22535&47933&21&249\\
            \bottomrule
        \end{tabular}
        \label{tab:continuous_datasets}
    \end{table}

    \subsection{Other}
    
    In addition to standard semaphoric gesture datasets, several specialized datasets have been introduced to support research on more specific gesture types. The First-Person Hand Action (FPHA) dataset \citep{FPHA}, published in 2018, contains 1,175 video sequences captured from a first-person perspective by six participants. It focuses on isolated manipulative hand gestures across 45 action classes, offering RGB, depth, and skeleton data, and is particularly valuable for studies involving egocentric interaction.
    Another notable example is the DP Dataset \citep{DeePoint}, released in 2023, which, to the best of our knowledge, is the only publicly available benchmark specifically designed for pointing gesture recognition. Collected from 33 participants in two indoor environments, the dataset comprises approximately 2.8 million frames and 6,335 annotated gesture instances. It includes detailed annotations such as gesture timings and pointing directions, enabling both binary gesture classification (gesture vs. non-gesture) and directional estimation tasks.
    Finally, the SocialGesture \citep{SocialGesture-CVPR2025-cao2025socialgesture} focuses on the recognition of natural gestures in multi-person environments. It comprises 9,889 in-the-wild videos containing 42,533 instances and provides fine-grained annotations including bounding boxes, relations (for example, a subject is pointing to a target) and Visual Question Answering (VQA) details.

    \subsection{Datasets Comparison}

    After examining the existing datasets for VHGR with significant impact on the current research, in this subsection we aim to determine pros and cons of the most important ones as well as the use cases that they fit. For gesture-based HCI, HaGRID \citep{WACV_kapitanov2024hagrid} and its versions \citep{HAGRIDv2_ARXIV} seem to be a reliable solution, containing an enormous number of samples corresponding to 19 distinct classes (that can be mapped to computer commands). Given that these gestures are also static, even simple architectures - like the ones discussed in Sections \ref{sec:static} \& \ref{sec:comparison} - can achieve sufficient performance. Furthermore, this dataset could also be used for pretraining, enhancing the backbone's expression capability. However, its size implies that training takes a long time, which constitutes a deterrent in some cases, compared to other datasets for HCI like OUHANDS \citep{ouhands}. For gesture-based HRI, URGR \citep{ULTRA-RANGE-UGV-Bamani} is the only one for UGV guidance and considers various recognition ranges, which may be crucial for real-world scenarios. Even though it contains a large number of samples, the small number of classes may limit the commands given by gestures and, therefore, the functionalities of the entire gesture-based interface. Regarding SLU datasets, for ASL recognition, OpenASL \citep{OpenASL} is the largest one, including almost 100K RGB videos (288h in total) collected from the web. MSASL \citep{MS-ASL} comprises almost the same number of subjects, but it is much smaller. Furthermore, OpenASL was proposed for CSLR whereas MSASL was proposed for IDSLR, thus it may be more suitable for real-world use cases, where the temporal boundaries are not known. BSL-1K \citep{bsl-1K} is also a very large dataset, proposed for British IDSLR and comprising the largest number of samples (280K, 1060h in total). BOBSL \citep{BOBSL-BBC} was proposed for British CSLR and comprises videos of 1467h in total making it the longest dataset in terms of duration. Phoenix \citep{PHOENIX-2014} and Phoenix T \citep{NEURAL-SIGN-LANGUAGE-TRANSLATION-CVPR-CAMGOZ} remain the most suitable datasets for German sign language, covering a large sign vocabulary. Nevertheless, the small number of samples and the low diversity - due to the limited number of signers and the constrained environment - make the collection of a new, larger and more diverse dataset for DGS. For Chinese SLR, NMFs-CSL \citep{NMFs-CSL} is significantly smaller than CSLD \citep{tim-slr_csld_dataseta}, CSL \citep{CSL500} and NCSL \citep{2plus1-D-SLR_and_NCSL} but includes much more signs (1064 instead of less than 500). CSL-Daily \citep{Back-Translation-CVPR} remains the only one for Chinese CSLR, covering a large sign vocabulary of 2000 signs in total.

\section{Evaluation Metrics}
\label{sec:metrics}

    In this section, we present the key evaluation metrics adopted in the literature for measuring the performance of HGR methods. Broadly, metrics fall into two main categories: \textit{classification metrics} (used in static and isolated dynamic HGR) and \textit{sequence-to-sequence metrics} (used in continuous dynamic HGR). Additionally, other indicators such as computational resource usage and overfitting-related metrics are also considered.
    
    \subsection{Classification Metrics}

    In multi-class HGR problems-especially when dealing with class imbalance-researchers often report both \textbf{micro-averaged} (per-instance) and \textbf{macro-averaged} (per-class) versions of common classification metrics. Micro-averaged metrics aggregate contributions from all instances across classes, thus being sensitive to class imbalance. Macro-averaged metrics compute the metric independently for each class and then average, treating all classes equally regardless of their frequency. The formulae for each metric can be found in Table \ref{tab:macro-micro-metrics}.

    \begin{table}[]
        \centering
        \setlength{\tabcolsep}{3pt}
        \renewcommand{\arraystretch}{1.5}
        \caption{Commonly utilized metrics for the evaluation of hand gesture classification. Here, $TP_i$, $FP_i$, $TN_i$, and $FN_i$ denote the number of true positives, false positives, true negatives, and false negatives for the $i$-th class, respectively, and $N$ is the total number of classes.}
        \begin{tabular}{ccc}
            \toprule
            \textbf{Metrics} & \textbf{Micro-averaged formulae} & \textbf{Macro-averaged formulae} \\
            \midrule
            Accuracy & $\frac{\sum_{i=1}^{N}(TP_i + TN_i)}{\sum_{i=1}^{N}(TP_i + TN_i + FP_i + FN_i)}$ & $\frac{1}{N} \sum_{i=1}^{N} \frac{TP_i + TN_i}{TP_i + FP_i + TN_i + FN_i}$ \\
            Precision & $\frac{\sum_{i=1}^{N}TP_i}{\sum_{i=1}^{N}(TP_i + FP_i)}$ & $\frac{1}{N} \sum_{i=1}^{N} \frac{TP_i}{TP_i + FP_i}$ \\
            Recall & $\frac{\sum_{i=1}^{N}TP_i}{\sum_{i=1}^{N}(TP_i + FN_i)}$ & $\frac{1}{N} \sum_{i=1}^{N} \frac{TP_i}{TP_i + FN_i}$ \\
            Jaccard-Index & $\frac{\sum_{i=1}^{N}TP_i}{\sum_{i=1}^{N}(TP_i + FP_i + FN_i)}$ & $\frac{1}{N} \sum_{i=1}^{N} \frac{TP_i}{TP_i + FP_i + FN_i}$ \\
            \midrule
            F1-Score & \multicolumn{2}{c}{$2 \cdot \frac{\text{Precision} \cdot \text{Recall}}{\text{Precision} + \text{Recall}}$} \\
            \bottomrule
        \end{tabular}
        \label{tab:macro-micro-metrics}
    \end{table}
    
    \subsection{Sequence-to-Sequence Metrics}

    For continuous HGR (CDHGR), particularly in Continuous Sign Language Recognition (CSLR), sequence-level metrics are used.
    \textbf{Word Error Rate (WER)}~\citep{VAC-CSLR,ADADOGLOU} is the standard metric, measuring the minimum number of substitutions ($S$), deletions ($D$), and insertions ($I$) needed to transform the predicted sequence into the ground truth sequence of length $N$:
    \begin{equation}
        WER = \frac{S + D + I}{N}
    \end{equation}
    
    \textbf{Edit Distance Accuracy}~\citep{FA-STGCN,CTCX_WANG2022123} is a complementary metric representing the ratio of correctly predicted elements in the sequence:
    \begin{equation}
        \text{Edit-Accuracy} = \frac{N - S - D - I}{N}
    \end{equation}

    These metrics provide insight not only into classification accuracy but also into temporal and sequential consistency.

    \subsection{Computational Resource Metrics}
    
    In addition to recognition performance, computational efficiency is often evaluated, especially for real-time or embedded HGR applications. Common metrics include:
    
    \begin{itemize}[leftmargin=*]
        \item \textbf{Inference Time:} Time (in ms or seconds) taken to process one input sample.
        \item \textbf{Frames Per Second (FPS):} Number of frames processed per second; higher values indicate faster processing.
        \item \textbf{Model Size in Megabytes (MBs):} The total memory footprint of the model.
        \item \textbf{Number of Parameters:} The number of trainable parameters.
        \item \textbf{Floating Point Operations (FLOPs):} Measures the number of numerical operations required to run a model (either during training or inference time). Being independent of the hardware utilized, it can be considered - along with the number of parameters - more reliable.
    \end{itemize}
    
    These metrics are crucial when deploying HGR systems on edge devices or in latency-sensitive environments.

\section{Current Challenges and Future Research Directions}
\label{sec:challenges}

    As mentioned in the Methods sections (Sections \ref{sec:static}, \ref{sec:isolated}, and \ref{sec:continuous}), researchers have developed sophisticated techniques and large-scale datasets to achieve satisfactory performance. However, as shown in the relevant tables (Tables \ref{tab:static-accuracy}, \ref{tab:idhgr_quantitative}, \ref{tab:cslr_performance}), recognition accuracy remains low in many cases---particularly in more complex tasks such as CSLR. In this section, we discuss the current challenges and limitations that hinder efficient VHGR, posing ongoing concerns for the research community and shaping future research directions. Additionally, we map the main bottlenecks to existing mitigation strategies and emerging research opportunities, aiming to provide a more forward-looking research route for the field. These observations address research question Q4 (see Section \ref{sec:intro}).

    \subsection{Gesture-Irrelevant Factors}

        \subsubsection{Current Challenges}
            
            The presence of gesture-irrelevant factors (or redundant information) hinders accurate recognition in diverse real-world scenarios. Some of these factors include the following~\citep{EDenseNet,Mohammad_Mahmudul_static_hgr,2DPSTPP,ULTRA-RANGE-UGV-Bamani,EnGesto,RAAR3DNet}:
            \begin{itemize}[leftmargin=*]
                \item \textbf{Environmental factors:} Cluttered background, varying illumination conditions, and self-occlusions make it difficult or even impossible to automatically spot the hand region, which is an important step for efficient HGR.
                \item \textbf{Subject characteristics:} Subject independence --i.e., the ability to correctly identify a gesture regardless of the signer's characteristics such as skin color, hand shape variations, or clothing---is a key prerequisite for achieving sufficient generalization and enabling real-world deployment.
                \item \textbf{Variations in perception device position:} The distance from the camera and the viewing angle may reduce the recognition performance.
                \item \textbf{Input data quality:} Low resolution and blurriness may corrupt the discriminative characteristics of the hand gesture region. 
            \end{itemize}

        \subsubsection{Mitigation Strategies \& Future Directions}
        
            Aiming to address these factors, researchers could apply various methods to eliminate their interference:
            \begin{itemize}[leftmargin=*]
                \item \textbf{Hand cropping:} Authors often deploy palm segmentation to isolate the hand region from the background \citep{FGDSNet,tuan_linh_pham_dang2023lightweight,DRCAM_TIM}.
                \item \textbf{Spatial normalization:} Object detectors are applied to strengthen scale invariance, especially in long range recognition scenarios \citep{ULTRA-RANGE-UGV-Bamani,SlowFast-Transformer,Mohammad_Mahmudul_static_hgr}.
                \item \textbf{Super resolution:} Super resolution methods can enhance image quality \citep{ULTRA-RANGE-UGV-Bamani} to reduce the impact of low resolution and motion blur.
                \item \textbf{Attention mechanisms:} They force the visual encoders to focus on the most informative regions \citep{RAAR3DNet}.
                \item \textbf{Other:} Information-theoretic losses~\citep{Yunan_Li} and feature selection techniques~\citep{SeST} have been proposed to reduce information redundancy.
            \end{itemize}
                    
            However, many of these approaches rely on computationally expensive off-the-shelf models (e.g. segmentation models and object detectors), which increase the complexity of the overall HGR pipelines and set an upper bound to their accuracy. Further, the efficacy of custom attention modules relies on the available data. Therefore, mitigating these challenges still remains a promising field for future work; the following could be considered as potential solutions:
            \begin{itemize}[leftmargin=*]
                \item \textbf{Explainable Artificial Intelligence (XAI):} XAI \citep{Rodis2024Access} could be adopted to better understand the behavior of the networks. For instance, Grad-CAM could be further utilized to determine the image region that most affects the predicted gesture class (as has already been done by a few researchers: \citep{CorrNet,SWIN_MSTP_Neurocomputing_2025_CSLR}). Such results may indicate the effectiveness of attention mechanisms or other related modules embedded in the network in emphasizing the most important parts of the input image. t-SNE visualizations could also be leveraged to evaluate the quality of the extracted features, and in particular, how much they are influenced by the presence of redundant information. Thus, guided by XAI tools or similar frameworks, the interference of gesture-irrelevant factors could be mitigated through the careful design of the future architectures.
                \item \textbf{Specialized data augmentation and collection:} Researchers could consider applying data augmentation methods specifically to mitigate these issues. Further, the data collection should be driven by the edge cases; it is essential to include many environments, distances, viewing angles and image qualities, to make the models invariant and robust to these conditions. In other words, the data should be captured in the wild. For example, authors could explore the potential of using internet data.
            \end{itemize}

    \subsection{Computational Cost}

        \subsubsection{Current Challenges}
    
            Even lightweight methods for SHGR and IDHGR achieve high accuracy; however, for the more complex task of CDHGR, the cost is a critical factor and can generally be decomposed into two parts:
            \begin{itemize}[leftmargin=*]
                \item \textbf{Floating Point Operations (FLOPs):}
                The RGB-based CSLR methods listed in Table~\ref{tab:cslr_performance}, which generally outperform pose-only methods on the reported benchmarks, require more than 366.3 GFLOPs (\citep{DSTEN-CSLR-yu2025dsten}) and reach up to 2950.5 GFLOPs (\citep{TemporalLiftPooling}). This issue prevents not only the deployment of such methods but also the efficient training, which takes longer. Pose-based methods \citep{CoSign} are significantly lighter, but the pose estimation cost is not taken into account.
                \item \textbf{Number of parameters:} The number of parameters does not change dramatically between IDHGR and CDHGR methods, while many SHGR methods require more parameters (e.g. Gesture-CNN \citep{G-CNN-SHARMA2021115657} has 67.25M parameters). Yet, recent approaches use large-scale generative models (e.g. Uni-Sign \cite{UNISIGNli2025} leverages an LLM and comprises 592.1M parameters in total).
            \end{itemize}
            The need for large computational resources prevents not only the real-time and commercial deployment of VHGR systems, but also poses barriers to the evolution of the field, as the conducted experiments take many hours or even days to complete.

        \subsubsection{Mitigation Strategies and Future Directions}

            To reduce the computational complexity of the models, researchers could consider the following:
            \begin{itemize}[leftmargin=*]
                \item \textbf{Non-parametric layers:} Zero-parameter layers can reduce the computational cost without affecting the capacity, as in \citep{GestFormer} and \cite{tim-slr_csld_dataseta}, who replaced attention with pooling layers and introduced a non-parametric temporal modeling layer, respectively.
                \item \textbf{Smaller architectures:} Minimizing the network size while maintaining accuracy has been studied by some researchers \citep{LHGR-Net}.
                \item \textbf{Model compression:} DL model compression \citep{DL_MODEL_COMPRESSION_marino2023deep} has not yet been applied to VHGR methods despite rapid evolution. Therefore, relevant techniques such as pruning (i.e., removing less important weights) and quantization (i.e., reducing the arithmetic precision of the parameters), may constitute future research.
            \end{itemize}

    \subsection{Data Scarcity}

        \subsubsection{Current Challenges}

            The inadequacy of most of the existing datasets can be broken down as:
            \begin{itemize}[leftmargin=*]
                \item \textbf{Limited diversity:} Even though there are datasets that contain a sufficient number of samples, they lack diversity. In particular, among SHGR, only the versions of HaGRID~\citep{WACV_kapitanov2024hagrid,HAGRIDv2_ARXIV} have been collected from more than one hundred participants, although some others also contain a plausible number of samples~\citep{LaRED,ULTRA-RANGE-UGV-Bamani}. A similar situation is observed in IDHGR datasets, where only the Jester dataset~\citep{Jester-Dataset} includes data from over one hundred subjects.
                \item \textbf{Laboratory samples:} Datasets are still often captured in constrained environments \citep{ADADOGLOU,KETI_MDPI,FDMSE-hierarchicalwindowedgraphattention,Lsa64argentiniansignlanguage,bosphorussign22k}, potentially degrading in-the-wild recognition performance.
                \item \textbf{Data imbalance:} The number of samples for specific modalities, such as IR data, is limited compared with the size of RGB-based datasets \citep{IR_data_imbalance_mobilenet}. Further, most of the datasets for SLU consider more widespread sign languages, so there are only a few datasets for rare ones.
            \end{itemize}
            The lack of large-scale, diverse datasets leads to overfitting and limited generalization, which is closely related to the challenges discussed earlier. This issue is further exacerbated in tasks that involve costly annotation processes and require data-intensive deep learning models with a large number of parameters, such as CSLR.

        \subsubsection{Mitigation Strategies and Future Directions}

            Several strategies have been proposed to mitigate data scarcity, while other promising directions remain underexplored:
            \begin{itemize}[leftmargin=*]
                \item \textbf{Multilingual Data Aggregation:} Combining datasets (e.g., CSL and DGS) to exploit cross-lingual visual similarities increases the overall amount of training data~\citep{Fangyun_Wei_ICCV}.
                \item \textbf{Self-Supervised Learning:} In the context of the self-supervised learning setting \citep{konstantakos2025self}, gloss-to-gloss translation tasks have been used for parameter initialization~\citep{CVT-SLR}, providing linguistic priors to sign language models. Pretraining with pose reconstruction has also been investigated to incorporate prior knowledge of hand structure~\citep{SignBERT+,SignBERT_ICCV,omar_ikne}.
                \item \textbf{Advanced Data Augmentation:} Adversarial learning has been employed to optimize augmentation hyperparameters~\citep{ADVERSARIAL_DATA_AUGM}, making the models robust to spatial deformations.
                \item \textbf{Few-/Zero-shot Learning}: Approaches using only a small number of samples to correctly recognize gestures---often based on CLIP---have also been proposed~\citep{ZSSLR,FSSLR,ZS-DHGR,ZS-GR,ZERO-SHOT-SLR}.
                \item \textbf{Generative AI:} Diffusion models (DMs) for image augmentation \citep{alimisis2025advances} could help address data scarcity in HGR; however they have not yet been examined. DMs can increase dataset diversity by generating additional samples with varied scenes, subjects, and lighting, or support specialized domains such as rare sign languages and human-robot interaction in firefighting or military scenarios. Thus, leveraging DMs represents a promising direction for future work.
            \end{itemize}

    \subsection{Optimization Difficulties}

        \subsubsection{Current Challenges}

\begin{figure}[H]
                \centering
                \includegraphics[width=0.8\linewidth]{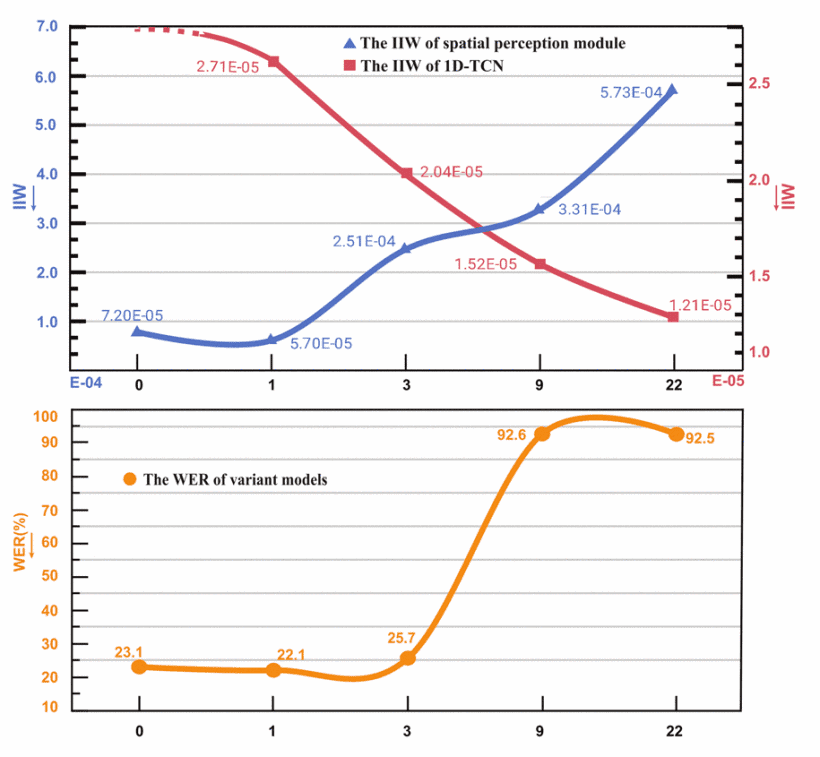}
                \caption{Experimental results from \citep{CTCA} showing the effect of the depth of the temporal aggregation module---a 1D-TCN with a varying number of layers followed by a two-layer BiLSTM---on the training of the visual encoder (spatial perception module). In both plots, the x-axis indicates the number of 1D-TCN layers. The top plot shows the information stored in weights (IIW) \citep{pac-bayes-wang2022pacbayes}, which reflects generalization ability, while the bottom plot reports the WER. Increasing the number of 1D-TCN layers improves short-term temporal generalization but degrades overall performance because the visual encoder becomes undertrained.}
                \label{fig:TAM_CTCA_DEPTH}
            \end{figure}

            Training the visual encoder effectively remains difficult, especially in deep architectures for CSLR:
            \begin{itemize}[leftmargin=*]
                \item \textbf{Gradient Attenuation:} As shown in CTCA~\citep{CTCA} (Fig.~\ref{fig:TAM_CTCA_DEPTH}), increasing the depth of the temporal module (e.g., 1D-TCN followed by BiLSTM) improves its generalization but hampers the encoder's learning due to gradient attenuation in backpropagation.
                \item \textbf{CTC Spike Phenomenon:} The model often over-focuses on a few keyframes causing rapid overfitting and limiting encoder training~\citep{VAC-CSLR,SMKD}.
            \end{itemize}
    
        \subsubsection{Mitigation Strategies and Future Directions}
        
            To address these issues, several strategies have been proposed and could be expanded in the future:
            \begin{itemize}[leftmargin=*]
                \item \textbf{Auxiliary Supervision:} Adding early-stage classifiers to the encoder or short-term module has been proposed as a strategy to guide training by providing additional supervision~\citep{VAC-CSLR,CSGC_Rao_Sun_Wang_Wang_Zhang_2024,CorrNet,corrnetplus_hu2024corrnetsignlanguagerecognition,TCNet_Lu_Salah_Poppe_2024,C2SLR}.
                \item \textbf{Temporal Module Redesign:} Placing the 1D-TCN and BiLSTM in parallel instead of sequentially~\citep{CTCA}.
                \item \textbf{Linguistic Constraints and Gloss Segmentation:} Using gloss boundaries~\citep{SMKD} or prior knowledge~\citep{GPGN} to regularize learning.
            \end{itemize}
            The aforementioned approaches -- particularly auxiliary supervision~\citep{VAC-CSLR} -- have served as the foundation for many existing methods. However, the increasing complexity of modern deep architectures often exceeds the effectiveness of these techniques. As a result, optimization remains a significant challenge -- one that may be further exacerbated by the continued development of even deeper models.

    \subsection{Limitations of Architectures}

        \subsubsection{Current Challenges}

            DL methods for HGR have increased the recognition performance. However, the existing architectures suffer from limitations, including the following:
            \begin{itemize}[leftmargin=*]
                \item \textbf{Block-fixed models:} The existing architectures are based on hand-crafted design, consisting of blocks of layers with hyperparameters (e.g., number of channels, convolution kernel size) that are defined manually by the researchers. This approach is suboptimal \citep{RAAR3DNet}, as several other possible configurations have not been explored.
                \item \textbf{Insufficient graph representations of GCNs:} The standard GCNs utilize fixed-topology graphs, often based on the shape of the hand, to better capture the interactions between the various skeleton joints. However, this configuration has been proven to be suboptimal \citep{TD-GCN}, unable to efficiently capture the temporal dynamics.
                \item \textbf{Ineffective multimodal fusion:} Out of the existing multimodal methods, most of them perform late fusion to incorporate the information of multiple streams \citep{SignBERT_ICCV,SignBERT+,BEST,Weichao-Zhao,MASA,SKIM,SAM_SLR,jiang2021sign-sam-slr-v2}. This scheme does not fully leverage the available modalities, as the features extracted by each stream are isolated from the others during backbone processing. Thus, the obtained vector representations do not benefit from the utilization of multiple modalities.
            \end{itemize}
            
        \subsubsection{Mitigation Strategies and Future Directions}

            To address these limitations, several mitigation strategies and future directions can be identified:
            \begin{itemize}[leftmargin=*]
                \item \textbf{Neural Architecture Search (NAS):} NAS can be used to identify optimal model structures by searching over a large set of configurations \citep{RAAR3DNet}.
                \item \textbf{Adaptive hand topology:} Learnable or time-dependent skeleton adjacency matrices that implicitly determine the optimal way to integrate the features of each joint \citep{DSTA_SLR_hu2024dynamic,DSTSA-CGN_cui2025dstsa,ST-PSM_L-PSM,TD-GCN}.
                \item \textbf{Multi-level fusion:} Multi-level fusion for dual-stream approaches enhances the quality of the extracted features by effectively incorporating information from various sources \citep{CTFB_Hampiholi}. To the best of our knowledge, multi-level and mid-level multimodality fusion have not been investigated for three or more streams. Thus, future work could consider the intermediate fusion in such cases, e.g. modifying existing fusion modules \citep{MMTM,CTFB_Hampiholi} and identifying the optimal way to connect the different streams.
            \end{itemize}

    \subsection{Summary of Challenges \& Research Route}

        In summary, the previous analysis indicates that future VHGR research is shaped by a small set of recurring bottlenecks rather than by isolated technical issues. The main research routes can be summarized as follows:
        \begin{itemize}[leftmargin=*]
            \item \textbf{Eliminating gesture-irrelevant factors:} Tackling the interference of redundant information, such as cluttered backgrounds, lighting variations, occlusions, and viewpoint changes, remains essential for improving real-world accuracy \citep{ULTRA-RANGE-UGV-Bamani,RAAR3DNet,Yunan_Li}.
            \item \textbf{Improving data coverage and diversity:} Larger, more diverse, and more realistic datasets are needed, especially for rare sign languages, in-the-wild settings, underrepresented modalities, and subject-independent evaluation. Multilingual aggregation, self-supervised learning, and generative augmentation are promising routes for mitigating data scarcity \citep{Fangyun_Wei_ICCV,SignBERT+,alimisis2025advances}.
            \item \textbf{Reducing computational cost:} Minimizing computational overhead is crucial for real-time and resource-constrained deployment. Lightweight architectures, non-parametric modules, and model compression provide promising directions for improving deployability \citep{LHGR-Net,GestFormer,DL_MODEL_COMPRESSION_marino2023deep}.
            \item \textbf{Addressing optimization difficulties:} The CSLR 2DCNN+1DCNN+BiLSTM pipeline still suffers from inadequate training of the lower parts of the network, necessitating more effective learning strategies, auxiliary supervision, and temporal modeling schemes \citep{GPGN,SMKD,CTCA}.
            \item \textbf{Advancing multi-stream fusion:} For architectures involving more than two streams, late fusion remains dominant \citep{SAM_SLR,jiang2021sign-sam-slr-v2}. More sophisticated techniques, such as multi-level and intermediate fusion, could improve cross-modal feature extraction and reduce the redundancy of independent streams.
            \item \textbf{Incorporating foundation models efficiently:} Foundation models can support stronger visual encoding, linguistic priors, and open-vocabulary recognition, but their benefits have not yet been fully exploited in VHGR and currently come with significant computational cost \citep{CLIP_SLA_alyami2025clip_CSLR_2025,UNISIGNli2025,kas2025foundation_models}.
        \end{itemize}

\section{Conclusions}
\label{sec:conclusions}

    The undeniable importance of Vision-based Hand Gesture Recognition (VHGR) stems from its wide range of applications---spanning from human-robot interaction to sign language understanding. This broad applicability has made it a vibrant and active research field. Recent advances in deep learning (DL) have played a pivotal role in driving progress, enabling the development of increasingly sophisticated architectures and training strategies aimed at tackling more complex and fine-grained tasks, such as continuous sign language recognition (CSLR).
    
    The current study aims to address four key research questions, as defined in Section \ref{sec:intro}. Specifically, this survey determines the principal aspects of VHGR by providing a thorough taxonomy in Section \ref{sec:taxonomy} (answering Q1) and identifies the current SOTA and key developments for each field following a task-oriented presentation (Sections \ref{sec:static}, \ref{sec:isolated}, \ref{sec:continuous} and \ref{sec:comparison}), as required by Q2. Additionally, it performs a comparative analysis of the selected publications in the aforementioned sections to address Q3 and highlights challenges and future research directions in Section \ref{sec:challenges}, as requested by Q4. The summaries of datasets (Section \ref{sec:datasets}) and evaluation metrics (Section \ref{sec:metrics}) are considered essential for facilitating the previous analysis.
    
    This study focuses on DL-based VHGR methods. These methods were highlighted not only for their superior performance compared to traditional hand-crafted approaches, but also for their greater potential for deployment in real-world scenarios, unlike more cumbersome sensor-based techniques. From the surveyed literature, it becomes clear that even relatively simple and shallow DL models can achieve high accuracy in static HGR tasks. In such cases, the primary challenges are typical of general computer vision problems and are not inherently tied to HGR itself. For isolated dynamic and continuous VHGR, more advanced approaches have been proposed, aiming to capture more complex context. The complexity of the architectures (especially for CSLR) causes serious optimization difficulties, leading researchers to develop sophisticated training setups.

    Nevertheless, the most important challenges regarding VHGR remain unsolved, leaving substantial research gaps; data scarcity---especially for more specialized tasks, such as the recognition of rare sign languages---and high computational cost remain essentially unresolved. Furthermore, the existing methods are limited to a predefined list of gestures, which may not be optimal for real-world interaction with humanoid robots or interpretation of body language. Thus, VHGR could be extended to open-vocabulary settings, enabling deployment in broader applications. The potential of leveraging the extensive knowledge of foundation models \citep{kas2025foundation_models} in the context of VHGR enabling zero-shot gesture recognition, as well as the utilization of model compression techniques \citep{DL_MODEL_COMPRESSION_marino2023deep} to reduce the computational complexity are also rarely explored, leaving space for future work.

    \section*{Declarations}
    \textbf{Funding} The research leading to these results received funding from the European Commission under Grant Agreement No. 101168042 (TRIFFID) and No. 101189557 (TORNADO).\\\\
    \textbf{Conflict of interest} The authors have no competing interests to declare that are relevant to the content of this article.\\\\
    \textbf{Author contribution} 
    Konstantinos Foteinos, Manousos Linardakis, Iraklis Varlamis and Georgios Th. Papadopoulos performed the literature review and prepared a draft of the manuscript. Panagiotis Radoglou-Grammatikis, Vasileios Argyriou and Panagiotis Sarigiannidis reviewed and edited the manuscript. Georgios Th. Papadopoulos was responsible for securing the funding to implement the study.

\bibliographystyle{elsarticle-harv}
\bibliography{references}

\end{document}